\documentclass[10pt,twocolumn,letterpaper]{article}

\usepackage[pagenumbers]{cvpr}      % To produce the REVIEW version
\usepackage{pifont}

\usepackage{graphicx}
\usepackage{stfloats} 
\usepackage{balance}
\usepackage{fdsymbol}

\definecolor{cvprblue}{rgb}{0.21,0.49,0.74}
\usepackage[pagebackref,breaklinks,colorlinks,allcolors=cvprblue]{hyperref}

\def\paperID{1666} % *** Enter the Paper ID here
\def\confName{CVPR}
\def\confYear{2026}

\title{UARE: A Unified Vision-Language Model for Image Quality Assessment, Restoration, and Enhancement
}

\author{
Weiqi Li$^{1}$,
Xuanyu Zhang$^{1,2}$,
Bin Chen$^{1,2}$,
Jingfen Xie$^{2}$,
Yan Wang$^{2}$,
Kexin Zhang$^{2}$,\\
Junlin Li$^{2}$,
Li Zhang$^{2}$,
Jian Zhang$^{1}$$^\dagger$,
Shijie Zhao$^{2}$$^\vardiamondsuit$$^\dagger$\\
$^1$School of Electronic and Computer Engineering, Peking University, $^2$ByteDance Inc.
}

\usepackage{xcolor}
\newcommand{\best}[1]{\textbf{\textcolor{red}{#1}}}
\newcommand{\secondbest}[1]{\underline{\textcolor{blue}{#1}}}
\newcommand{\third}[1]{\textcolor{ForestGreen}{\textit{#1}}}
\usepackage{multirow}
\usepackage{nicematrix}
\usepackage{bbm}
\usepackage{makecell}
\usepackage{float}
\newcommand\shline{\specialrule{1.0pt}{0pt}{0pt}}

\begin{document}
% \maketitle

\twocolumn[{
\renewcommand\twocolumn[1][]{#1}
\maketitle
\centering
% \vspace{-0.8cm}
\includegraphics[width=\textwidth]{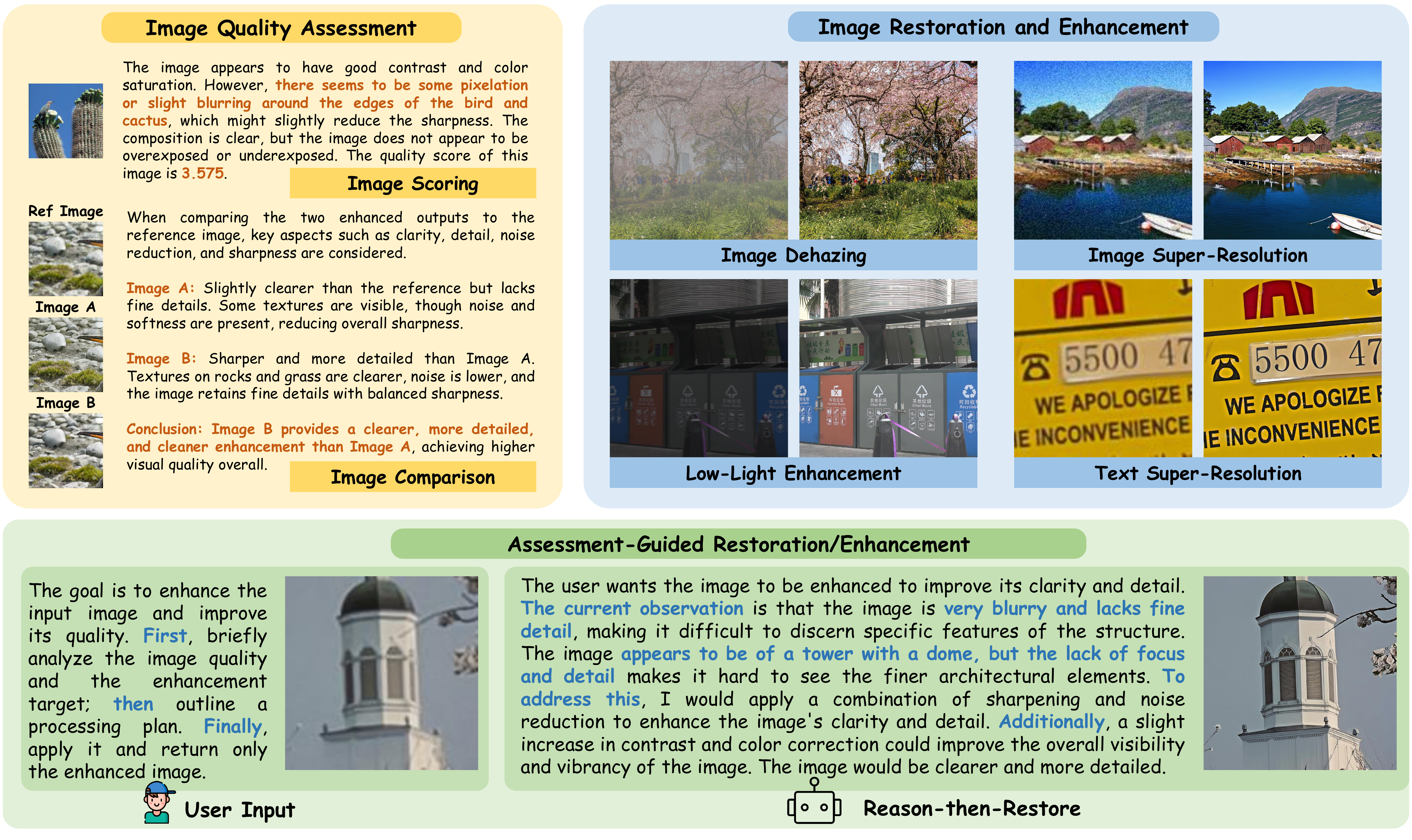}
\vspace{-0.7cm}
\captionsetup{type=figure}
\caption{\textbf{Showcase of UARE.} It supports image quality assessment (image scoring and comparison), image restoration/enhancement (super-resolution, dehazing, and low-light enhancement, etc.), and assessment-guided restoration and enhancement in \textbf{one unified model}.}
}
\vspace{0.4cm}]

\renewcommand*{\thefootnote}{$\vardiamondsuit$}
\footnotetext[1]{Project Lead. $\dagger$: Corresponding authors, zhangjian.sz@pku.edu.cn, zhaoshijie.0526@bytedance.com.}

\begin{abstract}
Image quality assessment (IQA) and image restoration are fundamental problems in low-level vision. Although IQA and restoration are closely connected conceptually, most existing work treats them in isolation. Recent advances in unified multimodal understanding-generation models demonstrate promising results and indicate that stronger understanding can improve generative performance. This motivates a single model that unifies IQA and restoration and explicitly studies how IQA can guide restoration, a setting that remains largely underexplored yet highly valuable. In this paper, we propose \textbf{UARE}, to our knowledge the first \textbf{U}nified vision-language model for image quality \textbf{A}ssessment, \textbf{R}estoration, and \textbf{E}nhancement. Built on pretrained unified understanding and generation models, we introduce a two-stage training framework. First, a progressive, easy-to-hard schedule expands from single-type distortions to higher-order mixed degradations, enabling UARE to handle multiple degradations. Second, we perform unified fine-tuning of quality understanding and restoration with interleaved text-image data, aligning IQA signals with restoration objectives. Through multi-task co-training, UARE leverages IQA to boost restoration and enhancement performance. Extensive experiments across IQA, restoration, and enhancement tasks demonstrate the effectiveness of UARE. The code and models will be available at \url{https://github.com/lwq20020127/UARE}.
\end{abstract}

\section{Introduction}
Image quality assessment (IQA)~\cite{wang2004image,lin2019kadid,zhang2011fsim,zhang2015feature} and image restoration/enhancement~\cite{dong2014learning,timofte2015a+,zhang2018image,liang2021swinir,wang2021real,ledig2017photo,DA-CLIP} are fundamental, long-standing problems in low-level vision. IQA aims to predict human-perceived quality, either as scalar scores~\cite{yang2022maniqa,wu2023QAlign,wang2024exploiting,you2025teaching} or as natural-language descriptions~\cite{you2024DQA,you2024DQAW,li2025qinsight,zhang2025vqinsight}, supporting tasks such as photo selection, camera capture tuning, and large-scale monitoring of streaming and video quality. Image restoration~\cite{liang2021swinir,DGUNet,PromptIR,DiffIR,DiffUIR} seeks to recover a clean image from a degraded observation, while enhancement~\cite{LOL,LOL-Blur,UHD-LL,wang2024zero} improves perceptual quality when no pristine reference is available. Typical applications include real-world photography, perception under adverse weather or low-light conditions, and compression-heavy media delivery.

IQA and restoration are logically closely connected: advances in IQA can guide the design of restoration tools that better match human perceptual preferences, while progress in restoration provides feedback to refine and calibrate IQA. However, most existing work treats them in isolation. On the one hand, IQA methods focus solely on assessment and do not consider compatibility with downstream restoration models. For example, recent multimodal large language model (MLLM)-based IQA methods produce detailed quality descriptions~\cite{you2024DQA,you2024DQAW,li2025qinsight,zhang2025vqinsight} but do not address how restoration methods can leverage such text. On the other hand, most existing restoration methods~\cite{li2024foundir,DiffIR,Real-esrgan} prioritize producing better results without leveraging IQA context. As a result, the two lines remain largely isolated.

Recently, multimodal understanding and generation have advanced rapidly, with numerous methods jointly optimizing both within a unified architecture~\cite{team2024chameleon,showo,deng2025bagel,tong2024metamorph,wu2024janus,ma2024janusflow}. For example, Chameleon~\cite{team2024chameleon} treats images as discrete tokens and unifies understanding and generation via next-token prediction. Transfusion~\cite{zhou2024transfusion} integrates diffusion modeling and switches to a diffusion mode at generation time. Bagel~\cite{deng2025bagel} employs a mixture-of-transformers (MoT) design, decoupling parameters for understanding and for generation while sharing multimodal self-attention. These approaches achieve strong results and suggest that a stronger understanding improves generation performance in high-level applications. This trend motivates a similar direction in low-level vision: the relationship between IQA and restoration, to some extent, mirrors that between understanding and generation. \textit{Therefore, unifying IQA and restoration in a single model and studying how IQA can guide image restoration and enhancement remains largely underexplored and highly valuable}.

% However, unified modeling of IQA and restoration in low-level vision remains largely unexplored and warrants further study.
% Due to the similar inherent conncetion between IQA and restoation, it remains underexplored
% and of great value to unify them in one model and expore the guidance of IQA for restoration.

In this paper, we propose \textbf{UARE}, to our knowledge, the first unified vision-language model for image quality assessment, restoration, and enhancement. As shown in Fig.~\textcolor{cvprblue}{1}, UARE supports image quality assessment tasks such as image scoring and comparison, and restoration tasks such as super-resolution, dehazing, and low-light enhancement. All these tasks are handled by a single model. Notably, UARE also enables assessment-guided restoration in a reason-then-restore style, showing that IQA can benefit restoration. To achieve this, we adopt a two-stage training framework based on a MoT backbone. The first stage employs a progressive easy-to-hard scheme that moves from single-type distortions~\cite{li2024foundir} to higher-order mixed degradations~\cite{Real-esrgan,wang2024apisr}, enabling UARE to handle multiple degradations. The second stage performs unified fine-tuning of quality understanding and restoration using interleaved text-image data. This strengthens quality assessment and aligns its signals with restoration objectives. Our empirical study shows that quality assessment boosts restoration performance through multi-task co-training. Extensive experiments across diverse image quality assessment, restoration, and enhancement tasks demonstrate the effectiveness of UARE. In summary, our contributions are:

\noindent\ding{113} We present \textbf{UARE}, to our knowledge, the first \textbf{U}nified vision-language model that enables image quality \textbf{A}ssessment (image scoring, description, and pairwise comparison) and image \textbf{R}estoration/\textbf{E}nhancement (real-world super-resolution, deblurring, denoising, deraining, dehazing, low-light enhancement, etc.) within a single model.

\noindent\ding{113} We introduce a two-stage training framework: (1) a progressive easy-to-hard training schedule that moves from single-type distortions to high-order mixed degradations; and (2) unified fine-tuning of quality assessment and restoration using interleaved text–image data, aligning assessment signals with restoration objectives.

\noindent\ding{113}  We find that image quality assessment boosts restoration through multi-task co-training. Extensive experiments across diverse image quality assessment, restoration, and enhancement tasks demonstrate the effectiveness of UARE.

\begin{figure*}[!t]
\centering
\includegraphics[width=\linewidth]{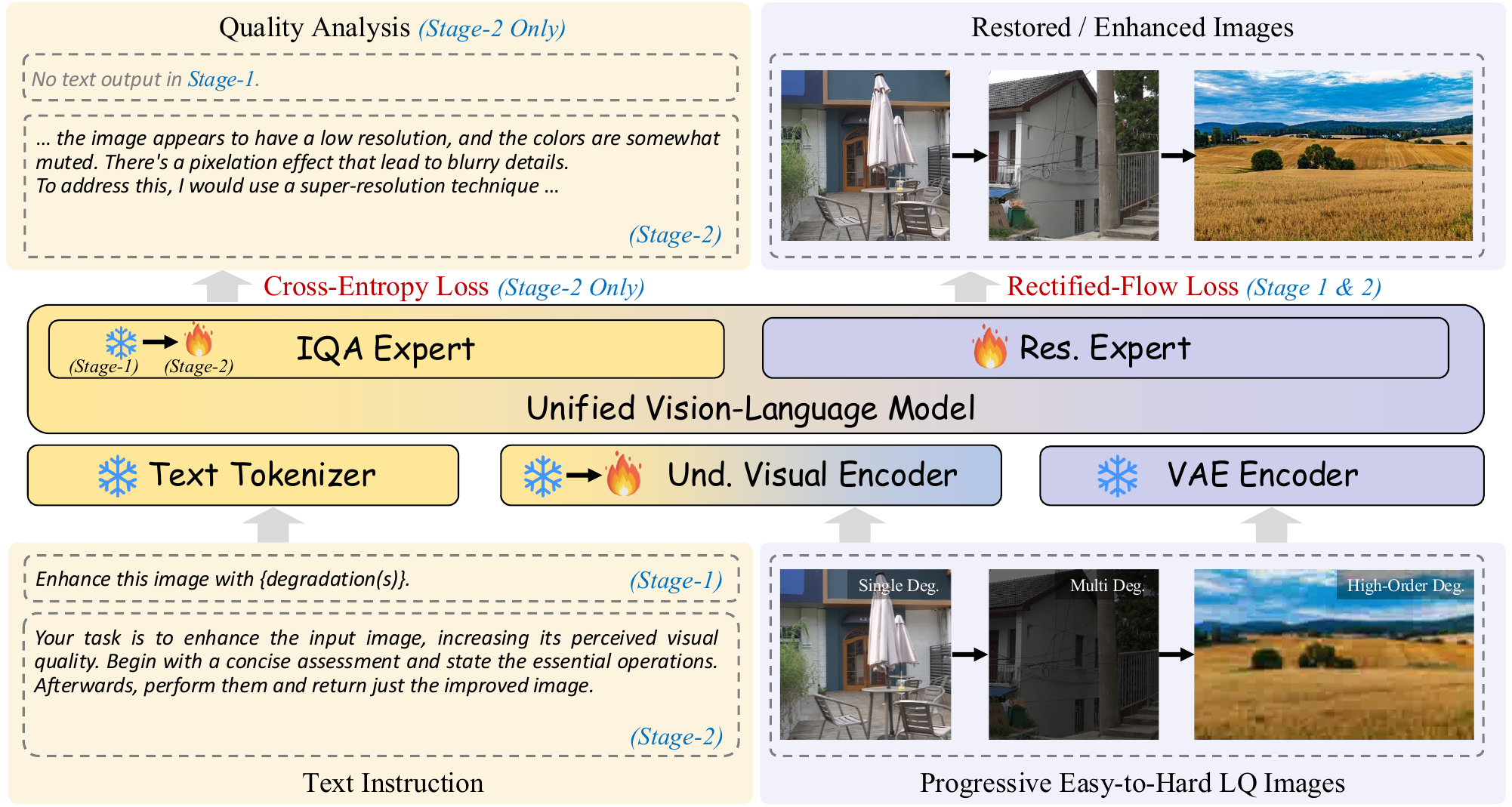}
\vspace{-14pt}
\caption{\textbf{Illustration of the architecture and two-stage training framework of UARE.} Two transformer experts are used to process IQA and restoration, respectively. Training stages include (1) a progressive, easy-to-hard schedule that moves from single-type to high-order degradations. In this stage, only the restoration expert is trained to make UARE handle multiple degradations. (2) Unified fine-tuning of the entire model to strengthen the IQA ability and align the IQA signals with restoration objectives through interleaved data.}
\label{fig:framework}
\vspace{-6pt}
\end{figure*}

\section{Related Work}
\subsection{Image Quality Assessment}
Traditional studies categorize IQA methods into two paradigms: full-reference and no-reference.
Full-reference approaches~\cite{wang2004image,sheikh2006image,zhang2011fsim} evaluate the perceptual similarity between a degraded image and its ground-truth counterpart, relying on both classic metrics such as PSNR~\cite{wang2004image} and learning-based perceptual measures~\cite{bosse2017deep,cao2022incorporating,ding2020image,ding2021locally,ghildyal2022shift,prashnani2018pieapp} including LPIPS~\cite{zhang2018unreasonable}. No-reference IQA methods predict visual quality without access to a clean reference, evolving from handcrafted natural scene statistics~\cite{ma2017learning,mittal2012no,mittal2012making,moorthy2010two,moorthy2011blind,saad2012blind} to data-driven neural models that learn quality priors directly from large-scale datasets~\cite{kang2014convolutional,ke2021musiq,liu2017rankiqa,pan2018blind,su2020blindly,zheng2021learning,zhu2020metaiqa,sun2022graphiqa,wang2024exploiting}. With the emergence of multimodal large language models (MLLMs), researchers have begun leveraging their vision-language understanding for image quality evaluation.
Score-based frameworks like Q-Align~\cite{wu2023QAlign} and DeQA-Score~\cite{you2025teaching} generate numerical assessments by exploiting MLLMs’ perception and factual knowledge. Description-driven methods~\cite{wu2023qbench,wu2024qinstruct,you2024DQA,you2024DQAW,wu2024towards,chen2024q,zhang2025qeval,zhang2025teaching,zhang2024qbenchvideo} emphasize interpretability and richer explanations, typically depending on extensive textual supervision for fine-tuning.
Recently, reinforcement learning has been incorporated into IQA tasks~\cite{li2025qinsight,zhang2025vqinsight,wu2025visualquality,zhao2025reasoning}, enabling models to jointly produce reasoning-based descriptions and quantitative scores, achieving stronger generalization.

\subsection{Universal Image Restoration and Enhancement}
Image restoration~\cite{liang2021swinir,DGUNet,chen2022real} seeks to recover a clean image from a degraded observation, with typical tasks including image super-resolution~\cite{lin2024diffbir,wu2024seesr,yue2024resshift,wang2024sinsr}, denoising~\cite{BSD,Urban}, deblurring~\cite{4KRD}, and deraining~\cite{Rain100}, among others. Image enhancement~\cite{LOL,LOL-Blur,UHD-LL,wang2024zero} improves perceptual quality when no pristine reference is available, with low-light enhancement as a representative example. Universal image restoration and enhancement aim to handle multiple tasks within a single model, which is challenging because degradation factors can be distinct and even mutually exclusive~\cite{GRIDS,Onerestore,kong2024towards}. To address this challenge, PromptIR~\cite{PromptIR} learns input-modulated embeddings that adaptively instruct the network to address different degradations. Diffusion-based methods~\cite{liu2024residual,DiffIR,DiffUIR} model degradation distributions using powerful diffusion priors. Pretrained multimodal large models have also been integrated into recent methods~\cite{MPerceiver,DA-CLIP,SUPIR} to enable high-fidelity, universal image restoration. FoundIR~\cite{li2024foundir} introduces a million-scale dataset covering twenty types of single and mixed degradations and presents a degradation-agnostic generalist model. Our UARE moves beyond existing all-in-one approaches by incorporating IQA capability within a unified model.

\subsection{Unified Understanding and Generation}
Unified multimodal architectures that jointly support image understanding and generation have emerged as a promising direction. Some methods~\cite{team2024chameleon,wang2024emu3,wu2024janus,chen2025januspro} treat image patches as discrete tokens and generate images with autoregressive next token prediction, as exemplified by Chameleon~\cite{team2024chameleon}. Other works~\cite{zhou2024transfusion,showo} integrate diffusion~\cite{dhariwal2021diffusion} to predict discrete text tokens and to model continuous images. A further line decouples understanding and generation by introducing external diffusion models~\cite{tong2024metamorph,wang2024illume,huang2025illume+,lin2025uniworld,chen2025blip3}. Recently, Bagel~\cite{deng2025bagel} adopts a mixture-of-transformers (MoT) design and decouples parameters for understanding and for generation while sharing multimodal self-attention. These methods show promising performance and indicate that improved understanding can enhance generation. In low-level vision, unifying IQA and restoration in a single model and studying how IQA can guide restoration remains largely underexplored and highly valuable. PURE~\cite{wei2025perceive} explores joint image degradation estimation and super-resolution within an autoregressive paradigm, but it outputs only simple degradation descriptions and performs only super-resolution. Moreover, PURE has not explored how IQA influences enhancement. Distinctly, our UARE is a more general-purpose unified model for IQA and universal image restoration and demonstrates that IQA can boost restoration.

\section{Methodology}
\subsection{Challenges}
Incorporating IQA and restoration into a unified framework presents three significant challenges: \textit{\textbf{Firstly,}} IQA and restoration are distinct tasks with different training objectives. The challenge lies in designing the model such that both tasks can be optimized simultaneously without causing performance degradation due to conflicting goals. \textit{\textbf{Secondly,}} can IQA truly benefit restoration? Specifically, how can the meaningful quality-assessment texts generated by IQA be effectively utilized by restoration? \textit{\textbf{Thirdly,}} within the various sub-tasks of image restoration, degradation types and intensities vary widely. How can the model balance performance across these diverse cases in a unified restoration framework? To address the above challenges, we introduce the network architecture, data construction pipeline, and training framework of UARE as follows. 

% Secs.~\ref{sec:3.2},~\ref{sec:3.3}, and~\ref{sec:3.4}, respectively.
% challenge，data construction，training framework, network structure

\begin{table*}[t]
\centering
\caption{\textbf{Quantitative comparison of different SR methods} on RealSR, DRealSR, and DIV2K. Throughout this paper, best, second-best, and third-best results are highlighted in \best{bold red}, \secondbest{underlined blue}, \third{italic green}. $\uparrow$ / $\downarrow$ indicates higher/lower is better.}
\vspace{-10pt}
\resizebox{\linewidth}{!}{
\renewcommand{\arraystretch}{1.15}
\begin{tabular}{l | l | c c c c c c c c c}
\shline
Test Dataset & Method & PSNR$\uparrow$ & SSIM$\uparrow$ & LPIPS$\downarrow$ & DISTS$\downarrow$ & NIQE$\downarrow$ & LIQE$\uparrow$ & MUSIQ$\uparrow$ & MANIQA$\uparrow$ & TOPIQ$\uparrow$ \\
\hline \hline
% ================= Set-A =================
\multirow{10}{*}{RealSR}
& StableSR~\cite{wang2024exploiting} & 23.73 & \third{0.6979} & 0.2792 & \third{0.2023} & 5.5914 & 3.0532 & 61.65 & 0.3826 & 0.5201 \\
& DiffBIR~\cite{lin2024diffbir} & 23.20 & 0.6346 & 0.3350 & 0.2162 & \best{4.5879} & 3.5529 & 65.25 & 0.4620 & 0.6033 \\
& SeeSR~\cite{wu2024seesr} & \secondbest{24.34} & \best{0.7187} & \third{0.2754} & 0.2134 & 6.4146 & 3.3938 & 65.53 & \third{0.4856} & 0.6246 \\
& PASD~\cite{yang2024pixel} & \best{24.50} & \secondbest{0.7115} & \best{0.2716} & \best{0.1954} & 6.0067 & 2.8541 & 58.52 & 0.3831 & 0.4969 \\
& ResShift~\cite{yue2024resshift} & \third{24.17} & 0.6528 & 0.4336 & 0.2812 & 8.6273 & 2.6610 & 53.38 & 0.3412 & 0.4210 \\
& SinSR~\cite{wang2024sinsr} & 23.68 & 0.6649 & 0.3490 & 0.2445 & 6.5101 & 3.2255 & 61.03 & 0.4230 & 0.5383 \\
& OSEDiff~\cite{wu2024one} & 23.07 & 0.6850 & 0.2941 & 0.2109 & 5.5054 & \best{4.0681} & \secondbest{68.95} & \secondbest{0.4876} & \secondbest{0.6441} \\
& S3Diff~\cite{zhang2024degradation} & 23.16 & 0.6810 & \secondbest{0.2748} & \secondbest{0.1986} & \third{5.3003} & \third{4.0080} & \third{67.57} & 0.4677 & \third{0.6301} \\
& PURE~\cite{wei2025perceive} & 21.31 & 0.5738 & 0.3859 & 0.2468 & 5.6419 & 3.7881 & 66.57 & 0.4829 & 0.6301 \\
& \textbf{UARE (Ours)} & 21.38 & 0.6464 & 0.3095 & 0.2344 & \secondbest{5.2981} & \secondbest{4.0658} & \best{69.67} & \best{0.5260} & \best{0.6796} \\
\hline \hline 
% ================= Set-B =================
\multirow{10}{*}{DRealSR}
&StableSR~\cite{wang2024exploiting}     & \best{28.28} & \best{0.7981} & \best{0.2687} & \best{0.2026} & 7.2816 & 2.5068 & 51.62 & 0.3226 & 0.4355 \\
&DiffBIR~\cite{lin2024diffbir}         & 26.08 & 0.6578 & 0.4144 & 0.2564 & \best{4.4856} & 3.3993 & 61.81 & 0.4612 & 0.6084 \\
&SeeSR~\cite{wu2024seesr}              & \third{28.14} & \secondbest{0.7798} & \secondbest{0.2832} & 0.2241 & 7.4833 & 2.7943 & 55.89 & 0.3976 & 0.5436 \\
&PASD~\cite{yang2024pixel}             & \secondbest{28.18} & \third{0.7722} & \third{0.2970} & \third{0.2108} & 7.4421 & 2.6129 & 51.42 & 0.3595 & 0.4587 \\
&ResShift~\cite{yue2024resshift}       & 27.39 & 0.6907 & 0.4996 & 0.3077 & 9.1788 & 1.7905 & 40.58 & 0.2457 & 0.3414 \\
&SinSR~\cite{wang2024sinsr}            & 26.72 & 0.6933 & 0.4031 & 0.2624 & 6.8825 & 2.7781 & 53.36 & 0.3677 & 0.4959 \\
&OSEDiff~\cite{wu2024one}              & 25.60 & 0.7403 & 0.3088 & 0.2158 & \third{6.1544} & \secondbest{3.9797} & \secondbest{65.24} & \secondbest{0.4879} & \secondbest{0.6273} \\
&S3Diff~\cite{zhang2024degradation}    & 26.18 & 0.7197 & 0.3161 & \secondbest{0.2099} & \secondbest{5.9531} & \third{3.9255} & \third{63.34} & \third{0.4635} & \third{0.6181} \\
&PURE~\cite{wei2025perceive}           & 23.04 & 0.5718 & 0.4461 & 0.2674 & 6.3939 & 3.7390 & 60.68 & 0.4362 & 0.5888 \\
&\textbf{UARE (Ours)}                           & 21.31 & 0.5736 & 0.4071 & 0.2613 & 6.4290 & \best{4.0445} & \best{67.71} & \best{0.5121} & \best{0.6652} \\
\hline \hline
% ================= Set-C =================
\multirow{10}{*}{DIV2K}
& StableSR~\cite{wang2024exploiting} & \best{19.85} & \best{0.4940} & 0.4796 & 0.2887 & 5.7479 & 1.8466 & 43.25 & 0.2181 & 0.3276 \\
& DiffBIR~\cite{lin2024diffbir} & 18.94 & 0.4332 & 0.4009 & 0.2238 & \best{3.6594} & 3.8573 & 67.20 & 0.4574 & 0.6467 \\
& SeeSR~\cite{wu2024seesr} & \third{19.11} & \secondbest{0.4580} & \third{0.3769} & 0.2339 & 4.5817 & 3.7445 & 66.31 & \third{0.4686} & 0.6330 \\
& PASD~\cite{yang2024pixel} & 18.98 & 0.4562 & 0.4293 & 0.2373 & 4.7846 & 3.6022 & 63.46 & 0.4025 & 0.5653 \\
& ResShift~\cite{yue2024resshift} & \secondbest{19.15} & 0.4311 & 0.4900 & 0.2808 & 7.4321 & 2.8862 & 56.02 & 0.3534 & 0.4662 \\
& SinSR~\cite{wang2024sinsr} & 18.58 & 0.4059 & 0.4483 & 0.2455 & 6.0533 & 3.4629 & 64.12 & 0.4483 & 0.5997 \\
& OSEDiff~\cite{wu2024one} & 18.86 & \third{0.4563} & \secondbest{0.3579} & \third{0.2209} & \third{4.1756} & 3.8877 & 67.83 & 0.4422 & 0.6269 \\
& S3Diff~\cite{zhang2024degradation} & 18.76 & 0.4490 & \best{0.3299} & \best{0.1990} & 4.2026 & \secondbest{4.2692} & \third{69.31} & 0.4675 & \secondbest{0.6679} \\
& PURE~\cite{wei2025perceive} & 16.71 & 0.3661 & 0.4449 & 0.2293 & 4.9545 & \best{4.2701} & \secondbest{70.06} & \best{0.5201} & \third{0.6621} \\
& \textbf{UARE (Ours)} & 16.59 & 0.3857 & 0.4074 & \secondbest{0.2138} & \secondbest{3.7931} & \third{4.2627} & \best{70.45} & \secondbest{0.5028} & \best{0.6864} \\
\shline
\end{tabular}}
\label{tab:SR}
\vspace{-12pt}
\end{table*}

\subsection{Network Architecture}
\label{sec:3.2}
An overview of UARE is illustrated in Fig.~\ref{fig:framework}. UARE is built upon a mixture-of-transformers (MoT) design~\cite{deng2025bagel}. Specifically, the network has two full-capacity experts: an IQA expert that receives text tokens from a text tokenizer and image tokens from an understanding visual encoder~\cite{tschannen2025siglip}, and a restoration expert that receives VAE~\cite{kingma2013auto} latent tokens. Both experts operate on a single interleaved token stream with shared self-attention in every block. The shared context preserves cross-modal alignment, while separate expert parameters reduce gradient interference between IQA and restoration, effectively alleviating task conflict. Illustrations of the three tasks' data flow are shown below.

\noindent \textbf{IQA:} \textbf{text} + \textbf{image} $\longrightarrow$ \textbf{text.} In this task, the input text instruction is firstly tokenized and encoded by the text encoder, and the input image is patched-embedded by the understanding visual encoder. The mixed stream is then processed by the IQA expert to produce an output answer text, while the restoration expert remains inactive.

\noindent \textbf{Image restoration:} \textbf{text} + \textbf{image} $\longrightarrow$ \textbf{image.} In text-guided restoration, a short command such as ``enhance this image with noise'' and a low-quality (LQ) image are fed into UARE together. The IQA expert encodes the command into instruction tokens, inserts these tokens into the shared token stream, and keeps them active in attention. The restoration expert attends to these tokens while processing the VAE latents, uses them to steer the latent updates toward the requested operation, refines the latents, and finally, the decoder produces a high-quality (HQ) image.

\noindent \textbf{Assessment-guided  restoration:} \textbf{text} + \textbf{image} $\longrightarrow$ \textbf{text} + \textbf{image.} In this task, an instruction such as “analyze the image quality and enhance” and an LQ image are given as input. The IQA expert first generates a compact quality analysis and enhancement recommendations. These analysis tokens remain in the shared stream. The restoration expert conditions on them through shared self-attention when updating the VAE latents. Consequently, the final latents and the decoded HQ image are guided by the analysis, and the model outputs both the analysis text and the restored image.

\subsection{Data Construction}
\label{sec:3.3}

Training a unified model requires large, high-quality, and task-diverse data. Notably, to study how IQA can benefit restoration, we construct an interleaved set of quality analysis text–image pairs so that, during training, the restoration expert can condition on outputs from the IQA expert, and both experts are updated in a single pass.

\noindent \textbf{IQA.} Our IQA data cover three tasks: image scoring, image degradation description, and image comparison. For image scoring, we utilize KONIQ~\cite{hosu2020koniq}, KADID~\cite{lin2019kadid}, PIPAL~\cite{jinjin2020pipal}, and SPAQ~\cite{fang2020perceptual}. The input includes an image and a question drawn from a small template pool, and the output is a quality description with a quality score. For image degradation description, the images and corresponding descriptions are obtained from DQ-495K~\cite{you2024DQAW}. The images in this dataset include common degradation types such as blur, noise, JPEG artifacts, and low-light, with multiple intensity levels. The descriptions cover both the image content and the specific degradations present in the image. For image comparison, we use DiffIQA~\cite{Chen2025afine}. Each sample contains a reference image and two super-resolved images produced by different methods. The output is a comparison analysis with quality descriptions for both images, generated using Q-Insight~\cite{li2025qinsight} by prompting the model to compare the two results in terms of fidelity and perceptual quality.

\noindent \textbf{Restoration and enhancement.} Our restoration and enhancement dataset spans single, multiple, and high-order degradations. Data with single and multiple degradations are obtained from FoundIR~\cite{li2024foundir}, which includes blur, noise, JPEG artifacts, haze, rain, and low-light. The multiple degradation data consists of compositions of the individual degradations. For high-order degradations, we follow the degradation processes used in RealESRGAN~\cite{Real-esrgan} and APISR~\cite{wang2024apisr}, applying degradation sequences to HQ ground-truth images from LSDIR~\cite{LSDIR} and FoundIR~\cite{li2024foundir}. Across all tasks, HQ images are cropped so that the longer side is 1024 pixels. For high-order degradations, we apply a final downsampling by a factor of 4. For single and multiple degradation tasks, the input instruction is ``enhance this image with {degradation}". For high-order degradation tasks, the input text is ``enhance this mix-degraded image."

\noindent \textbf{t} Building on the LQ-HQ image pairs from the above restoration datasets (with simple instructions), we create interleaved data that incorporates more detailed input and output text descriptions. Specifically, we generate input instructions from a prompt pool following an analyze-then-restore style, such as: ``Enhance the input image. Provide a brief analysis of defects and the target appearance, outline the main steps to address them, then apply the enhancements and return the improved image.'' The corresponding output text follows a four-step structure, including: (1) user intent, (2) current quality analysis, (3) enhancement plan, and (4) expected result. This structure enables the model to explicitly link quality analysis with restoration actions in a coherent and structured manner.

\begin{figure*}[!t]
\setlength{\tabcolsep}{0.5pt}
\centering
\footnotesize
\resizebox{\linewidth}{!}{%
\begin{minipage}{\linewidth}

% ---------- 第一行小图 ----------
\resizebox{\linewidth}{!}{\begin{tabular}{@{}ccccccc@{}}
Input & StableSR \cite{wang2024exploiting} & PASD \cite{yang2024pixel} & ResShift \cite{yue2024resshift} & S3Diff \cite{zhang2024degradation} & PURE~\cite{wei2025perceive} & \textbf{UARE (Ours)}\\
\includegraphics[width=0.14\textwidth]{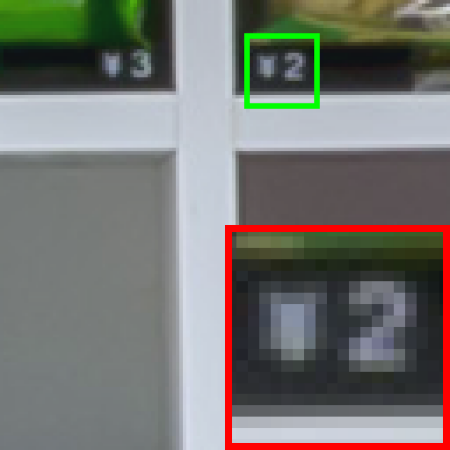}
&\includegraphics[width=0.14\textwidth]{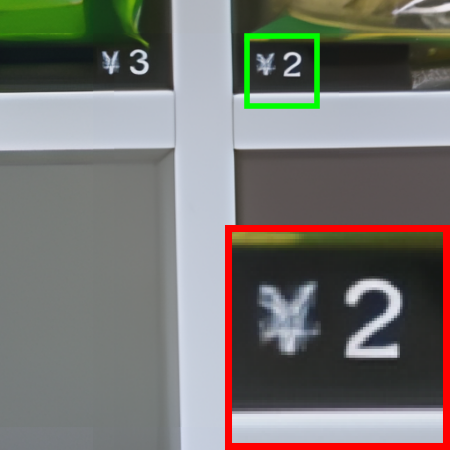}
&\includegraphics[width=0.14\textwidth]{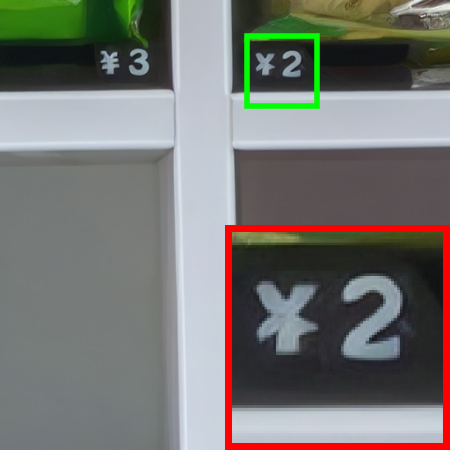}
&\includegraphics[width=0.14\textwidth]{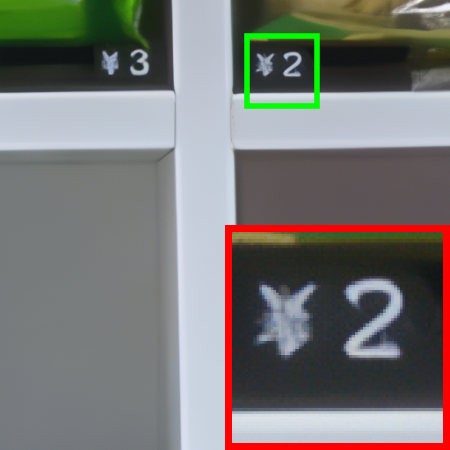}
&\includegraphics[width=0.14\textwidth]{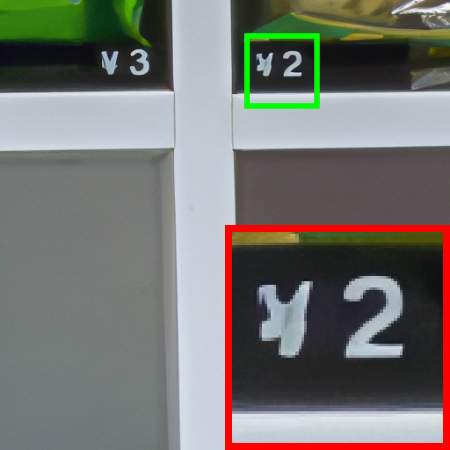} 
&\includegraphics[width=0.14\textwidth]{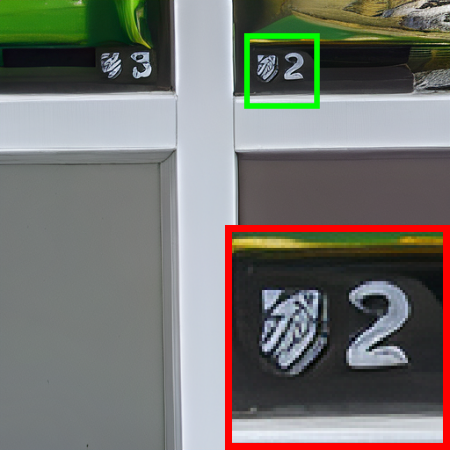} 
&\includegraphics[width=0.14\textwidth]{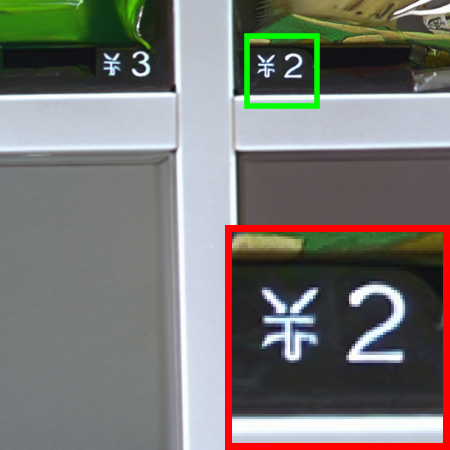}\\
\end{tabular}}

\vspace{-2pt}
% ---------- 中间整行图（与上表同宽，天然对齐） ----------
\includegraphics[width=\linewidth]{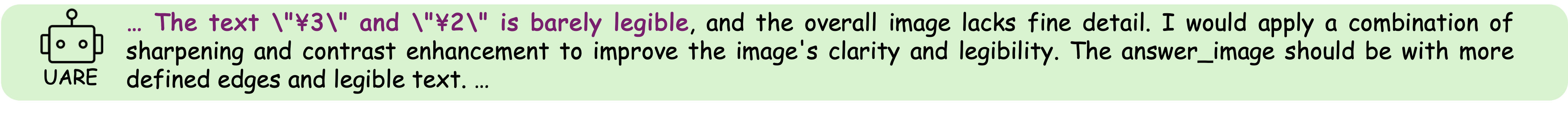}
\vspace{-15pt}

% ---------- 第二行小图 ----------
\resizebox{\linewidth}{!}{\begin{tabular}{@{}ccccccc@{}}
Input & DiffBIR \cite{lin2024diffbir} & SeeSR \cite{wu2024seesr} & SinSR \cite{wang2024sinsr} & OSEDiff \cite{wu2024one} & PURE~\cite{wei2025perceive} & \textbf{UARE \selectfont (Ours)}\\
\includegraphics[width=0.14\textwidth]{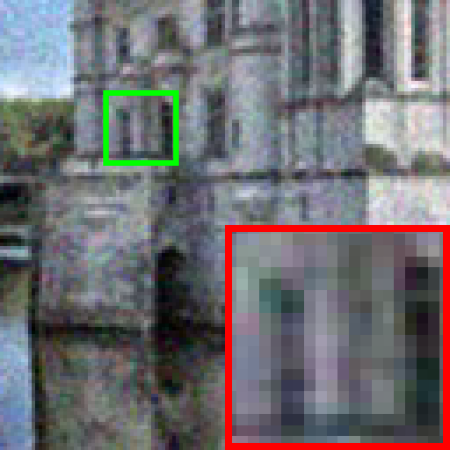}
&\includegraphics[width=0.14\textwidth]{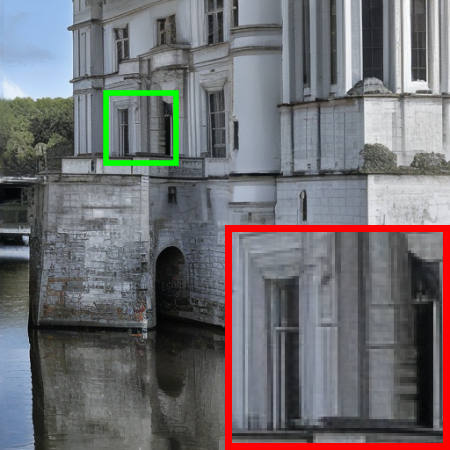}
&\includegraphics[width=0.14\textwidth]{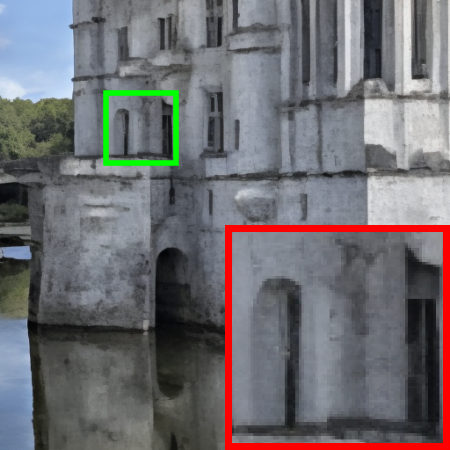}
&\includegraphics[width=0.14\textwidth]{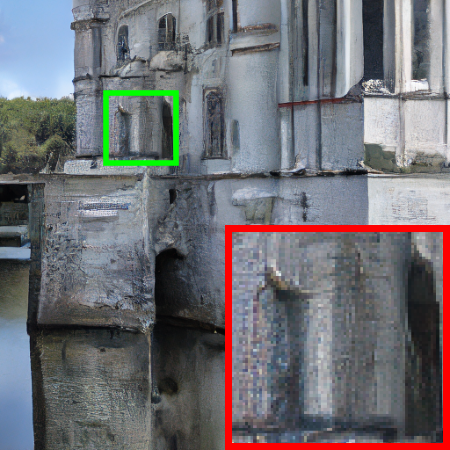}
&\includegraphics[width=0.14\textwidth]{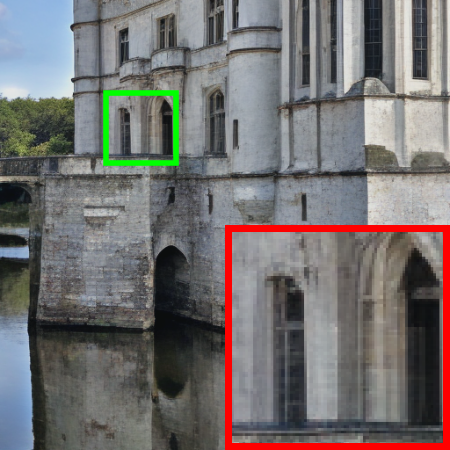}
&\includegraphics[width=0.14\textwidth]{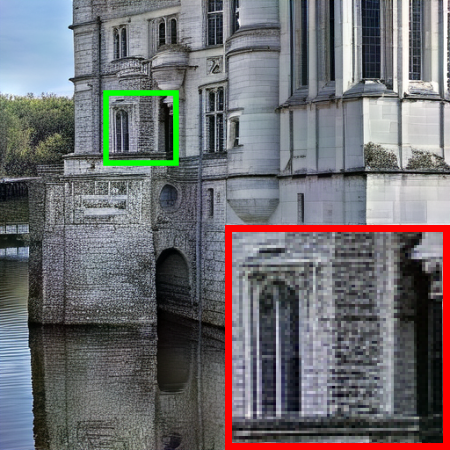}
&\includegraphics[width=0.14\textwidth]{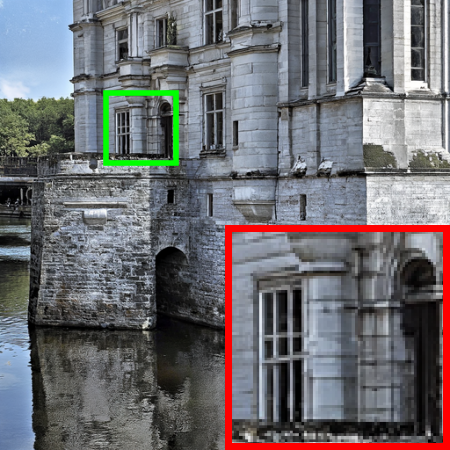}\\
\end{tabular}}

\vspace{-2pt}
% ---------- 中间整行图（与上表同宽，天然对齐） ----------
\includegraphics[width=\linewidth]{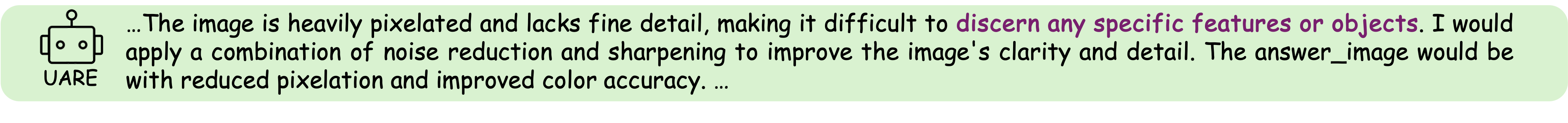}
\vspace{-15pt}

\end{minipage}%
} % end resizebox
\vspace{-10pt}
\caption{\textbf{Visual comparison of super-resolution} on images named ``Canon\_047" from RealSR (top) and ``0000065" from DIV2K-Val (bottom). Our UARE accurately understands both image content and degradations, achieving superior visual quality.}
\label{fig:sr_comparison}
\vspace{-10pt}
\end{figure*}

\subsection{Two-Stage Training Framework}
\label{sec:3.4}

As shown in 
Fig.~\ref{fig:framework}, we adopt a two-stage training framework, using a dynamic mixture of the curated data described above. Further illustrations are given below.

\noindent \textbf{Stage 1: Progressive easy-to-hard restoration training.} In this stage, we train only the restoration expert and keep all other components frozen. Using restoration data with simple instructions, UARE first learns to handle diverse degradations. As shown in Fig.~\ref{fig:framework}, the output is the restored image only, without text. Notably, we adopt a progressive learning strategy: the model is first trained on single degradations, then on multiple degradations, and finally on high-order degradations. This schedule guides the base model from simple to complex cases and steadily translates its generation capacity into strong restoration ability across degradation types and levels. This stage uses 9.6B image tokens for learning single degradations, 19.2B for multiple degradations, and 1.3B for high-order degradations.

\noindent \textbf{Stage 2: Unified fine-tuning of IQA and restoration.} After stage 1 stabilizes the core restoration ability, stage 2 jointly fine-tunes IQA and restoration. As shown in Fig.~\ref{fig:framework}, we update all model parameters except those of the VAE and the text tokenizer, using both IQA data and the interleaved text-image data. The corpus for this stage contains 0.4B IQA text tokens and 4.6B image tokens.

\subsection{Training Objectives}
\label{sec:3.5}
Both IQA and restoration training samples contain two parts, “condition” and ``response''. “Condition” refers to the task prompt, that is, the input text instruction and input image(s).  ``Response'' refers to the corresponding task targets. Denote a sample by $\mathbf{x} = (\mathbf{x}_\text{con}, \mathbf{x}_\text{res})$, with total sequence length as $l$, condition length as $l_\text{con}$ and response length as $l_\text{res}$. Let $\theta$ be the set of all trainable parameters in UARE.

\noindent \textbf{Autoregression objective.} For text prediction in IQA and IQA-guided restoration, where $\mathbf{x}_\text{con}$ with length $l_\text{con}$ contains the input instruction and image, and $\mathbf{x}_\text{res}$ contains only text tokens, UARE is trained by maximum likelihood, as:
\begin{equation}
\scalebox{0.87}{$
\mathcal{L}_{AR}(\theta) = -\mathbb{E}_{\mathbf{x}\sim \mathcal{D}_\text{IQA} }\left[\sum_{i=l_\text{con}}^{l-1} \text{log} ~ \text{P}_{\theta}(\mathbf{x}_{i+1}|\mathbf{x}_1, ... , \mathbf{x}_i) \right]$,}
\end{equation}
where $\mathcal{D}_\text{IQA}$ denotes the IQA data. $i$ is the token index starting from $l_\text{con}$. The loss is applied only to tokens in $\mathbf{x}_\text{res}$.

\noindent \textbf{Rectified flow objective.} For image restoration, $\mathbf{x}_\text{con}$ contains the input instruction and LQ image. For IQA-guided restoration, $\mathbf{x}_\text{con}$ also includes the generated quality analysis. The response $\mathbf{x}_\text{res}$ refers to the restored image. UARE is trained with the rectified flow~\cite{liu2022flow,lipman2022flow,esser2024scaling} objective, as:
\begin{equation}
\scalebox{0.85}{$
\mathcal{L}_{RF}(\theta) = \mathbb{E}_{\mathbf{x}\sim \mathcal{D}_\text{res},\mathbf{z}_0\sim\mathcal{N}(0,\mathbf{I})} 
 \left[\|v_{\theta}(\mathbf{z}_t, t | \mathbf{x}_\text{con})-(\mathbf{x}_\text{res}-\mathbf{z}_0)\|^2 \right]$,}
\end{equation}
where $\mathbf{z}_t=t\mathbf{x}_\text{res}+(1-t) \mathbf{z}_0$, $v_\theta$ denotes the velocity neural network with parameter $\theta$, $\mathcal{D}_\text{res}$ denotes the restoration data, and $t$ is the sampled diffusion timestep.

\noindent \textbf{Overall loss.} In 
stage 1, we use only the restoration loss: $\mathcal{L}_{s1}=\mathcal{L}_{RF}$. In stage 2, we use both losses with a trade-off: $\mathcal{L}_{s2}=\mathcal{L}_{RF}+\lambda\mathcal{L}_{AR}$. We set $\lambda=0.25$ in all experiments.
% During stage 1, UARE outputs only the restored image. Hence, the overall loss of stage 1 is $\mathcal{L}_{s1}(\theta_{1}) = \mathcal{L}_{RF}$, where $\theta_1$ contains only the parameters of the restoration expert. During the unified training in stage 2, the overall loss is $\mathcal{L}_{s2}(\theta_2) = \mathcal{L}_{RF} + \lambda\mathcal{L}_{AR}$, where $\lambda$ is a trade-off parameter, and $\theta_2$ includes the parameters of understanding expert, the restoration expert and understanding visual encoder.

\begin{table*}[t]
\centering
\caption{\textbf{Quantitative comparison} on nine multi-degradation subsets of FoundIR. For each method, the first row lists PSNR/LPIPS and the second row lists NIQE/MANIQA. Note that B., N., J., L., and H. denote Blur, Noise, JPEG compression, Low-light, and Haze, respectively.}
\vspace{-10pt}
\resizebox{\linewidth}{!}{
\renewcommand{\arraystretch}{1.3}
\begin{tabular}{l|ccccccccc}
\shline
Method & B.+N. & B.+J. & L.+H. & L.+B. & L.+N. & L.+J. & L.+B.+N. & L.+B.+J. & L.+N.+J. \\
\hline \hline

\multirow{2}{*}{Restormer~\cite{Restormer}} 
 & 22.70 / 0.3521 & 22.86 / 0.3397 & 11.33 / 0.7853 & 11.22 / 0.5734 & 8.77 / 0.7081 & 9.98 / 0.5076 & 9.54 / 0.5680 & 9.43 / 0.5432 & 10.43 / 0.3227 \\
 & 6.8794 / 0.1268 & 6.1987 / 0.1567 & 7.3745 / 0.2218 & 7.5792 / 0.1801 & 6.4635 / 0.2190 & 5.145 / 0.2466 & 7.3907 / 0.1621 & 6.6094 / 0.1788 & 5.6162 / 0.2560 \\
\hline

\multirow{2}{*}{PromptIR~\cite{PromptIR}} 
 & 22.80 / 0.3460 & \third{23.00} / 0.3677 & 16.49 / 0.6541 & \third{20.31} / 0.4606 & 11.23 / 0.6908 & 21.48 / 0.3645 & \secondbest{22.31} / 0.3918 & 12.57 / 0.5162 & 23.09 / 0.2091 \\
 & 7.7153 / 0.1354 & 7.8357 / 0.1718 & 6.8386 / 0.2391 & 8.4172 / 0.1615 & 6.9780 / 0.2286 & 5.4345 / 0.2874 & 7.3631 / 0.1339 & 8.2483 / 0.1914 & 5.8646 / 0.2915 \\
\hline

\multirow{2}{*}{DiffIR~\cite{DiffIR}} 
 & 22.78 / 0.3318 & 22.98 / 0.3276 & 20.03 / 0.2843 & \secondbest{21.94} / 0.3631 & 14.86 / 0.6158 & \secondbest{26.21} / \third{0.1385} & \third{21.74} / 0.3558 & \third{21.54} / 0.3305 & \secondbest{32.14} / \best{0.0532} \\
 & 7.0749 / 0.1315 & 7.0271 / 0.1681 & \third{3.5989} / 0.2751 & 6.9691 / 0.1508 & \secondbest{3.3768} / 0.2127 & \third{3.5363} / 0.3428 & 6.3189 / 0.1325 & 6.4128 / 0.1601 & \third{4.6687} / 0.3000 \\
\hline

\multirow{2}{*}{DiffUIR~\cite{DiffUIR}} 
 & \best{26.40} / \secondbest{0.1871} & 22.64 / 0.2459 & \secondbest{20.50} / \secondbest{0.2586} & 19.28 / \third{0.2934} & 14.39 / \third{0.4614} & 20.42 / \secondbest{0.1256} & 19.35 / \third{0.2380} & 19.52 / \third{0.2950} & 21.68 / 0.0800 \\
 & 5.9532 / \third{0.2660} & 6.2941 / 0.2671 & \secondbest{3.5501} / \secondbest{0.3081} & \third{6.4805} / \secondbest{0.2470} & 4.8987 / 0.2812 & 3.9225 / \third{0.3967} & \third{5.9545} / \secondbest{0.2469} & 6.7564 / \third{0.2571} & 5.0353 / \secondbest{0.3358} \\
\hline

\multirow{2}{*}{SUPIR~\cite{SUPIR}} 
 & 21.58 / 0.2533 & 21.36 / 0.3335 & 10.34 / 0.6731 & 10.75 / 0.5520 & 8.39 / 0.7636 & 8.22 / 0.4148 & 9.47 / 0.5456 & 9.14 / 0.5174 & 8.72 / 0.2817 \\
 & \best{4.1727} / \best{0.3330} & \secondbest{4.9403} / \secondbest{0.3053} & 4.9167 / 0.2753 & 7.8850 / 0.1815 & 7.5038 / 0.1973 & 4.3194 / 0.3418 & 7.3296 / 0.1594 & 7.8990 / 0.2078 & 5.0521 / 0.3065 \\
\hline

\multirow{2}{*}{InstructIR~\cite{InstructIR}} 
 & 21.52 / 0.2455 & 21.40 / \third{0.2067} & 13.71 / 0.5406 & 17.59 / 0.3900 & \secondbest{16.80} / 0.5926 & 18.69 / 0.2008 & 12.61 / 0.4020 & 17.55 / 0.3447 & 19.17 / 0.0929 \\
 & \third{5.8400} / 0.1768 & \third{6.0394} / 0.2511 & 4.0786 / 0.2880 & 6.7789 / 0.1558 & \best{3.0146} / \third{0.2846} & \secondbest{3.4827} / 0.3616 & 6.1334 / 0.1571 & \secondbest{6.2047} / 0.1709 & \best{4.0862} / 0.3191 \\
\hline

\multirow{2}{*}{AutoDIR~\cite{AutoDIR}} 
 & 21.74 / 0.3092 & 22.14 / 0.3098 & 14.77 / 0.6757 & 19.58 / 0.4300 & \best{17.73} / 0.5235 & 16.48 / 0.4028 & 18.37 / 0.3971 & 17.19 / 0.3706 & 18.89 / 0.2056 \\
 & 6.9000 / 0.1541 & 7.0662 / 0.1847 & 6.9955 / 0.1964 & 7.6655 / 0.1691 & 6.6220 / 0.2198 & 5.8992 / 0.2803 & 7.0586 / 0.1662 & 7.2479 / 0.2006 & 5.6556 / 0.2766 \\
\hline

\multirow{2}{*}{FoundIR~\cite{li2024foundir}} 
 & \third{23.81} / \third{0.2387} & \best{30.60} / \secondbest{0.1551} & \best{23.86} / \third{0.2774} & \best{25.13} / \secondbest{0.2282} & \third{16.47} / \secondbest{0.4054} & \best{30.32} / \best{0.0960} & \best{23.60} / \secondbest{0.2231} & \secondbest{22.79} / \secondbest{0.2183} & \best{34.27} / \secondbest{0.0559} \\
 & 6.1397 / 0.2282 & 6.4299 / \third{0.2851} & 4.1639 / \third{0.3033} & \secondbest{6.2341}/\third{0.2398} & 5.5928 / \secondbest{0.3285} & 3.8810 / \best{0.4067} & \secondbest{5.7256} / \third{0.2452} & \third{6.3009} / \secondbest{0.2629} & 4.9494 / \best{0.3360} \\
\hline

\multirow{2}{*}{\textbf{UARE (Ours)}} 
 & \secondbest{25.16} / \best{0.1573} & \secondbest{29.55} / \best{0.0891} & \third{20.10} / \best{0.2491} & 20.08 / \best{0.1878} & 15.08 / \best{0.3875} & \third{23.01} / 0.1540 & 20.20 / \best{0.1777} & \best{24.09} / \best{0.1225} & \third{29.30} / \third{0.0659} \\
 & \secondbest{4.8956} / \secondbest{0.2970} & \best{4.9061} / \best{0.3446} & \best{3.2736} / \best{0.3765} & \best{4.5594} / \best{0.3245} & \third{4.5500} / \best{0.3738} & \best{3.2550} / \secondbest{0.3999} & \best{4.6937} / \best{0.3169} & \best{5.0266} / \best{0.3379} & \secondbest{4.5995} / \third{0.3353} \\
\shline
\end{tabular}}
\label{tab:foundir_md}
\vspace{-6pt}
\end{table*}

\begin{figure*}[!t]
\setlength{\tabcolsep}{0.5pt}
\centering
\footnotesize
\resizebox{\linewidth}{!}{%
\begin{minipage}{\linewidth}

% ---------- 第一行小图 ----------
\resizebox{\linewidth}{!}{\begin{tabular}{@{}cccccccc@{}}
Input & DiffIR \cite{DiffIR} & DiffUIR \cite{DiffUIR} & SUPIR \cite{SUPIR} & InstructIR \cite{InstructIR} & FoundIR~\cite{li2024foundir} & \textbf{UARE (Ours)} & GT\\
\includegraphics[width=0.16\textwidth]{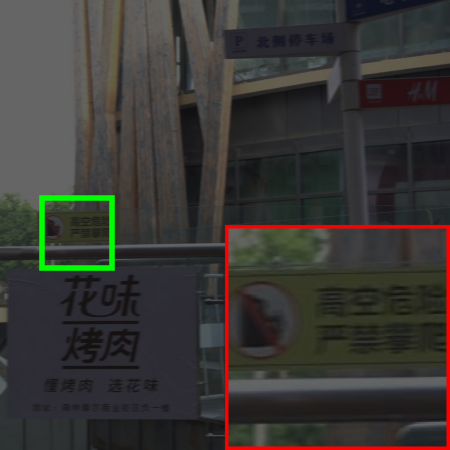}
&\includegraphics[width=0.16\textwidth]{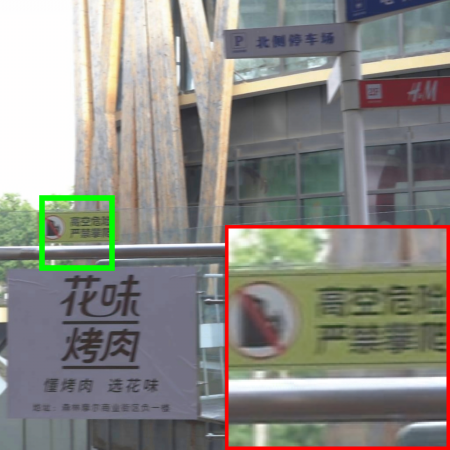}
&\includegraphics[width=0.16\textwidth]{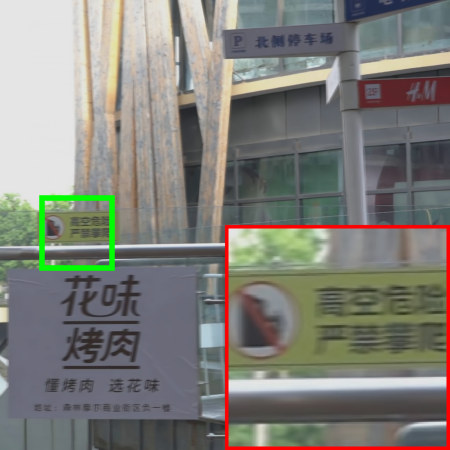}
&\includegraphics[width=0.16\textwidth]{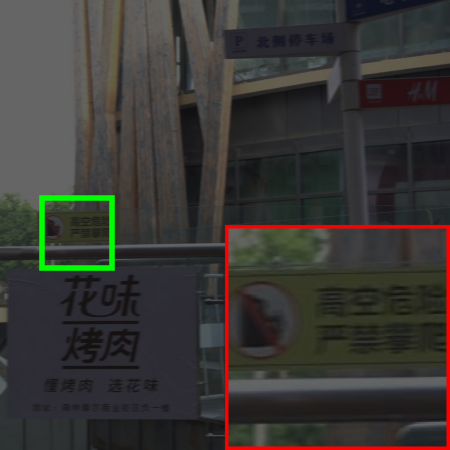}
&\includegraphics[width=0.16\textwidth]{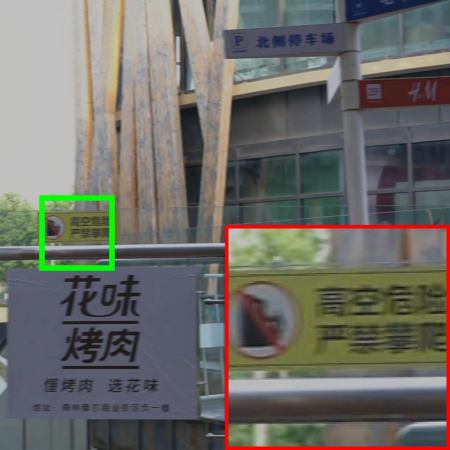}
&\includegraphics[width=0.16\textwidth]{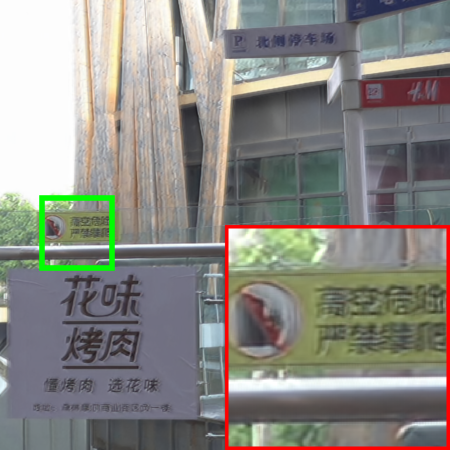}
&\includegraphics[width=0.16\textwidth]{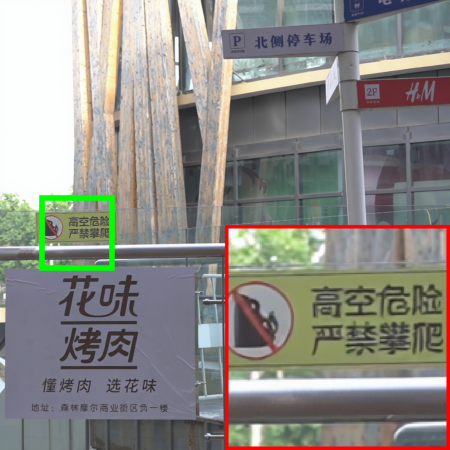}
&\includegraphics[width=0.16\textwidth]{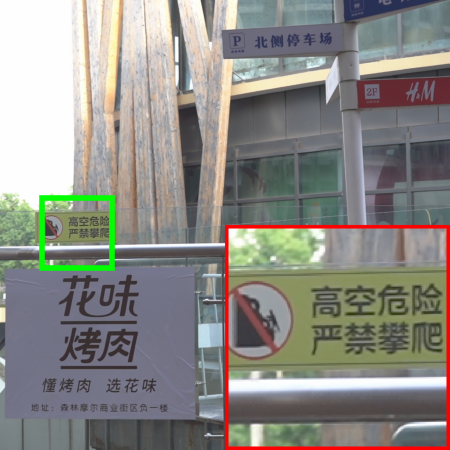}\\
\end{tabular}}

% \vspace{-2pt}

\vspace{-2pt}
% ---------- 中间整行图（与上表同宽，天然对齐） ----------
\includegraphics[width=\linewidth]{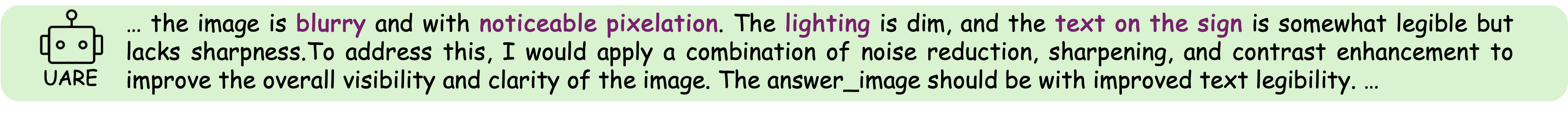}
\vspace{-15pt}

% ---------- 第二行小图 ----------
\resizebox{\linewidth}{!}{\begin{tabular}{@{}cccccccc@{}}
Input & DiffIR \cite{DiffIR} & DiffUIR \cite{DiffUIR} & SUPIR \cite{SUPIR} & InstructIR \cite{InstructIR} & FoundIR~\cite{li2024foundir} & \textbf{UARE (Ours)} & GT\\
\includegraphics[width=0.16\textwidth]{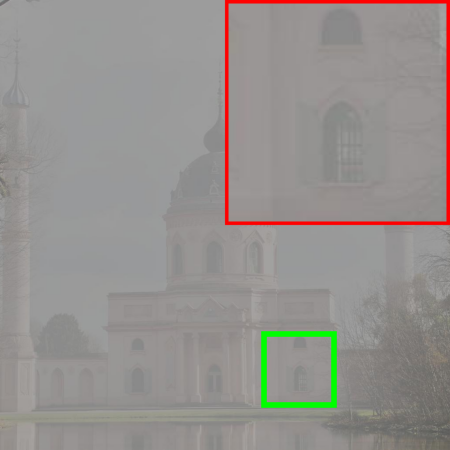}
&\includegraphics[width=0.16\textwidth]{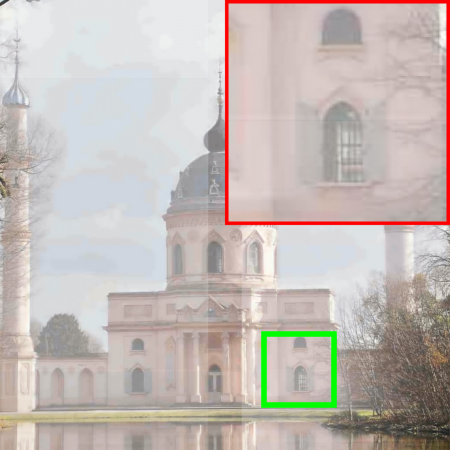}
&\includegraphics[width=0.16\textwidth]{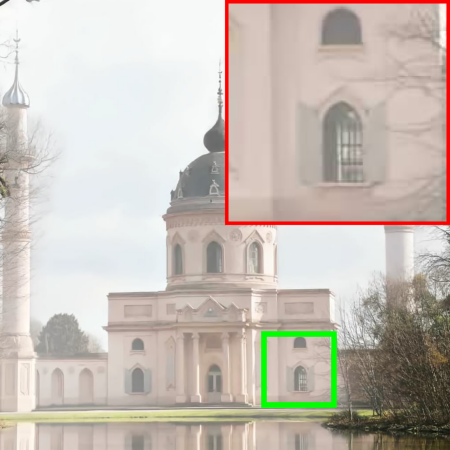}
&\includegraphics[width=0.16\textwidth]{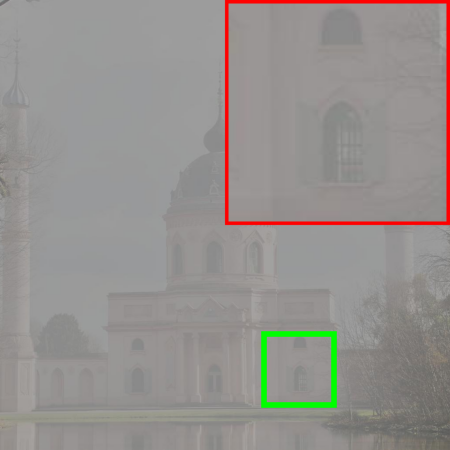}
&\includegraphics[width=0.16\textwidth]{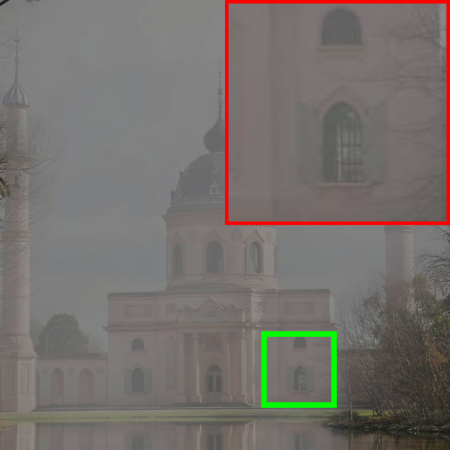}
&\includegraphics[width=0.16\textwidth]{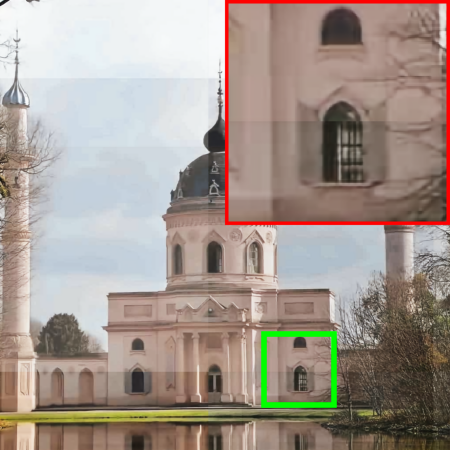}
&\includegraphics[width=0.16\textwidth]{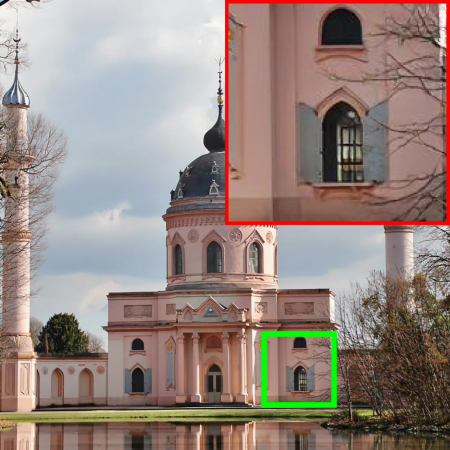}
&\includegraphics[width=0.16\textwidth]{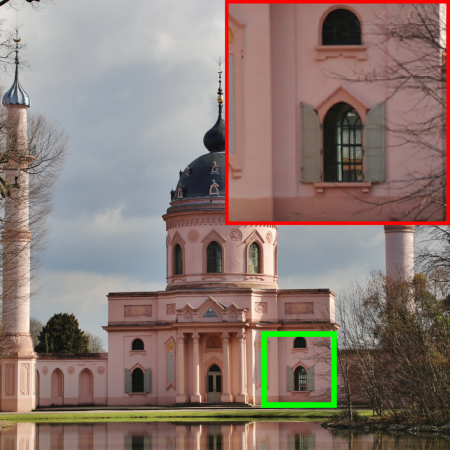}\\
\end{tabular}}

\vspace{-2pt}
% ---------- 中间整行图（与上表同宽，天然对齐） ----------
\includegraphics[width=\linewidth]{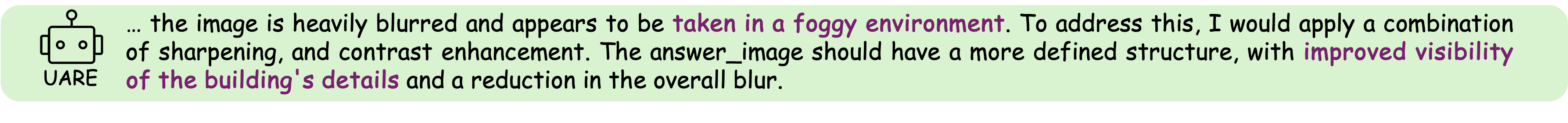}
\vspace{-15pt}

% \vspace{-2pt}

\end{minipage}%
} % end resizebox
\vspace{-10pt}
\caption{\textbf{Visual comparison} on images named ``1426" with low-light, blur, and noise (top),  and ``0525'' with haze (bottom) from FoundIR.}
\label{fig:foundir_comparison}
\vspace{-12pt}
\end{figure*}

\section{Experiments}
\subsection{Experimental Setup}
\textbf{Implementation details.}  We adopt Bagel~\cite{deng2025bagel} as our base model, whose understanding visual encoder uses a ViT architecture~\cite{dehghani2023patch}, and whose VAE is from FLUX~\cite{flux2024}. The training data, framework, and objective are detailed in Secs.~\ref{sec:3.3},~\ref{sec:3.4}, and~\ref{sec:3.5}, respectively. To enable classifier-free guidance~\cite{ho2022classifier}, we randomly drop text, ViT, and clean-VAE tokens with probability 0.1 each. 
% $\lambda$ in $\mathcal{L}_{s2}$ is set to $0.25$ to balance to two losses. 
We train UARE for 10K, 20K, and 1.5K steps in stage 1 for single, multiple, and high-order degradations, respectively, and 10K steps for stage 2. The learning rate is fixed at 2e-5, with a 500-step warmup.  In both stages, we use the AdamW~\cite{loshchilov2017decoupled} optimizer on 64 NVIDIA H20 (96GB) GPUs.  The whole training process takes about one week. More details of our training are provided in the \textcolor{blue}{\textbf{\textit{supplementary materials}}}.

\subsection{Super Resolution}
% \noindent\textbf{Test datasets.}
% \noindent\textbf{Compared methods.}
% \noindent\textbf{Evaluation metrics.}
\noindent\textbf{Settings.}
Following ~\cite{wu2024one,wu2024seesr,zhang2024degradation}, we evaluate UARE and compare it with prior methods on centor-cropped RealSR~\cite{cai2019toward}, DRealSR~\cite{wei2020component}, and DIV2K~\cite{agustsson2017ntire} with $512 \times 512$ resolution, performing $4\times$ SR. We compare UARE against eight diffusion-based methods:  StableSR~\cite{wang2024exploiting}, DiffBIR~\cite{lin2024diffbir}, SeeSR~\cite{wu2024seesr}, PASD~\cite{yang2024pixel}, ResShift~\cite{yue2024resshift}, SinSR~\cite{wang2024sinsr}, OSEDiff~\cite{wu2024one}, S3Diff~\cite{zhang2024degradation}, and the autoregressive method PURE~\cite{wei2025perceive}. We adopt both full- and no-reference metrics for evaluation. For reference-based fidelity, we use PSNR and SSIM~\cite{wang2004image}, calculated on the Y channel in YCbCr space. For reference-based perceptual quality, we apply LPIPS and DISTS. We also utilize NIQE~\cite{zhang2015feature}, LIQE~\cite{zhang2023blind}, MUSIQ~\cite{ke2021musiq}, MANIQA~\cite{yang2022maniqa}, and TOPIQ~\cite{chen2024topiq}.

\noindent \textbf{Results.} Tab.~\ref{tab:SR} presents that our UARE achieves strong results across
multiple metrics. Firstly, it ranks first in MUSIQ and TOPIQ across all datasets, and top-3 in MANIQA and LIQE. Secondly, it attains PSNR comparable to PURE while achieving better SSIM, LPIPS, and DISTS. Qualitative results are shown in Fig.~\ref{fig:sr_comparison}. UARE accompanies restoration with fine-grained, language-aligned quality analysis, correctly diagnosing issues (e.g., “barely legible text,” “heavy pixelation”) and suggesting appropriate enhancements. Moreover, UARE restores clearer text and well-formed window/brick patterns with markedly fewer artifacts. More visual results and a user study are provided in the \textcolor{blue}{\textbf{\textit{supplementary materials}}}.

\begin{table}[t]
\renewcommand{\arraystretch}{1.2}
\centering
\caption{
    \textbf{PLCC / SRCC comparison} on the image quality assessment task between our UARE and other competitive methods.
}
\vspace{-10pt}
\label{tab:score}
\resizebox{1.\linewidth}{!}{
\begin{tabular}{c|rrrrr}
\shline
Methods & SPAQ & KADID & LiveW & AGIQA & CSIQ \\
\hline \hline
NIQE~\cite{mittal2012making} & 
0.679 &
0.468 &
0.493 &
0.560 &
0.718  \\
(SPL 2012) & 
/\hspace{1pt}0.664 &
/\hspace{1pt}0.405 &
/\hspace{1pt}0.449 &
/\hspace{1pt}0.533 &
/\hspace{1pt}0.628  \\
BRISQUE~\cite{mittal2012no} & 
0.490 & 
0.429 & 
0.361 & 
0.541 & 
0.740  \\
(TIP 2012) & 
/\hspace{1pt}0.406 & 
/\hspace{1pt}0.356 & 
/\hspace{1pt}0.313 & 
/\hspace{1pt}0.497 & 
/\hspace{1pt}0.556  \\
MUSIQ~\cite{ke2021musiq} & 
0.868 & 
0.575 & 
0.789 & 
0.722 & 
0.771  \\
(ICCV 2021) & 
/\hspace{1pt}0.863 & 
/\hspace{1pt}0.556 & 
/\hspace{1pt}0.830 & 
/\hspace{1pt}0.630 & 
/\hspace{1pt}0.710  \\
CLIP-IQA+~\cite{wang2024exploiting} & 
0.866 & 
0.653 & 
0.832 & 
0.736 & 
0.772  \\
(AAAI 2023) & 
/\hspace{1pt}0.864 & 
/\hspace{1pt}0.654 & 
/\hspace{1pt}0.805 & 
/\hspace{1pt}0.685 & 
/\hspace{1pt}0.719  \\
ManIQA~\cite{yang2022maniqa} & 
0.768 & 
0.499 & 
0.849 & 
0.723 & 
0.623  \\
(CVPR 2022) & 
/\hspace{1pt}0.758 & 
/\hspace{1pt}0.465 & 
/\hspace{1pt}0.832 & 
/\hspace{1pt}0.636 & 
/\hspace{1pt}0.627  \\
Q-Align~\cite{wu2023QAlign} & 
0.886 & 
0.674 & 
0.853 & 
0.772 & 
0.671  \\
(ICML 2024) & 
/\hspace{1pt}0.887 & 
/\hspace{1pt}0.684 & 
/\hspace{1pt}\secondbest{0.860} & 
/\hspace{1pt}\secondbest{0.735} & 
/\hspace{1pt}0.737  \\
DeQA~\cite{you2025teaching} & 
0.895 & 
0.694 & 
\best{0.892} & 
\best{0.809} & 
\secondbest{0.787}  \\
(CVPR 2025) & 
/\hspace{1pt}0.896 & 
/\hspace{1pt}0.687 & 
/\hspace{1pt}\best{0.879} & 
/\hspace{1pt}0.729 & 
/\hspace{1pt}\secondbest{0.744} \\
Q-Insight~\cite{li2025qinsight} & 
\best{0.913} & 
\secondbest{0.757} & 
\secondbest{0.867} & 
\secondbest{0.805} & 
0.768  \\
(NeurIPS 2025) & 
/\hspace{1pt}\best{0.907} & 
/\hspace{1pt}\secondbest{0.765} & 
/\hspace{1pt}0.830 & 
/\hspace{1pt}\best{0.757} & 
/\hspace{1pt}0.740  \\
\textbf{UARE} & 
\secondbest{0.902} & 
\best{0.878} & 
0.855 & 
0.752 & 
\best{0.930}  \\
\textbf{(Ours)} & 
/\hspace{1pt}\secondbest{0.898} & 
/\hspace{1pt}\best{0.873} & 
/\hspace{1pt}0.814 & 
/\hspace{1pt}0.667 & 
/\hspace{1pt}\best{0.915}  \\
\shline
\end{tabular}}
\vspace{-12pt}
\end{table}

\subsection{Mix-Degraded Image Restoration}
\noindent \textbf{Settings.} We evaluate UARE and compare it with other methods on the center-cropped FoundIR test set~\cite{li2024foundir} with $1024\times 1024$ resolution. We compare UARE against eight state-of-the-art restoration methods: Restormer~\cite{Restormer}, PromptIR~\cite{PromptIR}, DiffIR~\cite{DiffIR}, DiffUIR~\cite{DiffUIR}, SUPIR~\cite{SUPIR}, InstructIR~\cite{InstructIR}, AutoDIR~\cite{AutoDIR}, and FoundIR~\cite{li2024foundir}. We adopt PSNR, LPIPS, NIQE, and MANIQA as evaluation metrics. 

\noindent\textbf{Results.} Tab.~\ref{tab:foundir_md} shows that our UARE achieves top-3 performance on nearly all metrics across the mixed-degradation subsets of FoundIR. Fig.~\ref{fig:foundir_comparison} shows that UARE precisely identifies degradation types and scene content in LQ inputs, removes blur, low-light, noise, and haze, and reconstructs vivid, realistic details in challenging regions (e.g., fine facade structures and small text). Results on single-degradation subsets are given in \textcolor{blue}{\textbf{\textit{supplementary materials}}}.

\subsection{Image Quality Assessment}
We compare UARE with handcrafted methods NIQE~\cite{mittal2012making} and BRISQUE~\cite{mittal2012no}, non-LLM approaches MUSIQ~\cite{ke2021musiq}, CLIP-IQA+~\cite{wang2024exploiting}, MANIQA~\cite{yang2022maniqa}, and LLM-based methods Q-Align~\cite{wu2023QAlign}, DeQA-Score~\cite{you2025teaching}, and Q-Insight~\cite{li2025qinsight}. Following Q-Insight~\cite{li2025qinsight}, we evaluate on SPAQ~\cite{fang2020perceptual}, KADID~\cite{lin2019kadid}, LIVEW~\cite{ghadiyaram2015live}, AGIQA~\cite{li2023agiqa}, and CSIQ~\cite{larson2010most}. PLCC and SRCC are adopted for image score regression. As shown in Tab.~\ref{tab:score}, UARE significantly surpasses the current state-of-the-art IQA method Q-Insight on KADID and CSIQ, while achieving comparable performance on the remaining benchmarks. It demonstrates that our method not only achieves superior reconstruction performance but also possesses strong visual quality understanding capability.

\subsection{Ablation Study}
\noindent \textbf{Effect of two-stage training.} In Tab.~\ref{tab:ablation-two-stage}, we show the results of pretrained Bagel and three variants after each progressive step in stage 1 on the RealSR dataset. We also train an ``all-in-one stage" variant by mixing all data in a single-stage training. Our progressive training framework balances fidelity and perceptual quality, achieving the best overall performance. Besides, comparing the “+ high-order deg.” variant with our final UARE confirms that introducing IQA data to jointly fine-tune the IQA and restoration experts yields a significant gain in perceptual quality.

\begin{table}[t]
\renewcommand{\arraystretch}{1.2}
\centering
\caption{\textbf{Ablation study} of the two-stage training framework.}
\vspace{-10pt}
\label{tab:ablation-two-stage}
\resizebox{1.0\linewidth}{!}{
\begin{tabular}{l|ccccc}
\shline
Methods & PSNR$\uparrow$ & LPIPS$\downarrow$ & LIQE$\uparrow$ & MUSIQ$\uparrow$ & MANIQA$\uparrow$ \\
\hline \hline
Pretrained Bagel & 21.00 & 0.5906 & 1.0990 & 29.66 & 0.1792 \\
+ single deg. & 22.26 & 0.4205 & 1.0080 & 28.10 & 0.2033 \\
+ multi deg. & 21.17 & 0.3989 & 1.0235 & 30.96 & 0.2152 \\
+ high-order deg. & \secondbest{22.74} & \best{0.2603} & \secondbest{3.5686} & \secondbest{60.21} & \secondbest{0.4082} \\
All-in-one stage & \best{23.65} & \secondbest{0.2662} & {2.8750} & {57.50} & {0.3760} \\
\textbf{UARE (Ours)} & 21.38 & {0.3095} & \best{4.0658} & \best{69.67} & \best{0.5260} \\
\shline
\end{tabular}}
\vspace{-12pt}
\end{table}

\begin{table}[t]
\renewcommand{\arraystretch}{1.2}
\centering
\caption{\textbf{Ablation study} of the IQA guidance in restoration.
}
\vspace{-10pt}
\label{tab:ablation-prompt}
\resizebox{1.0\linewidth}{!}{
\begin{tabular}{l|ccccc}
\shline
Methods & PSNR$\uparrow$ & LPIPS$\downarrow$ & LIQE$\uparrow$ & MUSIQ$\uparrow$ & MANIQA$\uparrow$ \\
\hline \hline
Simple prompt & \best{21.79} & \best{0.2772} & {3.6308} & \secondbest{66.52} & {0.4668} \\
Q-Insight prompt & {21.26} & {0.3175} & \secondbest{4.0498} & {64.43} & \secondbest{0.4817} \\
\textbf{UARE (Ours)} & \secondbest{21.38} & \secondbest{0.3095} & \best{4.0658} & \best{69.67} & \best{0.5260} \\
\shline
\end{tabular}}
\vspace{-12pt}
\end{table}

\noindent \textbf{Effect of the IQA guidance in restoration.} In Tab.~\ref{tab:ablation-prompt}, we study how different text prompts used at inference affect UARE’s restoration performance on RealSR. Specifically, the ``single prompt” setting uses the instruction ``enhance the mix-degraded image.” The ``Q-Insight prompt” first asks the Q-Insight model to produce a degradation analysis and enhancement suggestions, then feeds those suggestions as textual instructions together with the low-quality image into UARE. Our UARE follows an analysis-then-restore paradigm: it outputs a quality analysis first, and then the enhanced image. UARE achieves the best perceptual metrics while maintaining comparable fidelity. Compared with the single-prompt setting, UARE parses and exploits richer IQA textual information. Compared with the Q-Insight-prompt setting, UARE aligns its own IQA and restoration capabilities more effectively, thereby turning IQA insights into concrete gains in restoration.

\section{Conclusion}
In this paper, we propose UARE, a novel unified vision-language model for image quality assessment, restoration, and enhancement. We introduce a two-stage training framework. The first stage enables UARE to handle multiple degradations in a progressive way, while the second stage aligns IQA signals with restoration objectives. We find that IQA can boost the restoration performance of UARE through multi-task co-training. Extensive experiments across diverse tasks demonstrate the effectiveness of UARE. Looking ahead, UARE can extend its capabilities to a wide range of tasks, such as video quality understanding and restoration. As a unified model, UARE has the potential to revolutionize IQA and restoration/enhancement, providing a unified solution that can transform how image quality is evaluated, improved, and applied across various fields.
% \maketitlesupplementary
\setcounter{section}{0}
\renewcommand\thesection{\Alph{section}}
\setcounter{table}{0}
\renewcommand\thetable{\Alph{table}}
\makeatletter
  \renewcommand\theHtable{\Alph{table}}   
\makeatother
\setcounter{figure}{0}
\renewcommand\thefigure{\Alph{figure}}
\makeatletter
  \renewcommand\theHfigure{\Alph{figure}}
\makeatother
\counterwithin{figure}{section}
\counterwithin{table}{section}

\section*{Supplementary Material}
Our main paper presents the core ideas, architecture, and experimental results of UARE, a unified vision–language model for image quality assessment, restoration, and enhancement. In this \textbf{\textit{\textcolor{blue}{supplementary material}}}, we provide additional information. Sec.~\ref{sec:A} gives implementation details, including representative data examples in Sec.~\ref{sec:A.1} and training configurations in Sec.~\ref{sec:A.2}. Sec.~\ref{sec:B} reports more comparison results, with real-world super resolution in Sec.~\ref{sec:B.1}, image restoration/enhancement in Sec.~\ref{sec:B.2}, and a user study in Sec.~\ref{sec:B.3}. Finally, Sec.~\ref{sec:C} offers further discussion of UARE and a detailed analysis of its limitations.

\section{Implementation Details}
\label{sec:A}

\begin{figure*}[!b]
\centering
\includegraphics[width=1.0\linewidth]{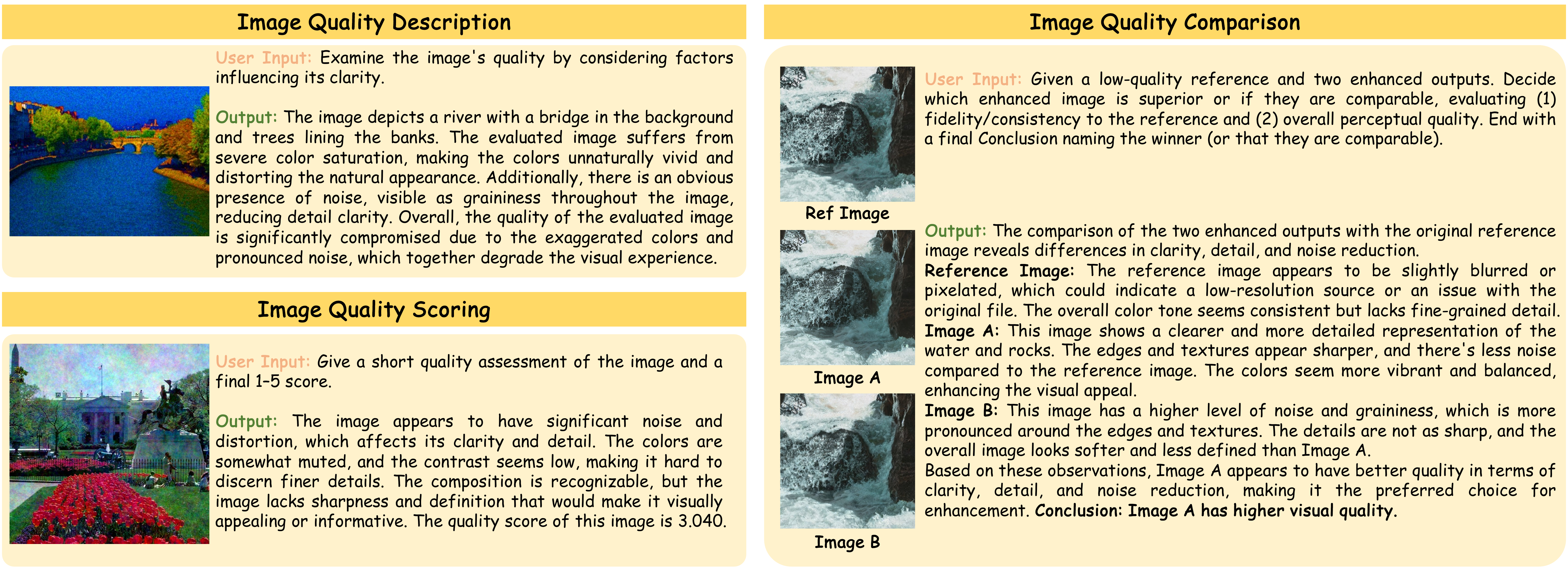}
\vspace{-14pt}
\caption{Data examples for IQA training in UARE, including quality description, image quality scoring, and image comparison.}
\label{fig:IQA_case}
\vspace{-6pt}
\end{figure*}

\subsection{Data Examples}
\label{sec:A.1}

\begin{figure*}[!t]
\centering
\includegraphics[width=1.0\linewidth]{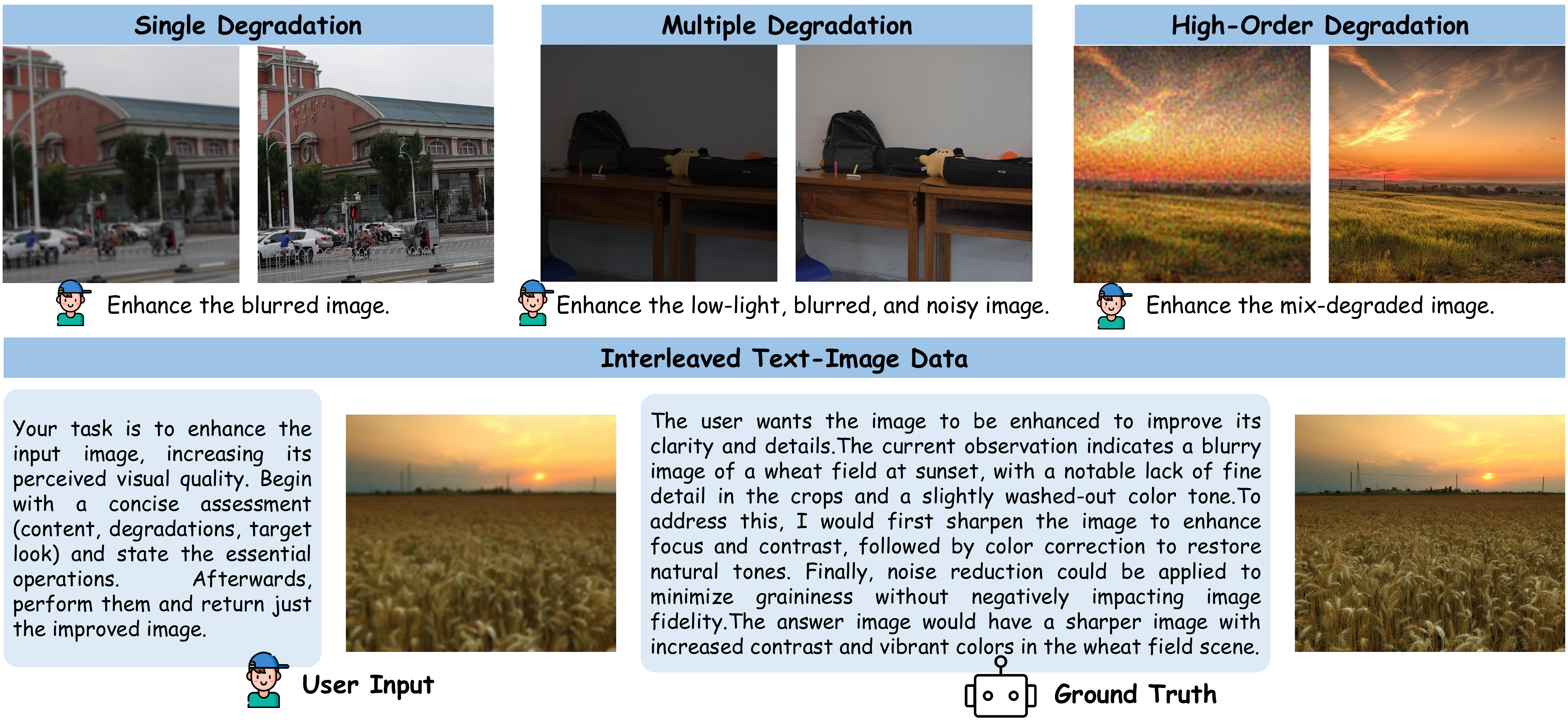}
\vspace{-14pt}
\caption{Data examples for restoration and enhancement training in UARE, covering single, multiple, and high-order degradations as well as interleaved text–image pairs.}
\label{fig:Res_case}
% \vspace{-6pt}
% \vspace{50pt}
\end{figure*}

We have detailed the data construction process in our main paper. Here, we provide data examples for training UARE in Fig.~\ref{fig:IQA_case} and Fig.~\ref{fig:Res_case} for IQA and restoration/enhancement, respectively. For IQA, our data include diverse instruction formats such as free-form quality description, scalar quality scoring, and reference-based pairwise comparison. For restoration and enhancement, we construct instructions for single, multiple, and high-order degradations, as well as interleaved text–image data where the model must first analyze the degradation and user intent and then plan the enhancement steps. These examples span a wide range of scenes (indoor/outdoor, day/night, natural and urban) and degradation types (blur, low light, noise, haze, and complex mixed artifacts), highlighting the richness and compositionality of our training corpus.

% \begin{figure*}[!t]
% \centering
% \includegraphics[width=\linewidth]{figs/IQA_case.pdf}
% \vspace{2mm} % 两张图之间留一点间距
% \includegraphics[width=\linewidth]{figs/Res_case.pdf}
% \vspace{-10pt}
% \caption{Data examples for training UARE. Top: IQA instructions and responses. Bottom: restoration/enhancement instructions and responses.}
% \label{fig:data_examples}
% \vspace{-6pt}
% \end{figure*}

% \clearpage

% \clearpage
% \clearpage
\subsection{Training Details}
\label{sec:A.2}
% \vspace{-5pt}
Table~\ref{tab:training_recipe} summarizes the full training recipe of UARE. For all stages, we use a constant learning rate of $2\times10^{-5}$, zero weight decay, gradient-norm clipping of 1.0, and AdamW ($\beta_1=0.9$,~$\beta_2=0.95$,~$\epsilon=1.0\times10^{-15}$) with EMA ratios of 0.990, 0.995, 0.995, and 0.995 for the single-degradation, multi-degradation, high-order degradation, and unified fine-tuning stages, respectively. The three stage-1 curricula are trained for 10K, 20K, and 1.5K steps with 250 warm-up steps, while the unified stage is trained for 10K steps with a 500-step warm-up. In the unified stage, we jointly optimize the IQA cross-entropy loss and the restoration MSE loss with a weight ratio of $0.25:1$. Restoration images are randomly resized such that the minimum short side and maximum long side fall in $(512, 1024)$, and IQA images are resized to $(378, 980)$. We apply a diffusion timestep shift of 4.0 for all diffusion-based restoration branches. The bottom part of Table A.1 lists the data sampling ratios: stage-1 curricula only sample their corresponding degradation type (single, multi, or high-order), whereas the unified stage mixes IQA, single-, multi-, and high-order degradation data and interleaved IQA–restoration pairs, forming a balanced curriculum for learning unified quality assessment and restoration.

\begin{table*}[t!]
% \small
\centering

\caption{\textbf{Training recipe of UARE.}}
\vspace{-10pt}
\setlength{\tabcolsep}{6pt}
\resizebox{0.9\linewidth}{!}{
\begin{tabular}{l|cccc}
\toprule
 & \textbf{single deg.} & \textbf{multi deg.} & \textbf{high-order deg.} & \textbf{Uni ft.} \\
\hline
\textbf{Hyperparameters} \\
Learning rate  & \multicolumn{4}{c}{$2\times10^{-5}$} \\
LR scheduler   &  \multicolumn{4}{c}{Constant} \\
Weight decay   & \multicolumn{4}{c}{0.0} \\
Gradient norm clip & \multicolumn{4}{c}{1.0} \\
Optimizer       & \multicolumn{4}{c}{AdamW ($\beta_1=0.9$, $\beta_2=0.95$, $\epsilon=1.0 \times 10^{-15}$)} \\
Loss weight (CE : MSE)  & - & - & - & 0.25 : 1 \\
Warm-up steps   & 250 & 250 & 250 & 500 \\
Training steps  & 10K & 20K & 1.5k & 10K \\
EMA ratio            & 0.990 & 0.995 & 0.995 & 0.995 \\
% Sequence length per rank (min, max) & (32K, 36K) & (32K, 36K) & (40K, 45K) & (40K, 45K) \\
Training seen tokens & 9.6B & 19.2B & 1.3B & 4.6B \\
% Max context window  & 16K & 16k & 40k & 40k \\
Res. resolution (min short side, max long side)      & \multicolumn{4}{c}{(512, 1024)} \\
IQA resolution (min short side, max long side)      & \multicolumn{4}{c}{(378, 980)} \\
Diffusion timestep shift & \multicolumn{4}{c}{4.0} \\
\midrule
\textbf{Data sampling ratio} \\
IQA                            & 0.0 & 0.0 & 0.0 & 0.25 \\
Single degradation          & 1.0 & 0.0 & 0.0 & 0.05 \\
Multiple degradation           & 0.0 & 1.0 & 0.0 & 0.1 \\
high-order degradation       & 0.0 & 0.0 & 1.0 & 0.2 \\
Interleaved IQA and restoration    & 0.0 & 0.0 & 0.0 & 0.4 \\
\bottomrule
\end{tabular}}
\label{tab:training_recipe}
\end{table*}

% \clearpage

\section{More Comparison Results}
\label{sec:B}
\subsection{Real-World Super Resolution}
\label{sec:B.1}

\begin{table*}[t]
\centering
\caption{Quantitative comparison of different methods on RealSR, DRealSR, and DIV2K. Throughout this paper, best, second-best, and third-best results are highlighted in \best{bold red}, \secondbest{underlined blue}, \third{italic green}. $\uparrow$ / $\downarrow$ indicate higher/lower is better.}
\vspace{-4pt}
\resizebox{\linewidth}{!}{
\renewcommand{\arraystretch}{1.15}
\begin{tabular}{l | l | c c c c c c c c c}
\shline
Test Dataset & Method & PSNR$\uparrow$ & SSIM$\uparrow$ & LPIPS$\downarrow$ & DISTS$\downarrow$ & NIQE$\downarrow$ & LIQE$\uparrow$ & MUSIQ$\uparrow$ & MANIQA$\uparrow$ & TOPIQ$\uparrow$ \\
\hline \hline
% ================= Set-A =================
\multirow{13}{*}{RealSR}
& Real-ESRGAN~\cite{Real-esrgan}   & 23.62 & \third{0.7185} & 0.2763 & 0.2063 & 5.7619        & 3.3163 & 59.87 & 0.3749 & 0.5097 \\
& FeMASR~\cite{chen2022real}        & 23.26 & 0.7030 & 0.2850 & 0.2254 & 5.7053        & 3.1587 & 58.05 & 0.3435 & 0.4848 \\
& SwinIR~\cite{SwinIR}        & 23.75 & \best{0.7250} & \best{0.2608} & \third{0.1981} & 5.6989        & 3.0798 & 58.95 & 0.3546 & 0.4816 \\
& InvSR~\cite{yue2025arbitrary}         & 22.90 & 0.6844 & \secondbest{0.2634} & \secondbest{0.1980} & 5.9996        & 3.7639 & 67.20 & 0.4270 & 0.5546 \\
& StableSR~\cite{wang2024exploiting}      & 23.73 & 0.6979 & 0.2792 & 0.2023 & 5.5914        & 3.0532 & 61.65 & 0.3826 & 0.5201 \\
& DiffBIR~\cite{lin2024diffbir}       & 23.20 & 0.6346 & 0.3350 & 0.2162 & \best{4.5879} & 3.5529 & 65.25 & 0.4620 & 0.6033 \\
& SeeSR~\cite{wu2024seesr}         & \secondbest{24.34} & \secondbest{0.7187} & 0.2754 & 0.2134 & 6.4146 & 3.3938 & 65.53 & \third{0.4856} & 0.6246 \\
& PASD~\cite{yang2024pixel}          & \best{24.50} & 0.7115 & \third{0.2716} & \best{0.1954} & 6.0067 & 2.8541 & 58.52 & 0.3831 & 0.4969 \\
& ResShift~\cite{yue2024resshift}      & \third{24.17} & 0.6528 & 0.4336 & 0.2812 & 8.6273 & 2.6610 & 53.38 & 0.3412 & 0.4210 \\
& SinSR~\cite{wang2024sinsr}         & 23.68 & 0.6649 & 0.3490 & 0.2445 & 6.5101 & 3.2255 & 61.03 & 0.4230 & 0.5383 \\
& OSEDiff~\cite{wu2024one}       & 23.07 & 0.6850 & 0.2941 & 0.2109 & 5.5054 & \best{4.0681} & \secondbest{68.95} & \secondbest{0.4876} & \secondbest{0.6441} \\
& S3Diff~\cite{zhang2024degradation}        & 23.16 & 0.6810 & 0.2748 & 0.1986 & \third{5.3003} & \third{4.0080} & \third{67.57} & 0.4677 & \third{0.6301} \\
& PURE~\cite{wei2025perceive}          & 21.31 & 0.5738 & 0.3859 & 0.2468 & 5.6419 & 3.7881 & 66.57 & 0.4829 & \third{0.6301} \\
& \textbf{UARE (Ours)}   & 21.38 & 0.6464 & 0.3095 & 0.2344 & \secondbest{5.2981} & \secondbest{4.0658} & \best{69.67} & \best{0.5260} & \best{0.6796} \\
\hline \hline 
% ================= Set-B =================
\multirow{13}{*}{DRealSR}
&Real-ESRGAN~\cite{Real-esrgan}  & 27.26 & \third{0.7745} & 0.2841 & \third{0.2085} & 6.6994 & 2.8595 & 53.43 & 0.3438 & 0.4559 \\
&FeMASR~\cite{chen2022real}       & 25.32 & 0.7221 & 0.3164 & 0.2241 & \secondbest{5.8831} & 2.9538 & 53.32 & 0.3169 & 0.4722 \\
&SwinIR~\cite{SwinIR}       & 27.01 & 0.7703 & \secondbest{0.2793} & \secondbest{0.2070} & 6.5370 & 2.8340 & 52.42 & 0.3310 & 0.4484 \\
&InvSR~\cite{yue2025arbitrary}        & 25.55 & 0.7087 & 0.3188 & 0.2192 & 6.0231 & \third{3.7525} & \third{64.25} & 0.4301 & 0.5726 \\
&StableSR~\cite{wang2024exploiting}     & \best{28.28} & \best{0.7981} & \best{0.2687} & \best{0.2026} & 7.2816 & 2.5068 & 51.62 & 0.3226 & 0.4355 \\
&DiffBIR~\cite{lin2024diffbir}      & 26.08 & 0.6578 & 0.4144 & 0.2564 & \best{4.4856} & 3.3993 & 61.81 & 0.4612 & 0.6084 \\
&SeeSR~\cite{wu2024seesr}  & \third{28.14} & \secondbest{0.7798} &\third{0.2832} & 0.2241 & 7.4833 & 2.7943 & 55.89 & 0.3976 & 0.5436 \\
&PASD~\cite{yang2024pixel} & \secondbest{28.18} & 0.7722 & 0.2970 & 0.2108 & 7.4421 & 2.6129 & 51.42 & 0.3595 & 0.4587 \\
&ResShift~\cite{yue2024resshift}     & 27.39 & 0.6907 & 0.4996 & 0.3077 & 9.1788 & 1.7905 & 40.58 & 0.2457 & 0.3414 \\
&SinSR~\cite{wang2024sinsr}        & 26.72 & 0.6933 & 0.4031 & 0.2624 & 6.8825 & 2.7781 & 53.36 & 0.3677 & 0.4959 \\
&OSEDiff~\cite{wu2024one}      & 25.60 & 0.7403 & 0.3088 & 0.2158 & 6.1544 & \secondbest{3.9797} & \secondbest{65.24} & \secondbest{0.4879} & \secondbest{0.6273} \\
&S3Diff~\cite{zhang2024degradation}       & 26.18 & 0.7197 & 0.3161 & 0.2099 & \third{5.9531} & 3.9255 & 63.34 & \third{0.4635} & \third{0.6181} \\
&PURE~\cite{wei2025perceive}         & 23.04 & 0.5718 & 0.4461 & 0.2674 & 6.3939 & 3.7390 & 60.68 & 0.4362 & 0.5888 \\
&\textbf{UARE (Ours)}  & 21.31 & 0.5736 & 0.4071 & 0.2613 & 6.4290 & \best{4.0445} & \best{67.71} & \best{0.5121} & \best{0.6652} \\
\hline \hline
% ================= Set-C =================
\multirow{13}{*}{DIV2K}
& Real-ESRGAN~\cite{Real-esrgan}   & \secondbest{19.41} & \secondbest{0.4901} & 0.4123 & 0.2586 & 4.5100 & 3.6731 & 61.63 & 0.3835 & 0.5449 \\
& FeMASR~\cite{chen2022real}        & 18.46 & 0.4339 & 0.4139 & 0.2382 & \third{4.1173} & 3.4794 & 61.01 & 0.3068 & 0.5151 \\
& SwinIR~\cite{SwinIR}        & 19.11 & \third{0.4772} & 0.4285 & 0.2647 & 4.7146 & 3.2109 & 58.22 & 0.3251 & 0.4844 \\
& InvSR~\cite{yue2025arbitrary}         & 18.93 & 0.4597 & 0.4182 & 0.2685 & 5.8936 & 3.5459 & 61.91 & 0.4066 & 0.5573 \\
& StableSR~\cite{wang2024exploiting}      & \best{19.85} & \best{0.4940} & 0.4796 & 0.2887 & 5.7479 & 1.8466 & 43.25 & 0.2181 & 0.3276 \\
& DiffBIR~\cite{lin2024diffbir}       & 18.94 & 0.4332 & 0.4009 & 0.2238 & \best{3.6594} & 3.8573 & 67.20 & 0.4574 & 0.6467 \\
& SeeSR~\cite{wu2024seesr}         & 19.11 & 0.4580 & \third{0.3769} & 0.2339 & 4.5817 & 3.7445 & 66.31 & \third{0.4686} & 0.6330 \\
& PASD~\cite{yang2024pixel}          & 18.98 & 0.4562 & 0.4293 & 0.2373 & 4.7846 & 3.6022 & 63.46 & 0.4025 & 0.5653 \\
& ResShift~\cite{yue2024resshift}      & \third{19.15} & 0.4311 & 0.4900 & 0.2808 & 7.4321 & 2.8862 & 56.02 & 0.3534 & 0.4662 \\
& SinSR~\cite{wang2024sinsr}         & 18.58 & 0.4059 & 0.4483 & 0.2455 & 6.0533 & 3.4629 & 64.12 & 0.4483 & 0.5997 \\
& OSEDiff~\cite{wu2024one}       & 18.86 & 0.4563 & \secondbest{0.3579} & \third{0.2209} & 4.1756 & 3.8877 & 67.83 & 0.4422 & 0.6269 \\
& S3Diff~\cite{zhang2024degradation}        & 18.76 & 0.4490 & \best{0.3299} & \best{0.1990} & 4.2026 & \secondbest{4.2692} & \third{69.31} & 0.4675 & \secondbest{0.6679} \\
& PURE~\cite{wei2025perceive}          & 16.71 & 0.3661 & 0.4449 & 0.2293 & 4.9545 & \best{4.2701} & \secondbest{70.06} &\best{0.5201} & \third{0.6621} \\
& \textbf{UARE (Ours)}   & 16.59 & 0.3857 & 0.4074 & \secondbest{0.2138} & \secondbest{3.7931} & \third{4.2627} & \best{70.45} & \secondbest{0.5028} & \best{0.6864} \\
\shline
\end{tabular}}
\label{tab:sr-all-supp}
\end{table*}

\begin{table}[t]
\centering
\caption{Quantitative comparison of different methods on RealSet80 without ground truth.}
\setlength{\tabcolsep}{5.2pt}
\resizebox{\linewidth}{!}{
\renewcommand{\arraystretch}{1.15}
\begin{tabular}{l | c c c c c}
\shline
Method & NIQE$\downarrow$ & LIQE$\uparrow$ & MUSIQ$\uparrow$ & MANIQA$\uparrow$ & TOPIQ$\uparrow$ \\
\hline \hline
BSRGAN~\cite{zhang2021designing} & 5.1655 & 3.8884 & 64.85 & 0.3941 & 0.5821 \\
StableSR~\cite{wang2024exploiting} & \best{4.0798} & 3.9074 & 67.67 & 0.4682 & 0.6440 \\
DiffBIR~\cite{lin2024diffbir} & 6.1069 & 4.1113 & 68.10 & \best{0.5527} & \secondbest{0.6736} \\
SeeSR~\cite{wu2024seesr} & 5.2244 & \best{4.3317} & \secondbest{69.70} & \third{0.5362} & \best{0.6887} \\
SinSR~\cite{wang2024sinsr} & 6.4250 & 3.6613 & 62.78 & 0.4483 & 0.5854 \\
OSEDiff~\cite{wu2024one} & \third{4.6362} & \third{4.2251} & 68.88 & 0.4995 & 0.6062 \\
PURE~\cite{wei2025perceive} & 5.3617 & \secondbest{4.2528} & \third{69.55} & 0.5215 & \third{0.6647} \\
\textbf{UARE (Ours)} & \secondbest{4.6044} & 4.1804 & \best{70.05} & \secondbest{0.5363} & 0.6446 \\
\shline
\end{tabular}}
\label{tab:realset80}
\end{table}

\noindent\textbf{More quantitative results.}
We further report more quantitative results in Tab.~\ref{tab:sr-all-supp} on RealSR~\cite{cai2019toward}, DRealSR~\cite{wei2020component} and DIV2K~\cite{agustsson2017ntire}. We compare UARE with \textbf{twelve} SR methods: Real-ESRGAN~\cite{Real-esrgan}, FeMASR~\cite{chen2022real}, SwinIR~\cite{SwinIR}, InvSR~\cite{yue2025arbitrary}, StableSR~\cite{wang2024exploiting}, DiffBIR~\cite{lin2024diffbir}, SeeSR~\cite{wu2024seesr}, PASD~\cite{yang2024pixel}, ResShift~\cite{yue2024resshift}, SinSR~\cite{wang2024sinsr}, OSEDiff~\cite{wu2024one}, S3Diff~\cite{zhang2024degradation}, and PURE~\cite{wei2025perceive}. The evaluation metrics follow the main paper: for reference-based fidelity, we report PSNR and SSIM~\cite{wang2004image} on the Y channel in YCbCr space; for reference-based perceptual quality, we use LPIPS and DISTS; for no-reference quality, we adopt NIQE~\cite{zhang2015feature}, LIQE~\cite{zhang2023blind}, MUSIQ~\cite{ke2021musiq}, MANIQA~\cite{yang2022maniqa}, and TOPIQ~\cite{chen2024topiq}. As shown in Tab.~\ref{tab:sr-all-supp}, UARE achieves higher SSIM and lower LPIPS/DISTS than PURE, and clearly outperforms all competing methods on MUSIQ, MANIQA, and TOPIQ, while ranking second-best on NIQE and LIQE. These results confirm that UARE delivers the best overall perceptual quality among all compared methods.

Additionally, we evaluate UARE on RealSet80~\cite{yue2024resshift}, which contains 80 low-resolution real-world images without ground-truth references. We compare UARE against BSRGAN, StableSR, DiffBIR, SeeSR, SinSR, OSEDiff, and PURE using five no-reference image quality metrics: NIQE, LIQE, MUSIQ, MANIQA, and TOPIQ. As reported in Tab.~\ref{tab:realset80}, UARE ranks first on MUSIQ and second on NIQE and MANIQA, further demonstrating its effectiveness in challenging real-world scenarios.

\noindent\textbf{More qualitative comparisons.} Figs.~\ref{fig:Q2_RealSR_Canon_043},~\ref{fig:Q3_RealSR_Canon_050},~\ref{fig:Q5_DRealSR_1425} and~\ref{fig:Q9_DIV2K512_0000098} present visual comparisons across super-resolution images produced by these approaches. It can be seen that our method effectively restores fine image details, such as knots, text, and petals, while producing noticeably fewer artifacts. These results comprehensively confirm the effectiveness of UARE in image super-resolution.

\begin{table}[t]
\centering
\caption{Quantitative comparison of different methods on four single-degradation subset of FoundIR. For each method, the first row lists PSNR/LPIPS and the second row lists NIQE/MANIQA.}
\setlength{\tabcolsep}{6pt}
\renewcommand{\arraystretch}{1.2}
\resizebox{\linewidth}{!}{
\begin{tabular}{l | c c c c}
\shline
Method & Blur & Haze & RainDrop & Lowlight  \\
\hline \hline
\multirow{2}{*}{Restormer~\cite{Restormer}}
  & 21.53 / 0.3821 & 13.70 / 0.5687 & 24.63 / 0.3029 &  9.20 / 0.6037 \\
  &  6.9553 / 0.1322 &  5.6395 / 0.2695 &  5.0340 / 0.2777 &  7.3095 / 0.2207 \\
\hline
\multirow{2}{*}{PromptIR~\cite{PromptIR}}
  & 21.64 / 0.4041 & 18.31 / 0.4762 & 26.67 / 0.2203 & 16.67 / 0.4804 \\
  &  8.0897 / 0.1361 &  5.7593 / 0.2815 &  4.8115 / 0.2449 &  7.7780 / 0.2317 \\
\hline
\multirow{2}{*}{DiffIR~\cite{lin2024diffbir}}
  & 21.61 / 0.3823 & 19.78 / 0.2345 & 26.19 / 0.2367 & 18.05 / 0.3146 \\
  &  7.4800 / 0.1390 & \secondbest{3.4547} / 0.3235 & \best{3.2762} / 0.2593 & \best{4.9243} / 0.2828 \\
\hline
\multirow{2}{*}{DiffUIR~\cite{DiffUIR}}
  & \best{26.99} / \third{0.1912} & \third{20.50} / \third{0.2037} & \secondbest{29.52} / \third{0.1292} & 14.88 / 0.2691 \\
  &  6.1600 / \third{0.2734} &  3.7435 / 0.3540 & \secondbest{4.0766} / 0.2697 &  6.0439 / \secondbest{0.3615} \\
\hline
\multirow{2}{*}{SUPIR~\cite{SUPIR}}
  & 20.63 / 0.3180 & 13.66 / 0.3883 & 21.41 / 0.3552 &  7.43 / 0.6460 \\
  & \secondbest{4.8374} / 0.2702 &  3.8944 / 0.3482 &  4.7940 / \third{0.2933} &  7.6004 / 0.2680 \\
\hline
\multirow{2}{*}{InstructIR~\cite{InstructIR}}
  & 19.81 / 0.2335 & 17.27 / 0.2267 & 21.75 / 0.3456 & \secondbest{20.78} / \secondbest{0.2473} \\
  &  6.1392 / 0.2394 & \third{3.4979} / \secondbest{0.3663} &  4.7295 / \secondbest{0.2935} & \secondbest{4.9299} / 0.3184 \\
\hline
\multirow{2}{*}{AutoDIR~\cite{AutoDIR}}
  & 19.98 / 0.3400 & 15.59 / 0.4130 & 21.46 / 0.3643 & \best{22.36} / 0.3487 \\
  &  6.8772 / 0.1652 &  4.9741 / 0.2909 &  5.2210 / 0.2836 &  7.6178 / 0.2567 \\
\hline
\multirow{2}{*}{FoundIR~\cite{li2024foundir}}
  & \secondbest{26.10} / \best{0.1709} & \best{23.29} / \secondbest{0.1896} & \best{30.86} / \best{0.0897} & \third{20.34} / \third{0.2499} \\
  & \third{5.6797} / \secondbest{0.2854} &  3.9543 / \third{0.3544} & \third{4.3657} / 0.2793 &  6.7511 / \third{0.3460} \\
\hline
\multirow{2}{*}{\textbf{UARE (Ours)}}
  & \third{22.38} / \secondbest{0.1904} & \secondbest{21.28} / \best{0.1635} & \third{28.26} / \secondbest{0.0981} & 19.64 / \best{0.1841} \\
  & \best{4.7313} / \best{0.3361} & \best{3.2674} / \best{0.4232} &  4.6515 / \best{0.3008} & \third{5.0119} / \best{0.4234} \\
\shline
\end{tabular}}
\label{tab:restoration_single}
\end{table}

\subsection{Image Restoration and Enhancement}
\label{sec:B.2}

\noindent\textbf{More quantitative results.}
We have reported the restoration/enhancement results on the multi-degradation subsets of FoundIR~\cite{li2024foundir} in the main paper. Here, we further compare UARE with Restormer~\cite{Restormer}, PromptIR~\cite{PromptIR}, DiffIR~\cite{DiffIR}, DiffUIR~\cite{DiffUIR}, SUPIR~\cite{SUPIR}, InstructIR~\cite{InstructIR}, AutoDIR~\cite{AutoDIR}, and FoundIR~\cite{li2024foundir} on the single-degradation subsets of FoundIR, including blur, haze, raindrop, and low-light, as shown in Tab.~\ref{tab:restoration_single}. UARE ranks first in MANIQA across all subsets. In addition, it achieves competitive PSNR, LPIPS, and NIQE results, indicating superior performance and a favorable trade-off between fidelity and perceptual quality.

\begin{figure*}[!t]
\setlength{\tabcolsep}{0.5pt}
\centering
\footnotesize
\resizebox{\linewidth}{!}{%
\begin{minipage}{\linewidth}
\centering
\begin{tabular}{@{}ccccccc@{}}
LR
& DGUNet~\cite{DGUNet}
& Restormer~\cite{Restormer}
& IDR~\cite{IDR}
& PromptIR~\cite{PromptIR}
& DiffIR~\cite{DiffIR}
& DiffUIR~\cite{DiffUIR} \\
\includegraphics[width=0.14\textwidth]{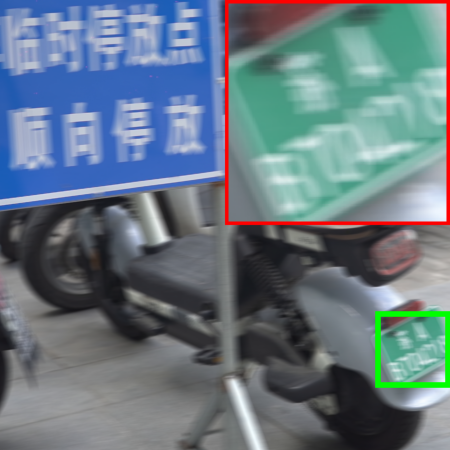}
& \includegraphics[width=0.14\textwidth]{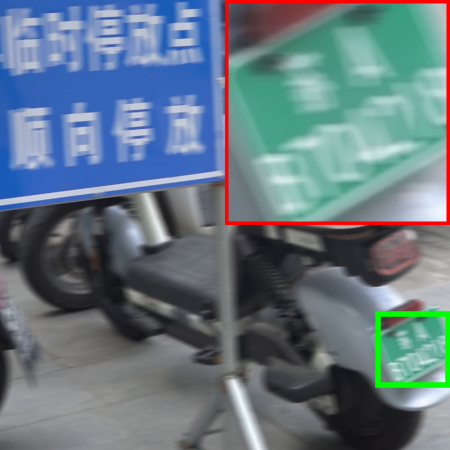}
& \includegraphics[width=0.14\textwidth]{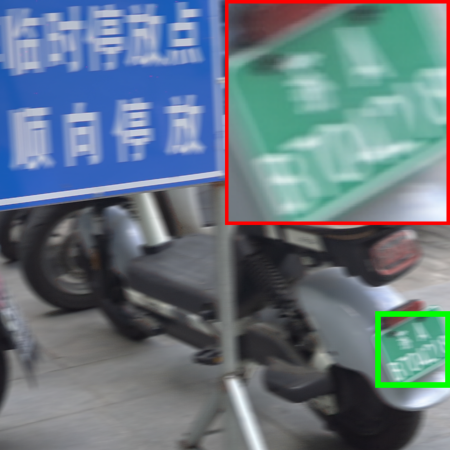}
& \includegraphics[width=0.14\textwidth]{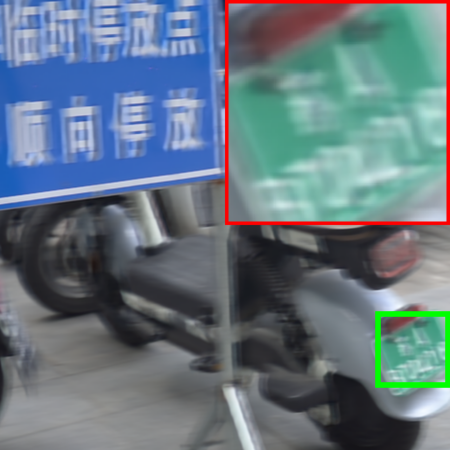}
& \includegraphics[width=0.14\textwidth]{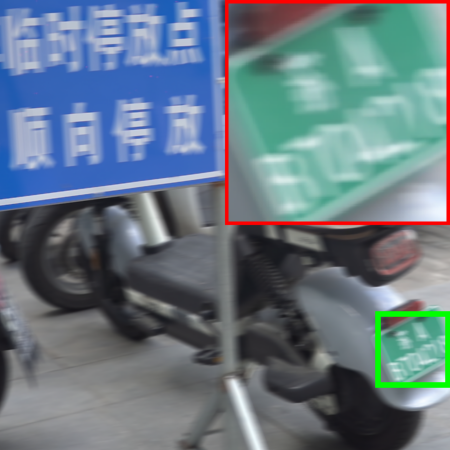}
& \includegraphics[width=0.14\textwidth]{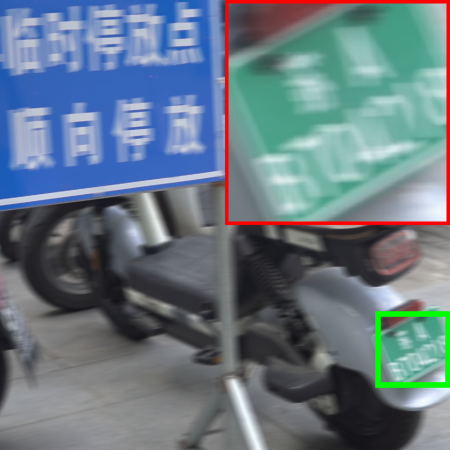}
& \includegraphics[width=0.14\textwidth]{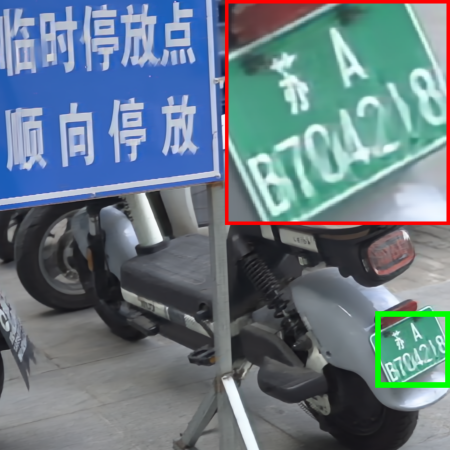} \\[1pt]

DA-CLIP~\cite{DA-CLIP}
& SUPIR~\cite{SUPIR}
& InstructIR~\cite{InstructIR}
& AutoDIR~\cite{AutoDIR}
& FoundIR~\cite{li2024foundir}
& 	\textbf{UARE (Ours)}
& GT \\
\includegraphics[width=0.14\textwidth]{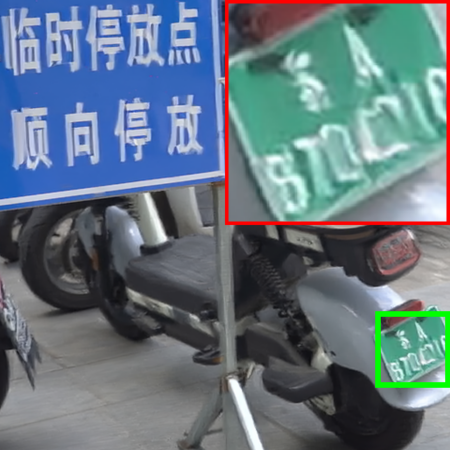}
& \includegraphics[width=0.14\textwidth]{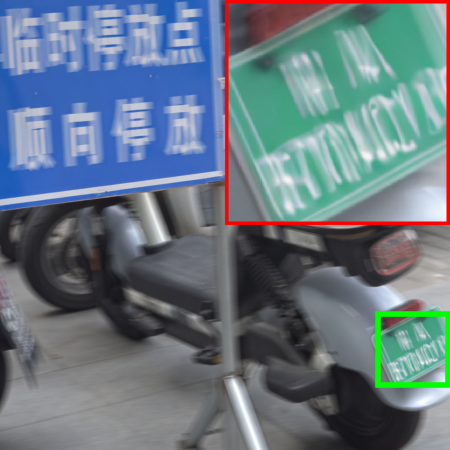}
& \includegraphics[width=0.14\textwidth]{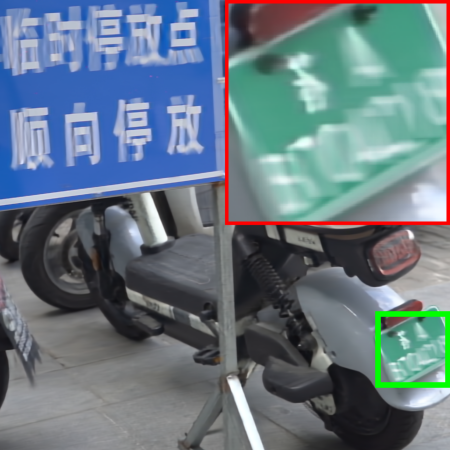}
& \includegraphics[width=0.14\textwidth]{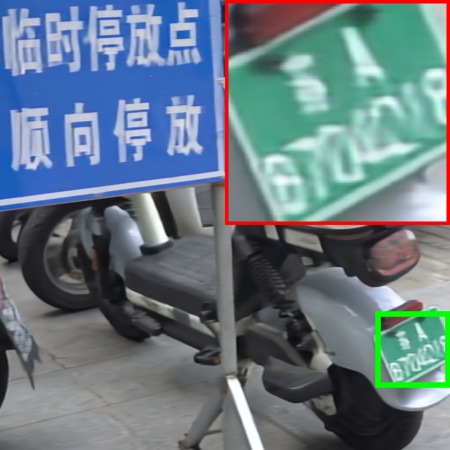}
& \includegraphics[width=0.14\textwidth]{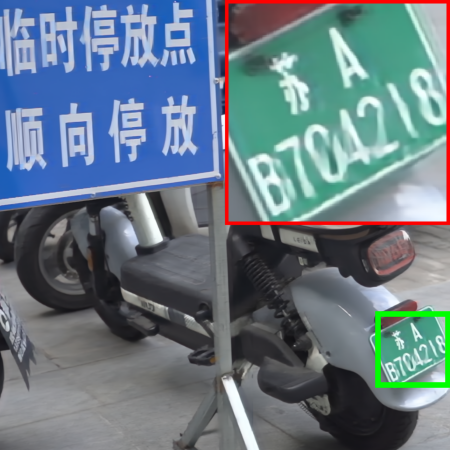}
& \includegraphics[width=0.14\textwidth]{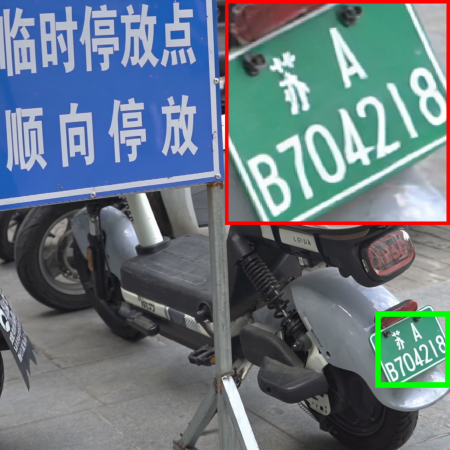}
& \includegraphics[width=0.14\textwidth]{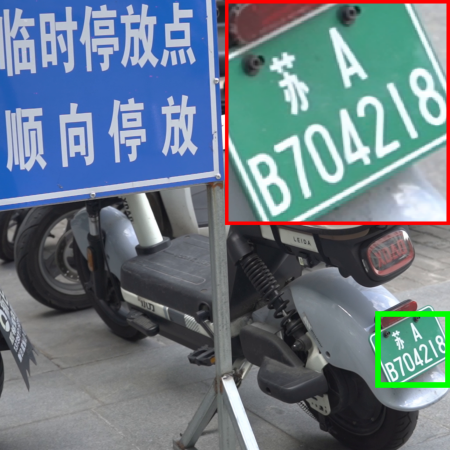} \\
\end{tabular}
\end{minipage}}
\vspace{-10pt}
\caption{Visual comparison on the image ``0131'' with blur from the FoundIR dataset.}
\label{fig:Q11_FoundIR_01_0131}
\end{figure*}

\begin{figure*}[!t]
\setlength{\tabcolsep}{0.5pt}
\centering
\footnotesize
\resizebox{\linewidth}{!}{%
\begin{minipage}{\linewidth}
\centering
\begin{tabular}{@{}ccccccc@{}}
LR
& DGUNet~\cite{DGUNet}
& Restormer~\cite{Restormer}
& IDR~\cite{IDR}
& PromptIR~\cite{PromptIR}
& DiffIR~\cite{DiffIR}
& DiffUIR~\cite{DiffUIR} \\
\includegraphics[width=0.14\textwidth]{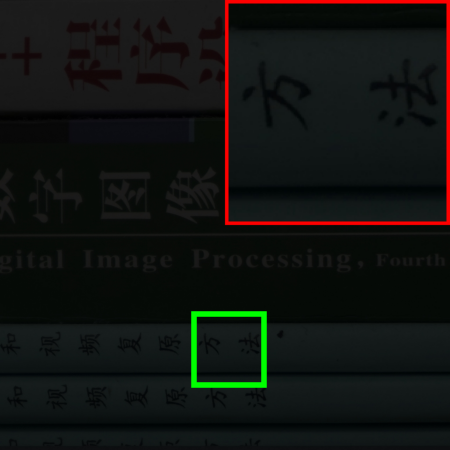}
& \includegraphics[width=0.14\textwidth]{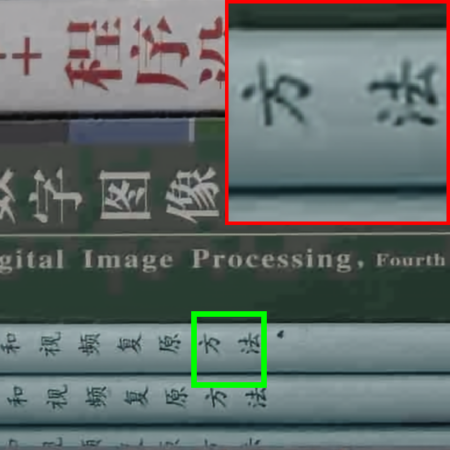}
& \includegraphics[width=0.14\textwidth]{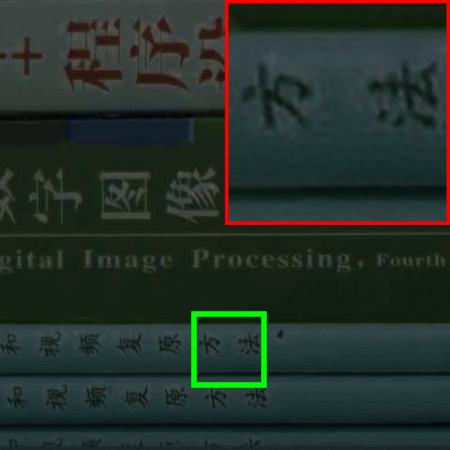}
& \includegraphics[width=0.14\textwidth]{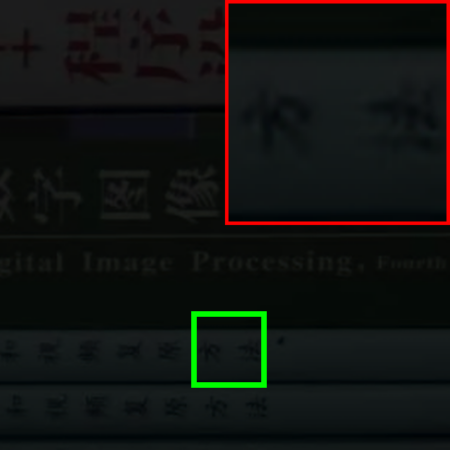}
& \includegraphics[width=0.14\textwidth]{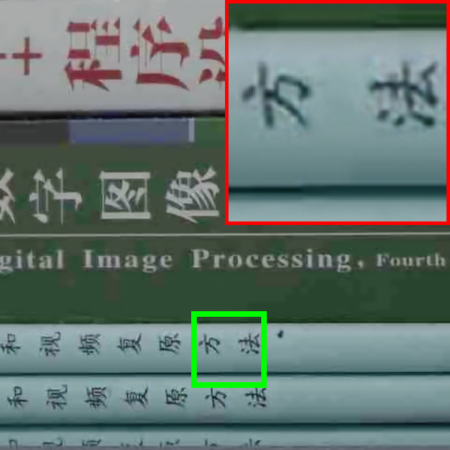}
& \includegraphics[width=0.14\textwidth]{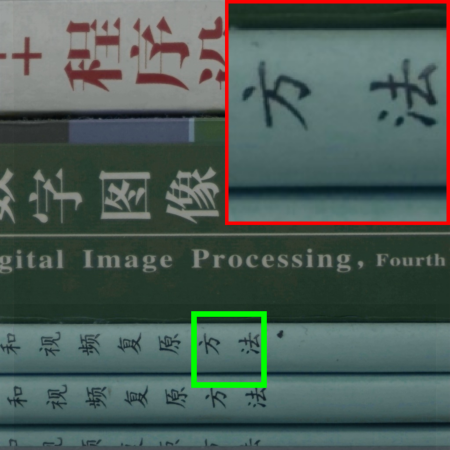}
& \includegraphics[width=0.14\textwidth]{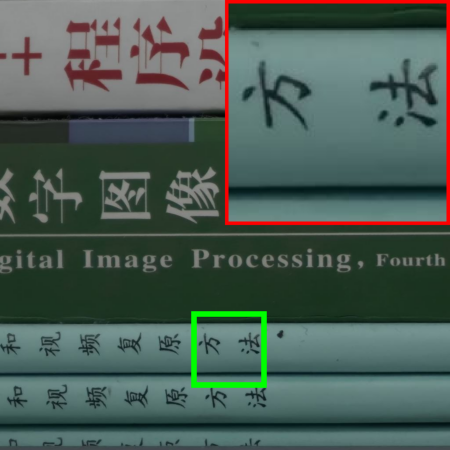} \\[1pt]

DA-CLIP~\cite{DA-CLIP}
& SUPIR~\cite{SUPIR}
& InstructIR~\cite{InstructIR}
& AutoDIR~\cite{AutoDIR}
& FoundIR~\cite{li2024foundir}
& 	\textbf{UARE (Ours)}
& GT \\
\includegraphics[width=0.14\textwidth]{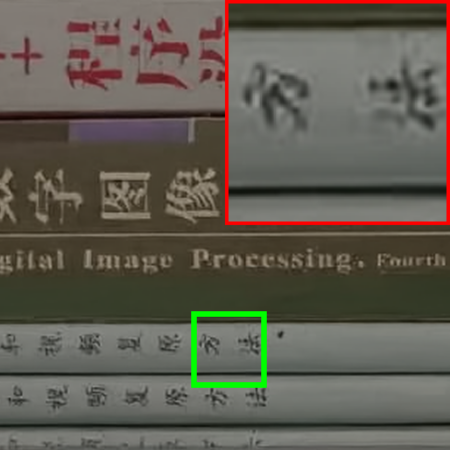}
& \includegraphics[width=0.14\textwidth]{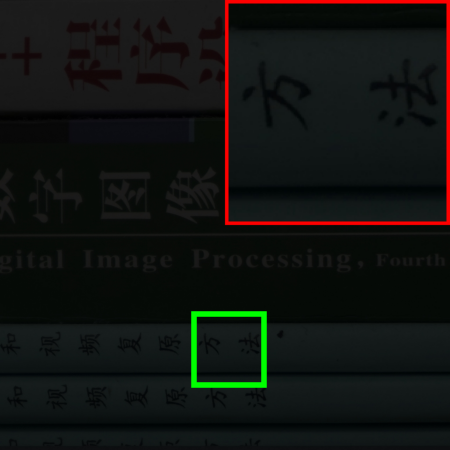}
& \includegraphics[width=0.14\textwidth]{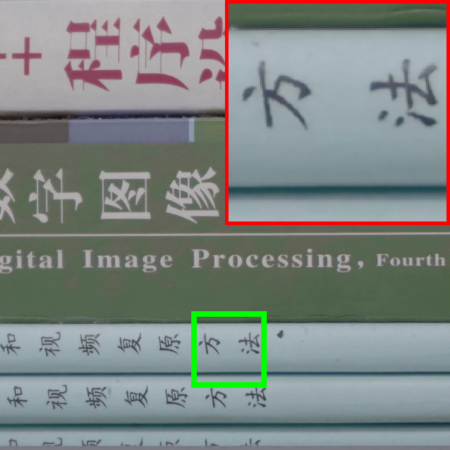}
& \includegraphics[width=0.14\textwidth]{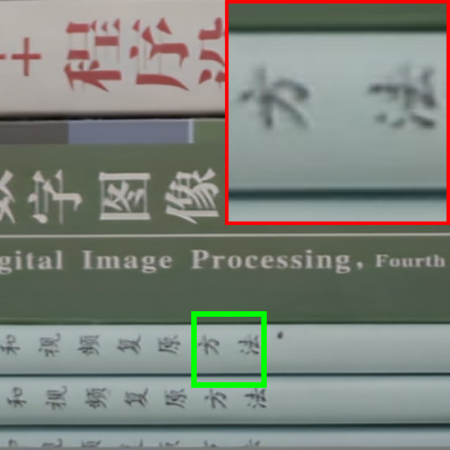}
& \includegraphics[width=0.14\textwidth]{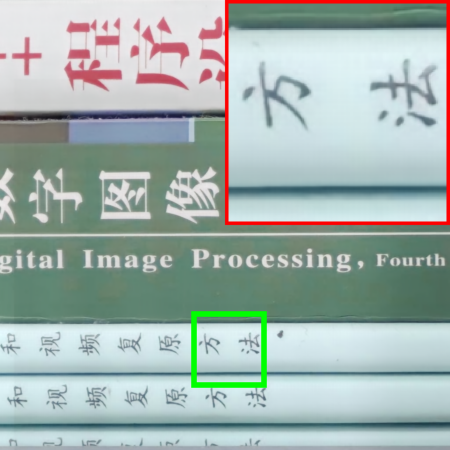}
& \includegraphics[width=0.14\textwidth]{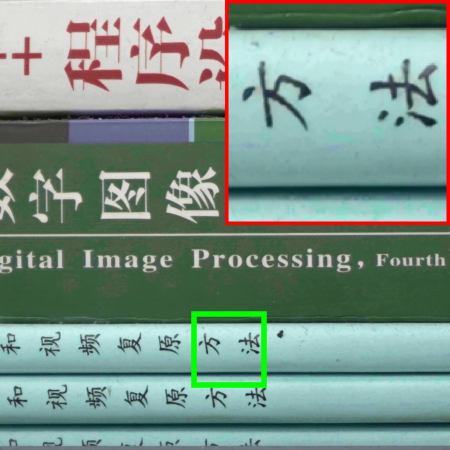}
& \includegraphics[width=0.14\textwidth]{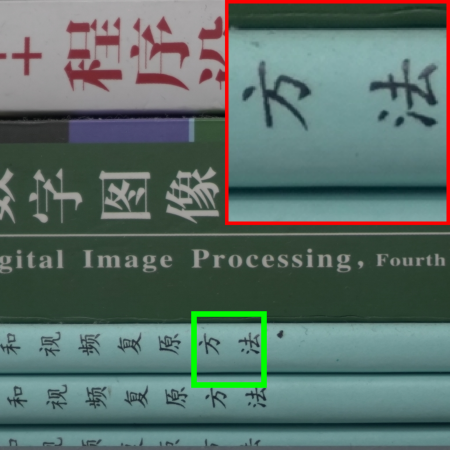} \\
\end{tabular}
\end{minipage}}
\vspace{-10pt}
\caption{Visual comparison on the image ``1143'' with low-light from the FoundIR dataset.}
\label{fig:Q14_FoundIR-14_1143}
\end{figure*}

\begin{figure*}[!t]
\setlength{\tabcolsep}{0.5pt}
\centering
\footnotesize
\resizebox{\linewidth}{!}{%
\begin{minipage}{\linewidth}
\centering
\begin{tabular}{@{}ccccccc@{}}
LR
& DGUNet~\cite{DGUNet}
& Restormer~\cite{Restormer}
& IDR~\cite{IDR}
& PromptIR~\cite{PromptIR}
& DiffIR~\cite{DiffIR}
& DiffUIR~\cite{DiffUIR} \\
\includegraphics[width=0.14\textwidth]{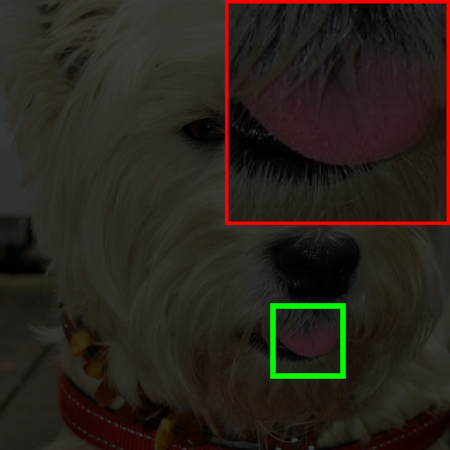}
& \includegraphics[width=0.14\textwidth]{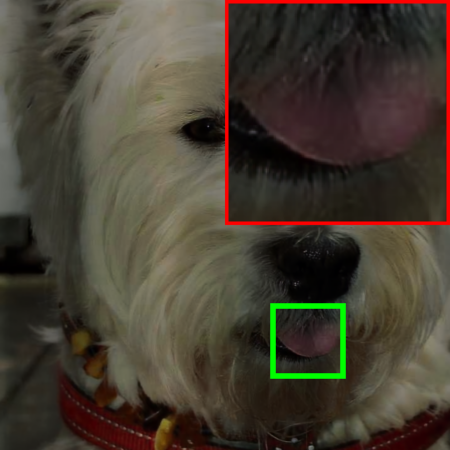}
& \includegraphics[width=0.14\textwidth]{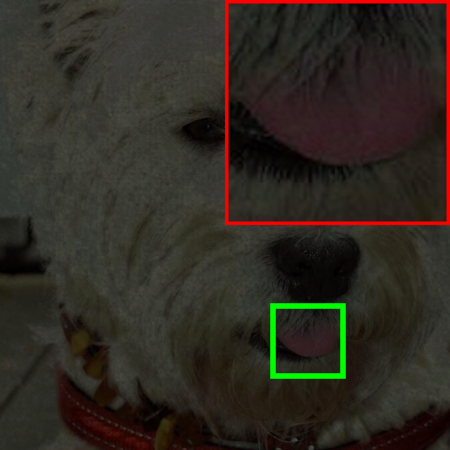}
& \includegraphics[width=0.14\textwidth]{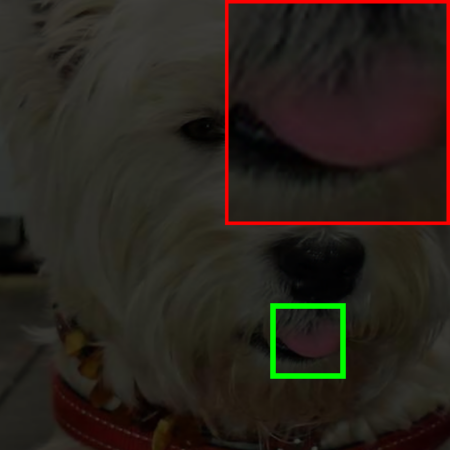}
& \includegraphics[width=0.14\textwidth]{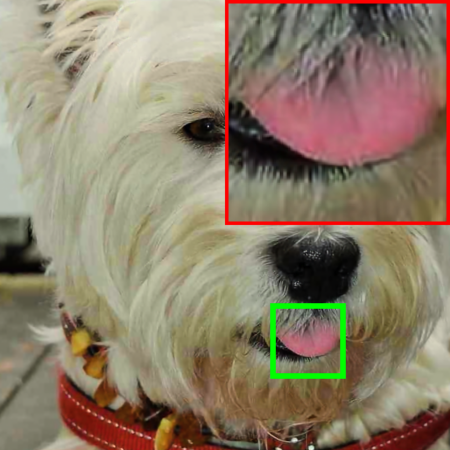}
& \includegraphics[width=0.14\textwidth]{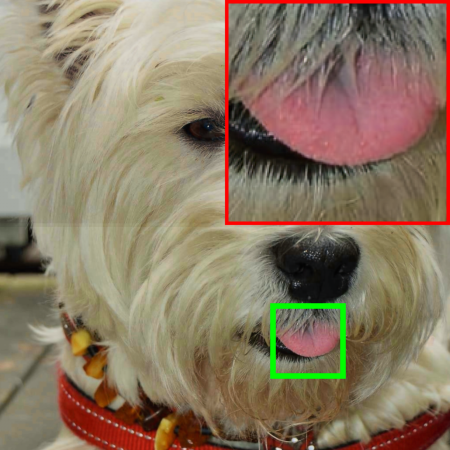}
& \includegraphics[width=0.14\textwidth]{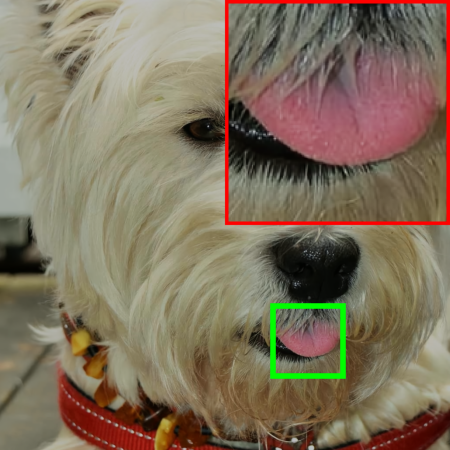} \\[1pt]

DA-CLIP~\cite{DA-CLIP}
& SUPIR~\cite{SUPIR}
& InstructIR~\cite{InstructIR}
& AutoDIR~\cite{AutoDIR}
& FoundIR~\cite{li2024foundir}
& 	\textbf{UARE (Ours)}
& GT \\
\includegraphics[width=0.14\textwidth]{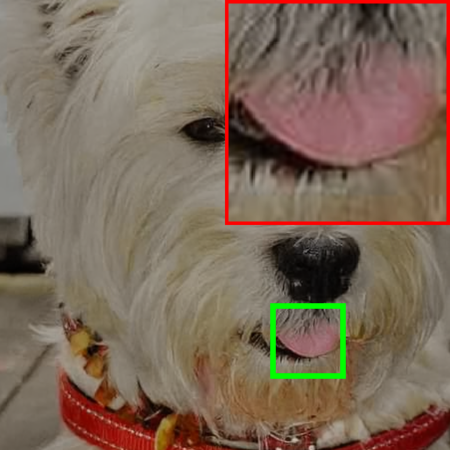}
& \includegraphics[width=0.14\textwidth]{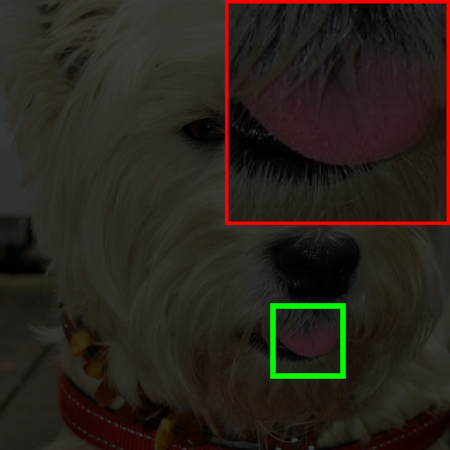}
& \includegraphics[width=0.14\textwidth]{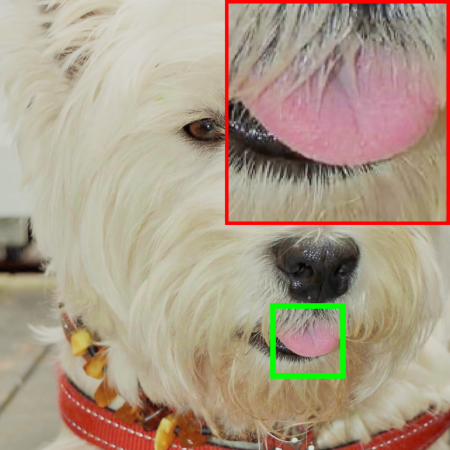}
& \includegraphics[width=0.14\textwidth]{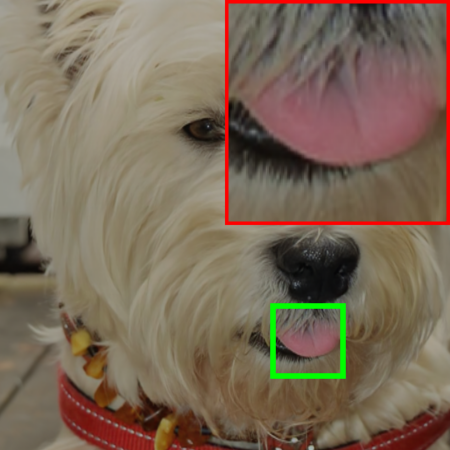}
& \includegraphics[width=0.14\textwidth]{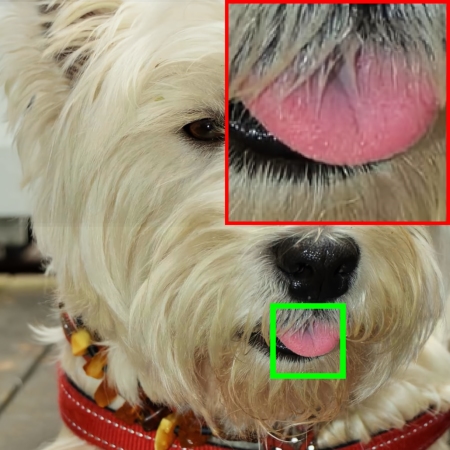}
& \includegraphics[width=0.14\textwidth]{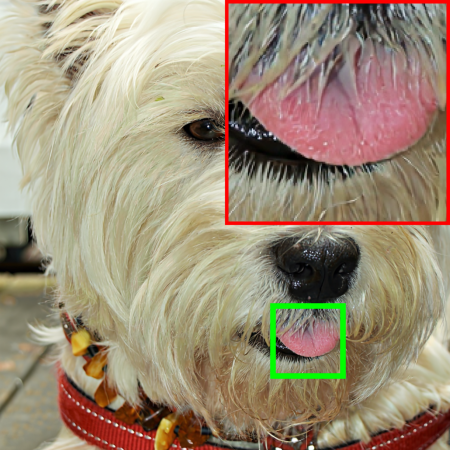}
& \includegraphics[width=0.14\textwidth]{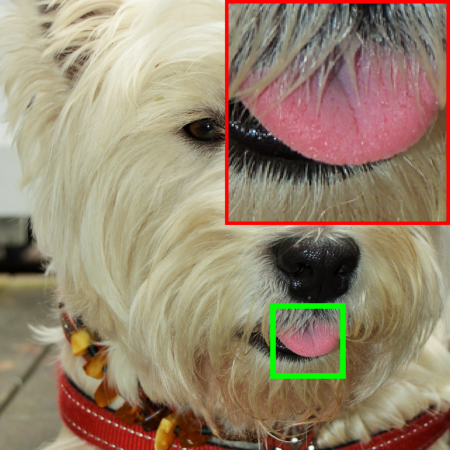} \\
\end{tabular}
\end{minipage}}
\vspace{-10pt}
\caption{Visual comparison on the image ``1304'' with low-light and JPEG compression from the FoundIR dataset.}
\label{fig:Q16_FoundIR_17_1304}
\end{figure*}

\begin{figure*}[!t]
\setlength{\tabcolsep}{0.5pt}
\centering
\footnotesize
\resizebox{\linewidth}{!}{%
\begin{minipage}{\linewidth}
\centering
\begin{tabular}{@{}ccccccc@{}}
LR
& DGUNet~\cite{DGUNet}
& Restormer~\cite{Restormer}
& IDR~\cite{IDR}
& PromptIR~\cite{PromptIR}
& DiffIR~\cite{DiffIR}
& DiffUIR~\cite{DiffUIR} \\
\includegraphics[width=0.14\textwidth]{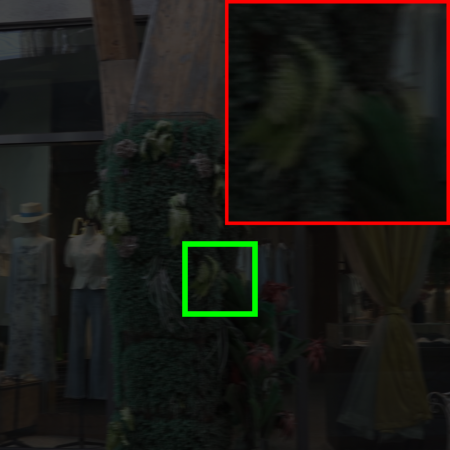}
& \includegraphics[width=0.14\textwidth]{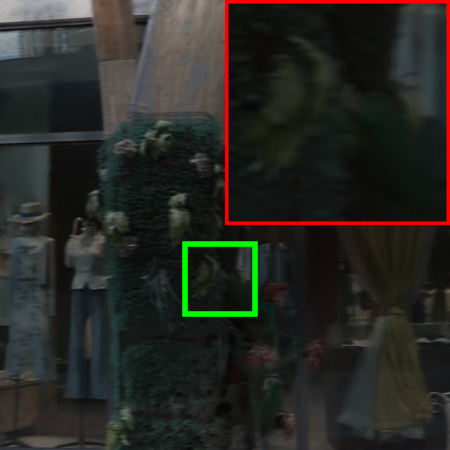}
& \includegraphics[width=0.14\textwidth]{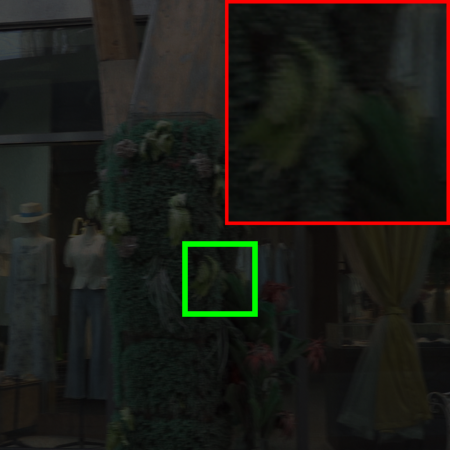}
& \includegraphics[width=0.14\textwidth]{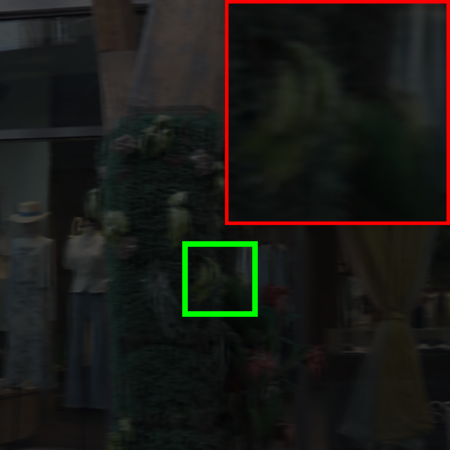}
& \includegraphics[width=0.14\textwidth]{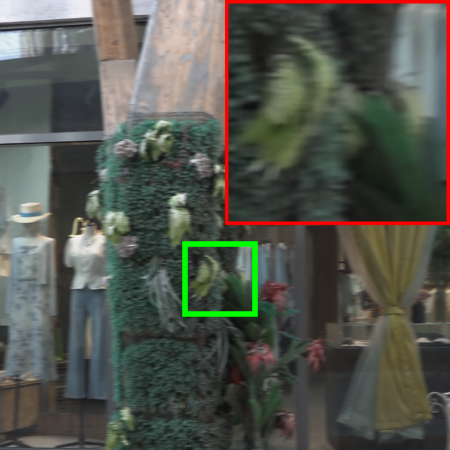}
& \includegraphics[width=0.14\textwidth]{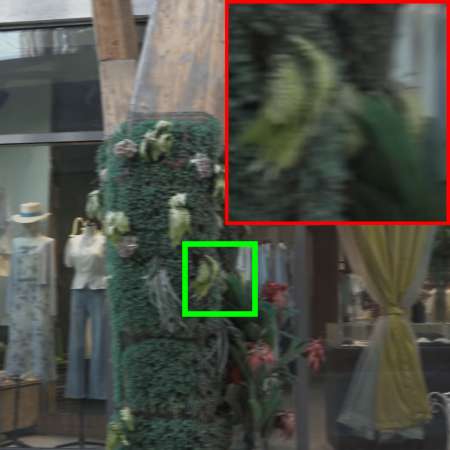}
& \includegraphics[width=0.14\textwidth]{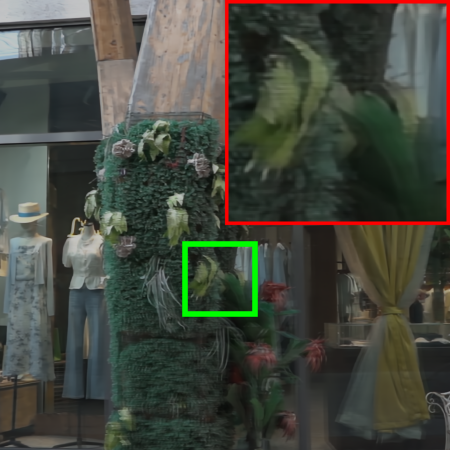} \\[1pt]

DA-CLIP~\cite{DA-CLIP}
& SUPIR~\cite{SUPIR}
& InstructIR~\cite{InstructIR}
& AutoDIR~\cite{AutoDIR}
& FoundIR~\cite{li2024foundir}
& 	\textbf{UARE (Ours)}
& GT \\
\includegraphics[width=0.14\textwidth]{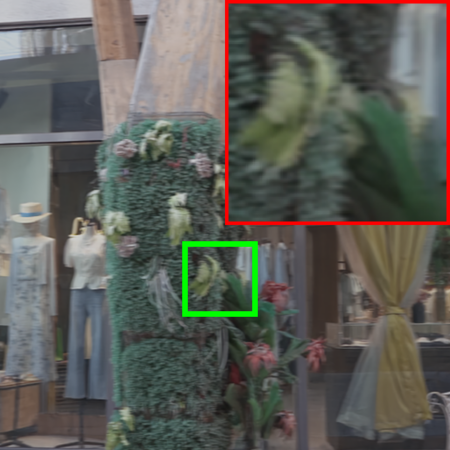}
& \includegraphics[width=0.14\textwidth]{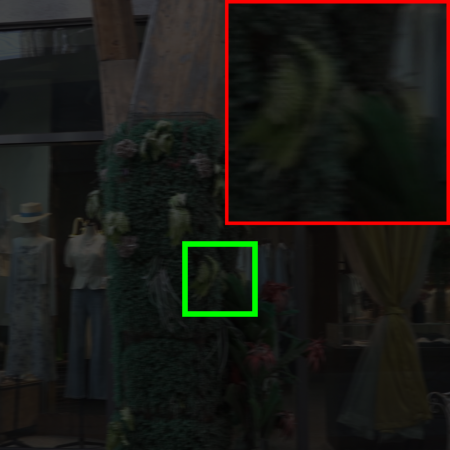}
& \includegraphics[width=0.14\textwidth]{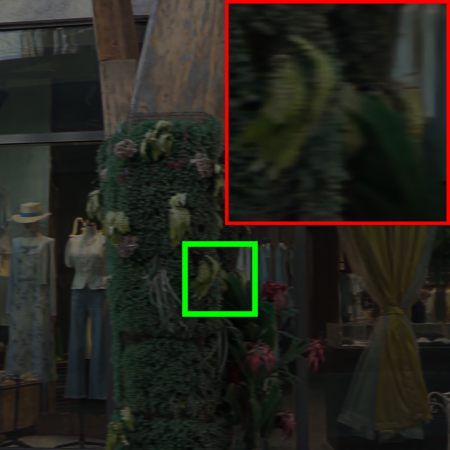}
& \includegraphics[width=0.14\textwidth]{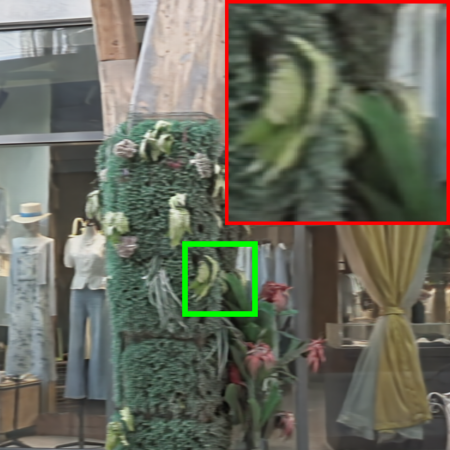}
& \includegraphics[width=0.14\textwidth]{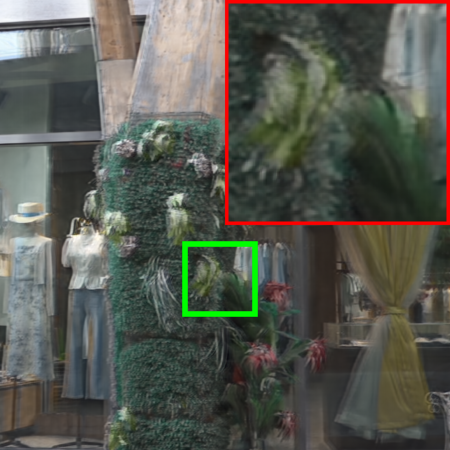}
& \includegraphics[width=0.14\textwidth]{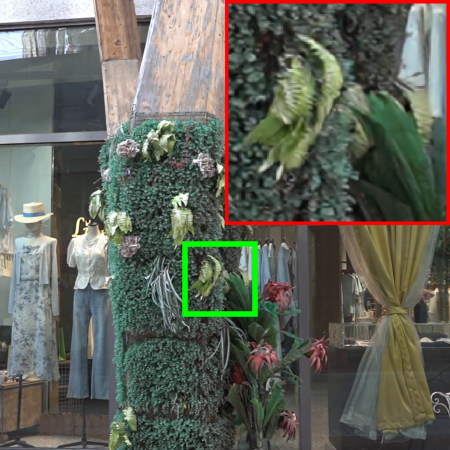}
& \includegraphics[width=0.14\textwidth]{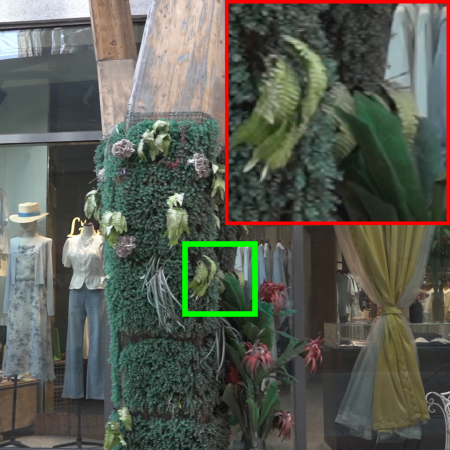} \\
\end{tabular}
\end{minipage}}
\vspace{-10pt}
\caption{Visual comparison on the image ``1397'' with low-light, blur and noise from the FoundIR dataset.}
\label{fig:Q18_FoundIR_18_1397}
\end{figure*}

\begin{figure*}[!t]
\setlength{\tabcolsep}{0.5pt}
\centering
\footnotesize
\resizebox{\linewidth}{!}{%
\begin{minipage}{\linewidth}
\centering
\begin{tabular}{@{}ccccccc@{}}
LR
& DGUNet~\cite{DGUNet}
& Restormer~\cite{Restormer}
& IDR~\cite{IDR}
& PromptIR~\cite{PromptIR}
& DiffIR~\cite{DiffIR}
& DiffUIR~\cite{DiffUIR} \\
\includegraphics[width=0.14\textwidth]{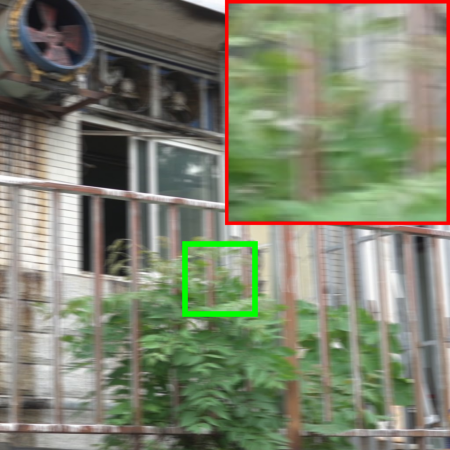}
& \includegraphics[width=0.14\textwidth]{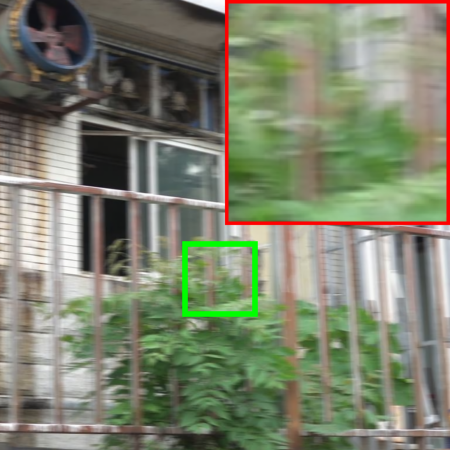}
& \includegraphics[width=0.14\textwidth]{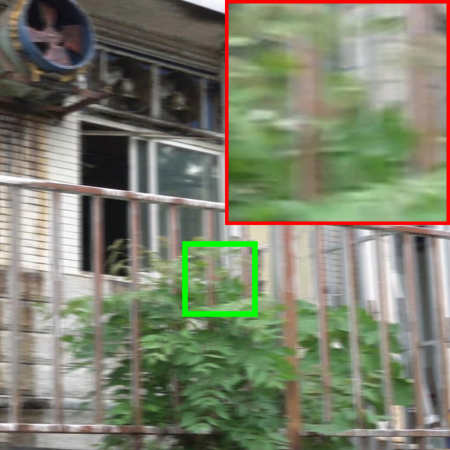}
& \includegraphics[width=0.14\textwidth]{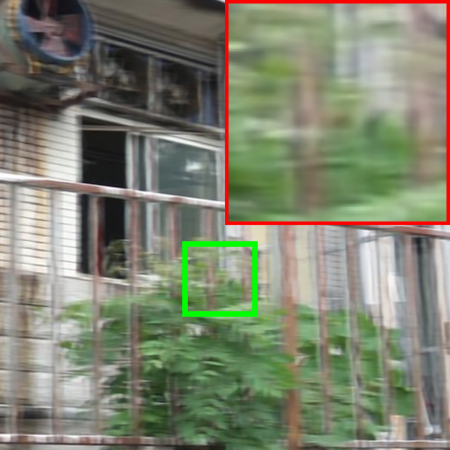}
& \includegraphics[width=0.14\textwidth]{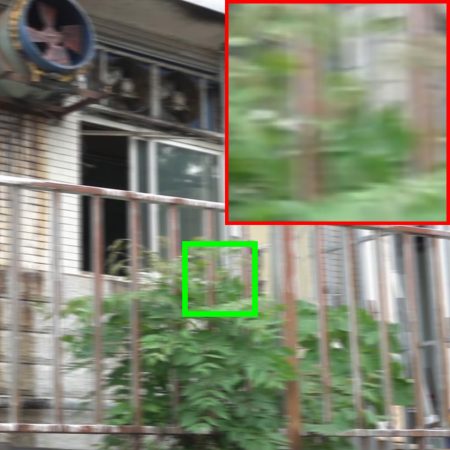}
& \includegraphics[width=0.14\textwidth]{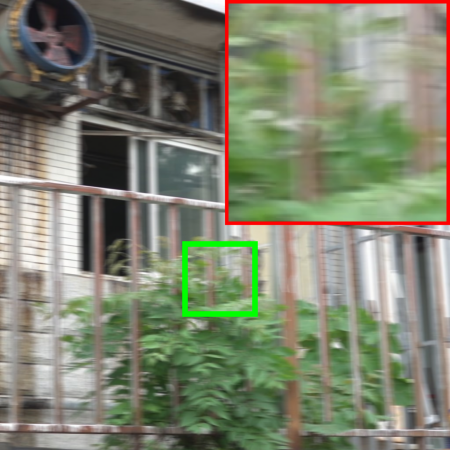}
& \includegraphics[width=0.14\textwidth]{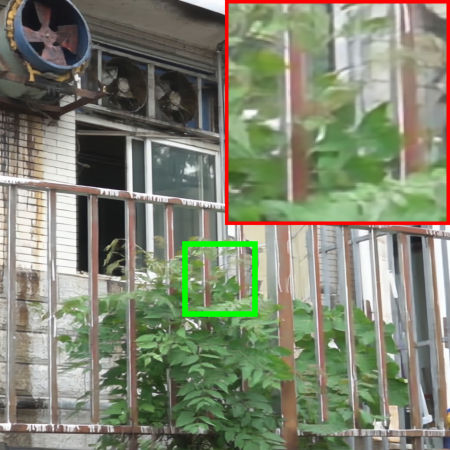} \\[1pt]

DA-CLIP~\cite{DA-CLIP}
& SUPIR~\cite{SUPIR}
& InstructIR~\cite{InstructIR}
& AutoDIR~\cite{AutoDIR}
& FoundIR~\cite{li2024foundir}
& 	\textbf{UARE (Ours)}
& GT \\
\includegraphics[width=0.14\textwidth]{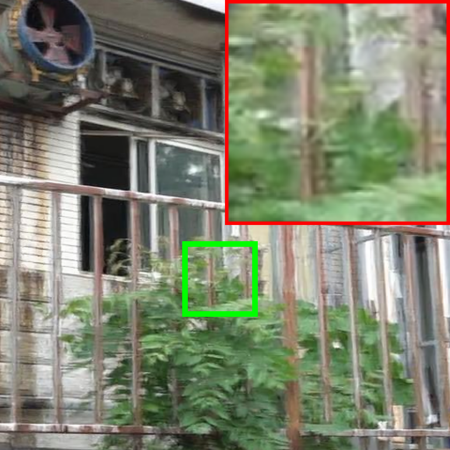}
& \includegraphics[width=0.14\textwidth]{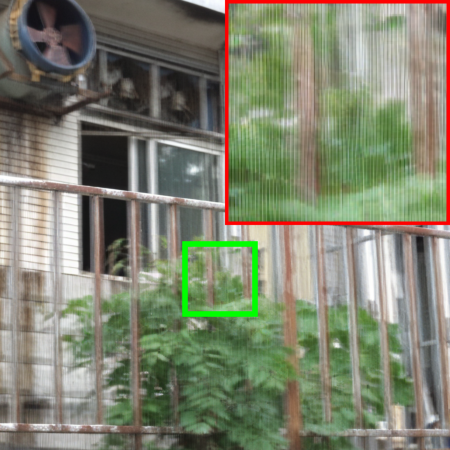}
& \includegraphics[width=0.14\textwidth]{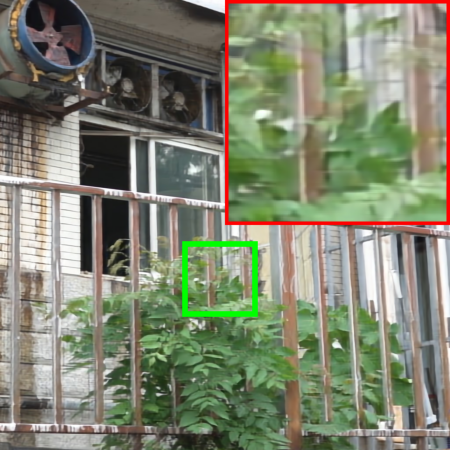}
& \includegraphics[width=0.14\textwidth]{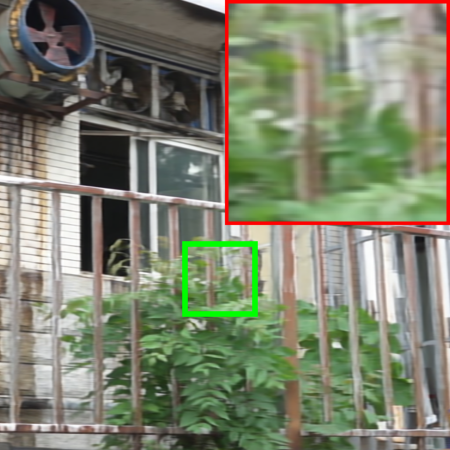}
& \includegraphics[width=0.14\textwidth]{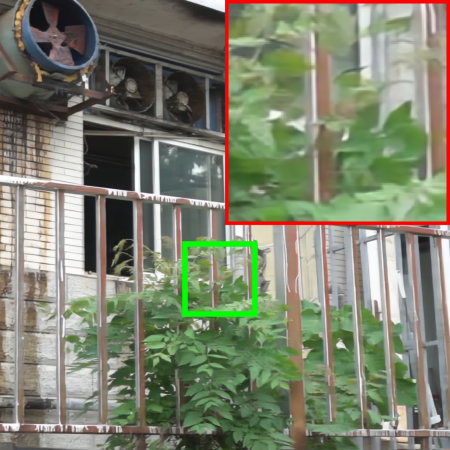}
& \includegraphics[width=0.14\textwidth]{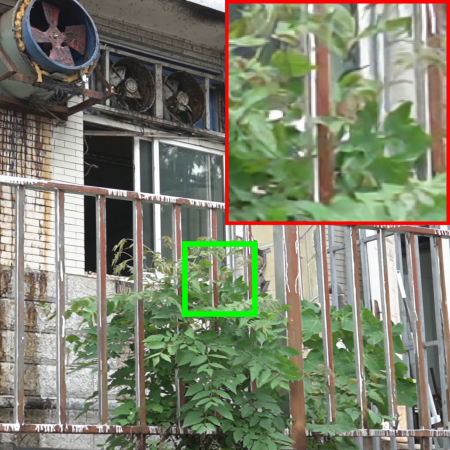}
& \includegraphics[width=0.14\textwidth]{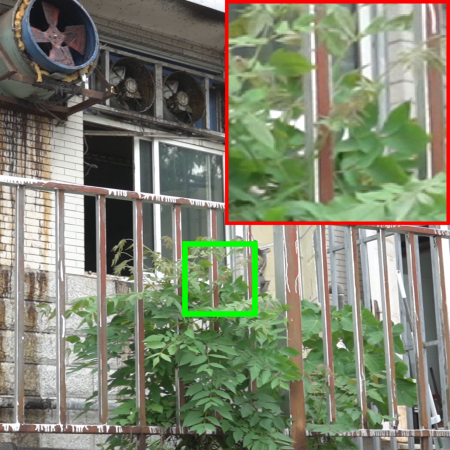} \\
\end{tabular}
\end{minipage}}
\vspace{-10pt}
\caption{Visual comparison on the image ``0243'' with blur and JPEG compression from the FoundIR dataset.}
\label{fig:Q20_FoundIR_03_243}
\end{figure*}

\noindent\textbf{More qualitative results.} Figs.~\ref{fig:Q11_FoundIR_01_0131},~\ref{fig:Q14_FoundIR-14_1143},~\ref{fig:Q16_FoundIR_17_1304},~\ref{fig:Q18_FoundIR_18_1397} and~\ref{fig:Q20_FoundIR_03_243} present visual comparison across restored/enhanced images produced by these approaches. It can be seen that UARE faithfully reconstructs challenging details, such as text in blurred or low-light regions, hair, and the fine structures of flowers and vegetation. These results comprehensively confirm the effectiveness of UARE across multiple image restoration and enhancement tasks.

\begin{figure*}[!t]
\setlength{\tabcolsep}{0.5pt}
\centering
\footnotesize
\resizebox{\linewidth}{!}{%
\begin{minipage}{\linewidth}
\centering
\begin{tabular}{@{}ccccccc@{}}
Input
& Real-ESRGAN~\cite{Real-esrgan}
& FeMASR~\cite{chen2022real}
& SwinIR~\cite{liang2021swinir}
& StableSR~\cite{wang2024exploiting}
& DiffBIR~\cite{lin2024diffbir}
& SeeSR~\cite{wu2024seesr} \\
\includegraphics[width=0.14\textwidth]{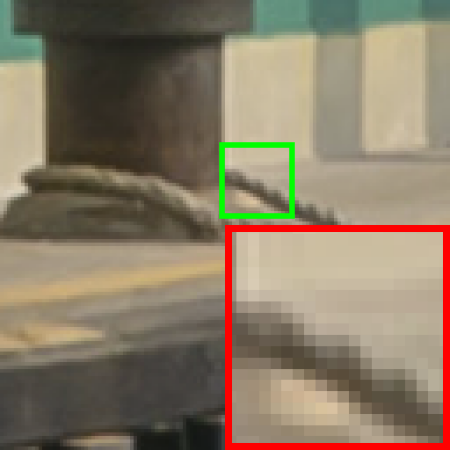}
& \includegraphics[width=0.14\textwidth]{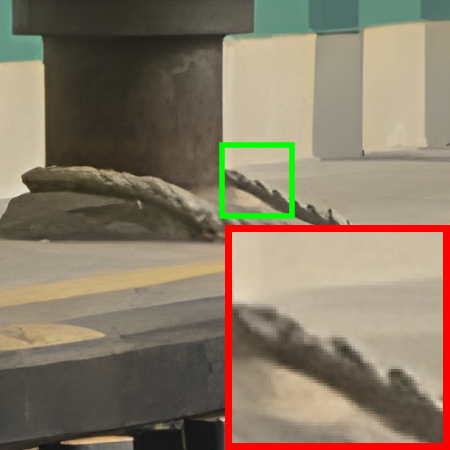}
& \includegraphics[width=0.14\textwidth]{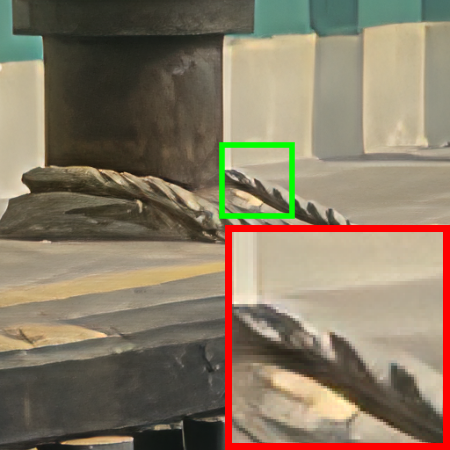}
& \includegraphics[width=0.14\textwidth]{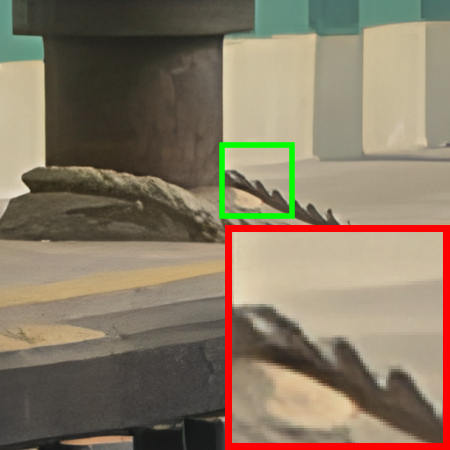}
& \includegraphics[width=0.14\textwidth]{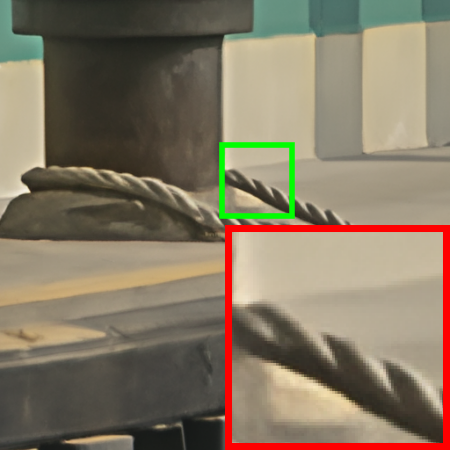}
& \includegraphics[width=0.14\textwidth]{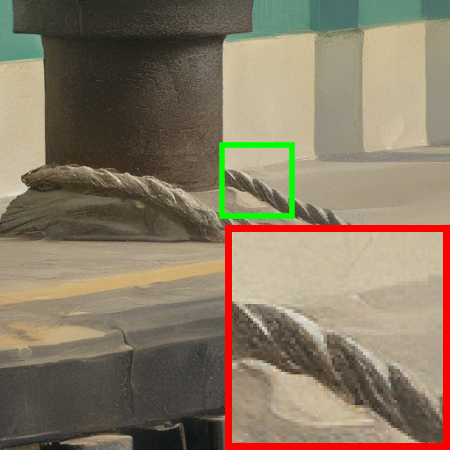}
& \includegraphics[width=0.14\textwidth]{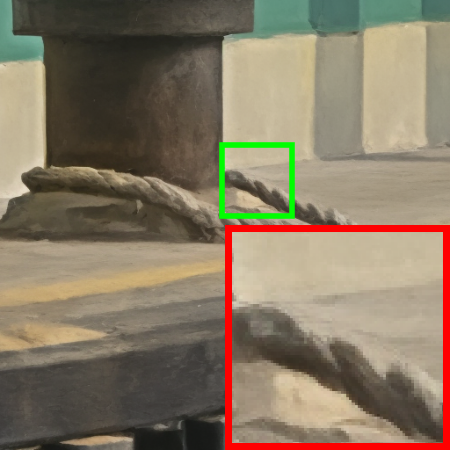} \\[1pt]

PASD~\cite{yang2024pixel}
& ResShift~\cite{yue2024resshift}
& SinSR~\cite{wang2024sinsr}
& OSEDiff~\cite{wu2024one}
& S3Diff~\cite{zhang2024degradation}
& PURE~\cite{wei2025perceive}
& \textbf{UARE (Ours)} \\
\includegraphics[width=0.14\textwidth]{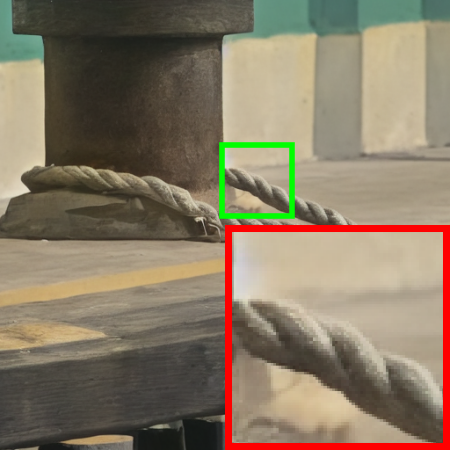}
& \includegraphics[width=0.14\textwidth]{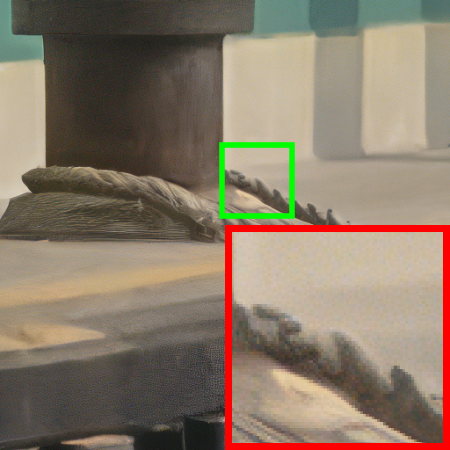}
& \includegraphics[width=0.14\textwidth]{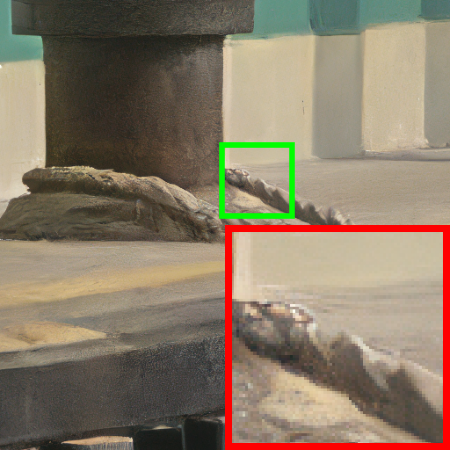}
& \includegraphics[width=0.14\textwidth]{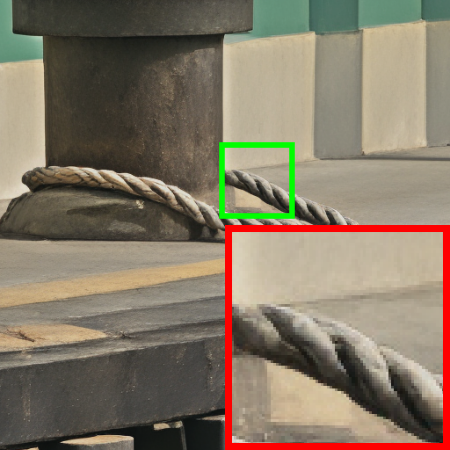}
& \includegraphics[width=0.14\textwidth]{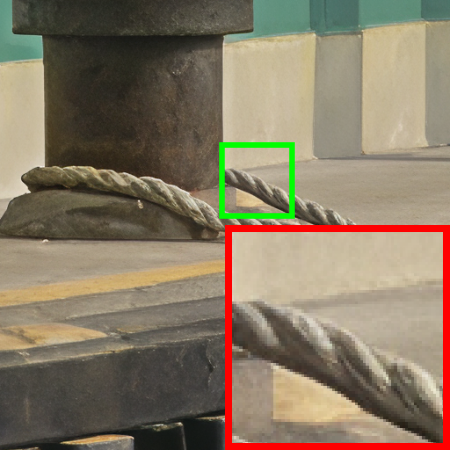}
& \includegraphics[width=0.14\textwidth]{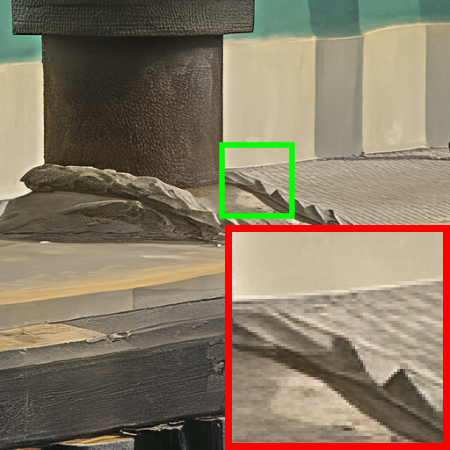}
& \includegraphics[width=0.14\textwidth]{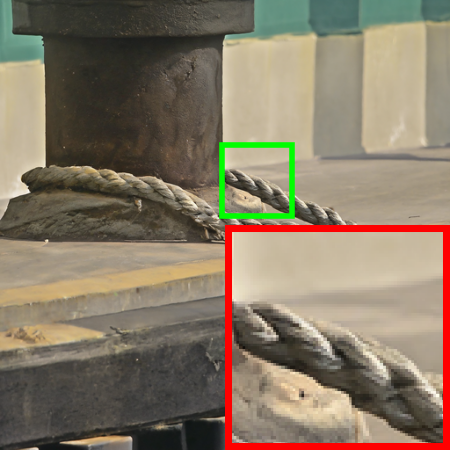} \\
\end{tabular}
\end{minipage}}
\vspace{-10pt}
\caption{Visual comparison on the image named ``Canon\_043'' from the RealSR dataset.}
\label{fig:Q2_RealSR_Canon_043}
\end{figure*}

\begin{figure*}[!t]
\setlength{\tabcolsep}{0.5pt}
\centering
\footnotesize
\resizebox{\linewidth}{!}{%
\begin{minipage}{\linewidth}
\centering
\begin{tabular}{@{}ccccccc@{}}
Input
& Real-ESRGAN~\cite{Real-esrgan}
& FeMASR~\cite{chen2022real}
& SwinIR~\cite{liang2021swinir}
& StableSR~\cite{wang2024exploiting}
& DiffBIR~\cite{lin2024diffbir}
& SeeSR~\cite{wu2024seesr} \\
\includegraphics[width=0.14\textwidth]{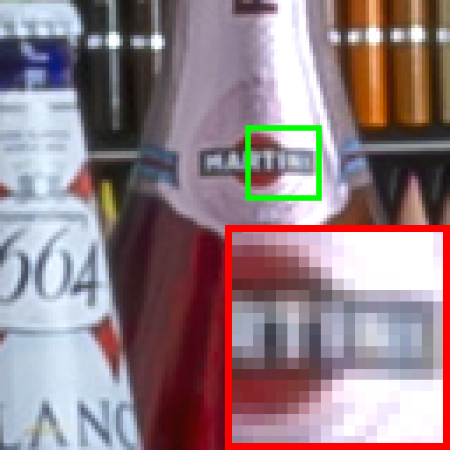}
& \includegraphics[width=0.14\textwidth]{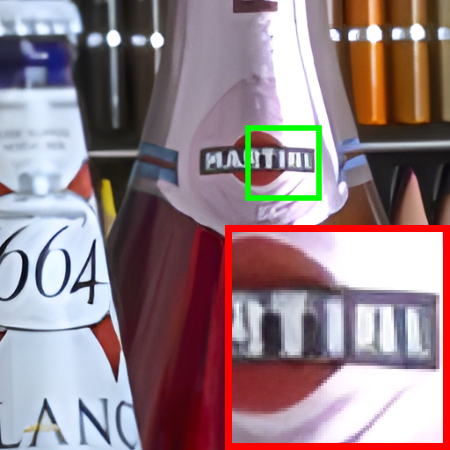}
& \includegraphics[width=0.14\textwidth]{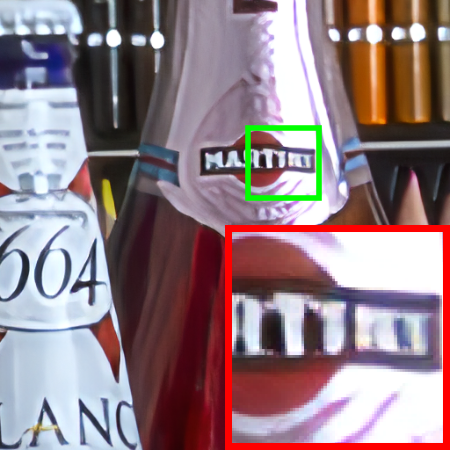}
& \includegraphics[width=0.14\textwidth]{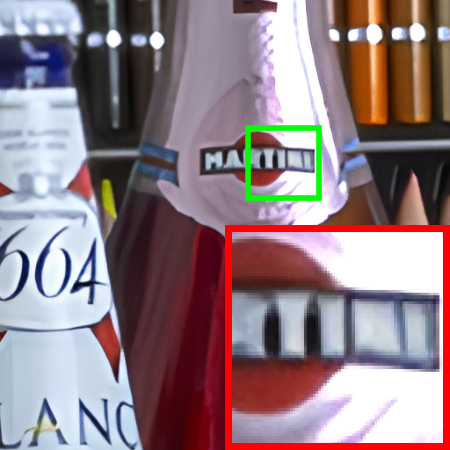}
& \includegraphics[width=0.14\textwidth]{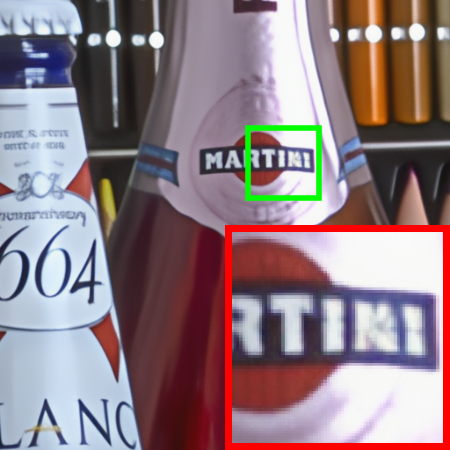}
& \includegraphics[width=0.14\textwidth]{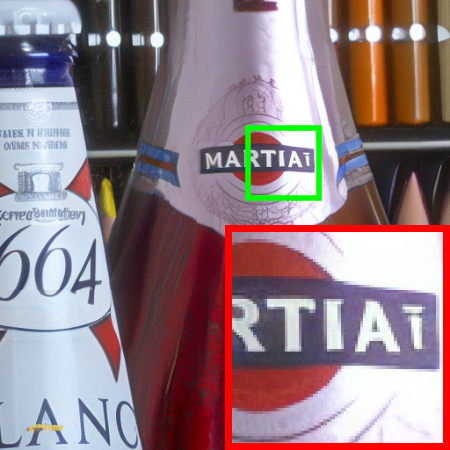}
& \includegraphics[width=0.14\textwidth]{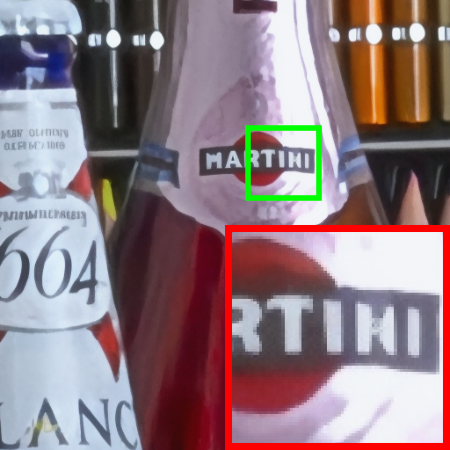} \\[1pt]

PASD~\cite{yang2024pixel}
& ResShift~\cite{yue2024resshift}
& SinSR~\cite{wang2024sinsr}
& OSEDiff~\cite{wu2024one}
& S3Diff~\cite{zhang2024degradation}
& PURE~\cite{wei2025perceive}
& \textbf{UARE (Ours)} \\
\includegraphics[width=0.14\textwidth]{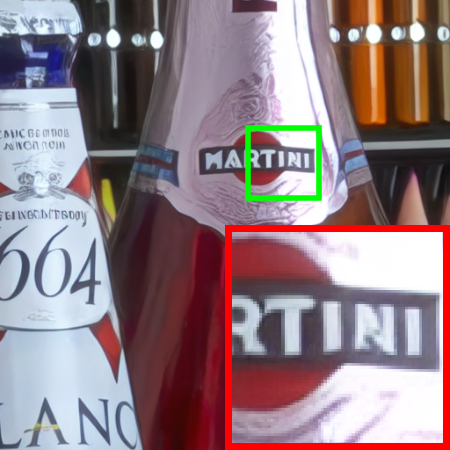}
& \includegraphics[width=0.14\textwidth]{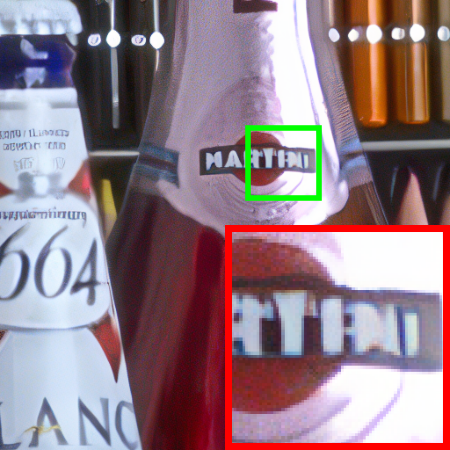}
& \includegraphics[width=0.14\textwidth]{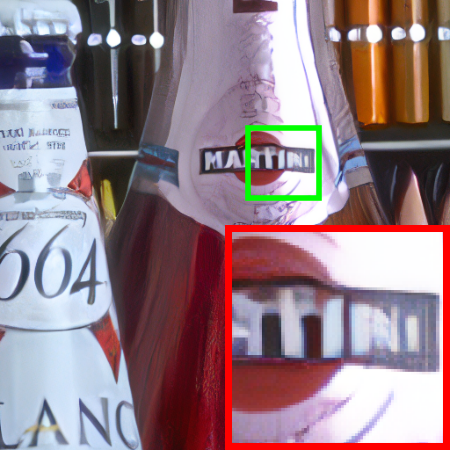}
& \includegraphics[width=0.14\textwidth]{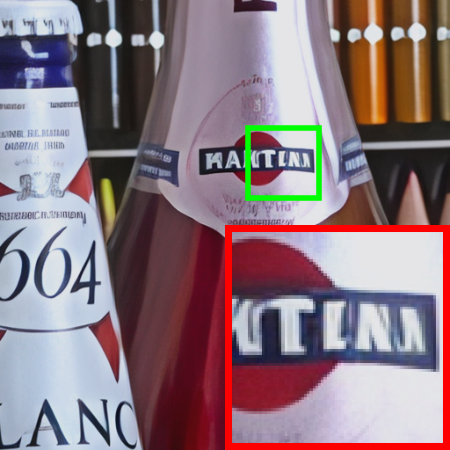}
& \includegraphics[width=0.14\textwidth]{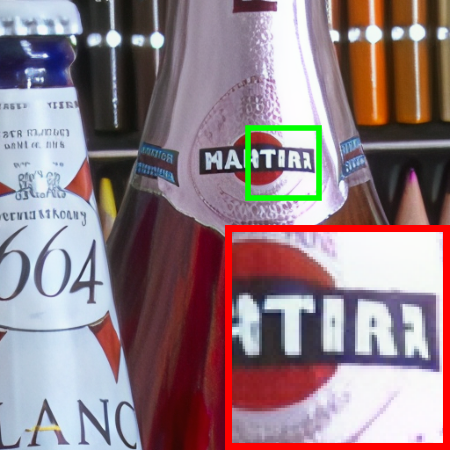}
& \includegraphics[width=0.14\textwidth]{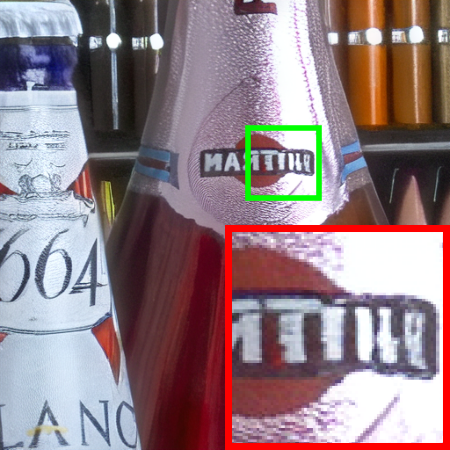}
& \includegraphics[width=0.14\textwidth]{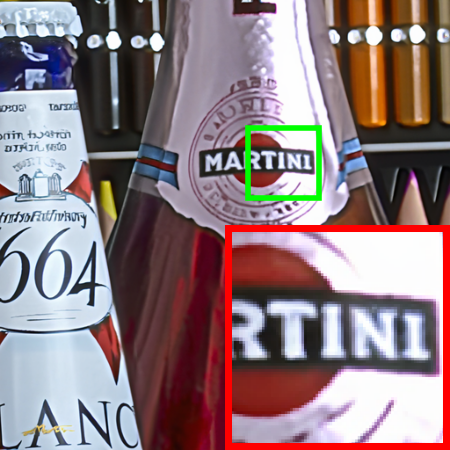} \\
\end{tabular}
\end{minipage}}
\vspace{-10pt}
\caption{Visual comparison on the image named ``Canon\_050'' from the RealSR dataset.}
\label{fig:Q3_RealSR_Canon_050}
\end{figure*}

\begin{figure*}[!t]
\setlength{\tabcolsep}{0.5pt}
\centering
\footnotesize
\resizebox{\linewidth}{!}{%
\begin{minipage}{\linewidth}
\centering
\begin{tabular}{@{}ccccccc@{}}
Input
& Real-ESRGAN~\cite{Real-esrgan}
& FeMASR~\cite{chen2022real}
& SwinIR~\cite{liang2021swinir}
& StableSR~\cite{wang2024exploiting}
& DiffBIR~\cite{lin2024diffbir}
& SeeSR~\cite{wu2024seesr} \\
\includegraphics[width=0.14\textwidth]{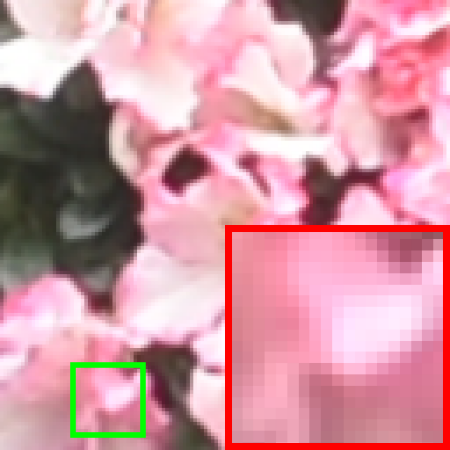}
& \includegraphics[width=0.14\textwidth]{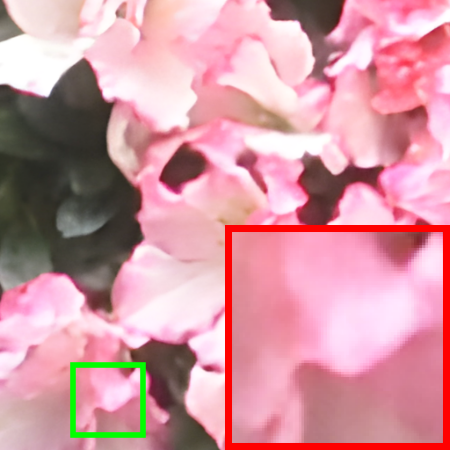}
& \includegraphics[width=0.14\textwidth]{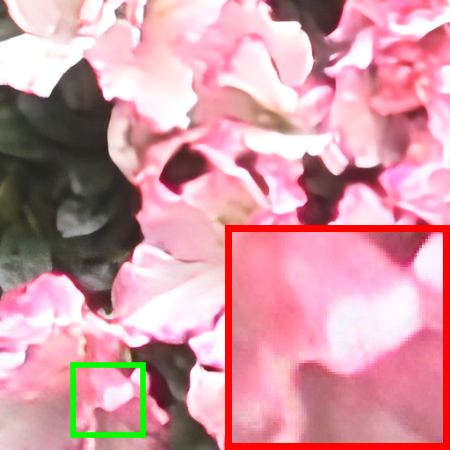}
& \includegraphics[width=0.14\textwidth]{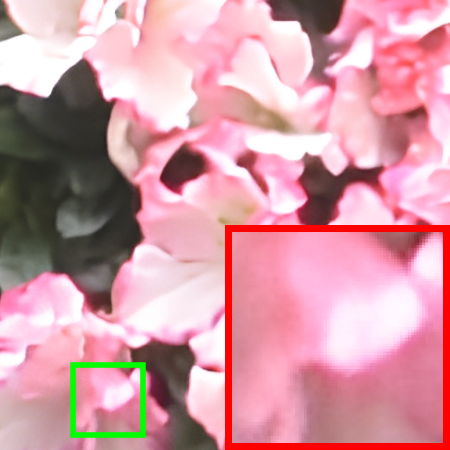}
& \includegraphics[width=0.14\textwidth]{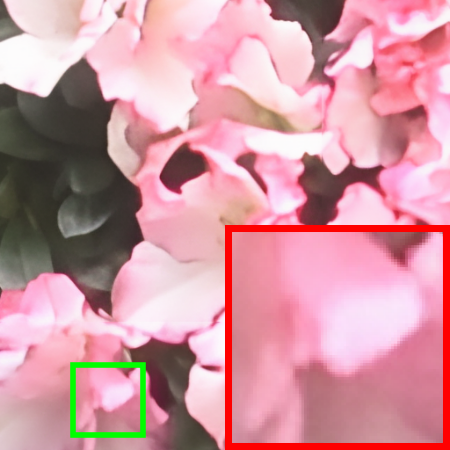}
& \includegraphics[width=0.14\textwidth]{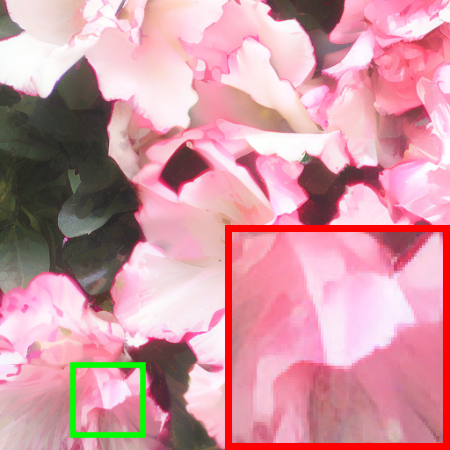}
& \includegraphics[width=0.14\textwidth]{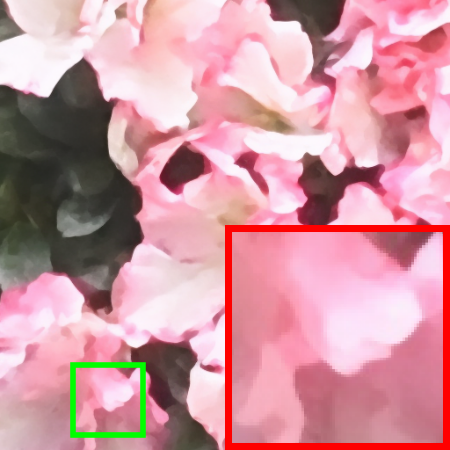} \\[1pt]

PASD~\cite{yang2024pixel}
& ResShift~\cite{yue2024resshift}
& SinSR~\cite{wang2024sinsr}
& OSEDiff~\cite{wu2024one}
& S3Diff~\cite{zhang2024degradation}
& PURE~\cite{wei2025perceive}
& \textbf{UARE (Ours)} \\
\includegraphics[width=0.14\textwidth]{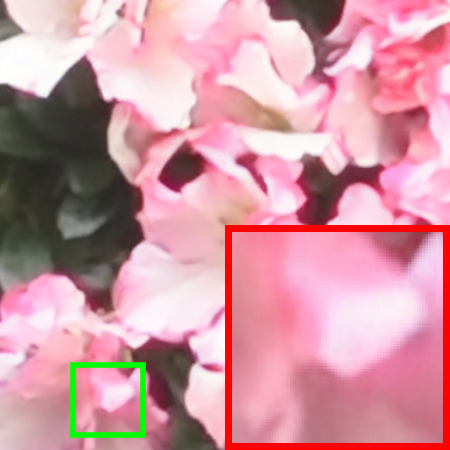}
& \includegraphics[width=0.14\textwidth]{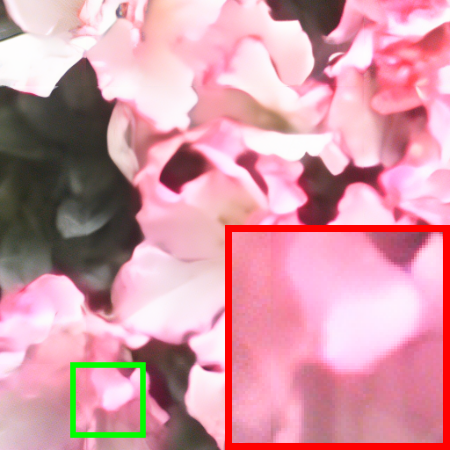}
& \includegraphics[width=0.14\textwidth]{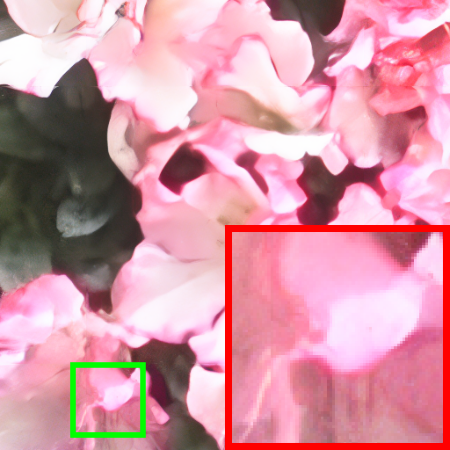}
& \includegraphics[width=0.14\textwidth]{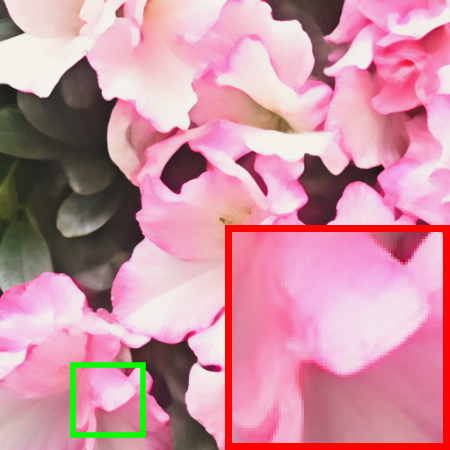}
& \includegraphics[width=0.14\textwidth]{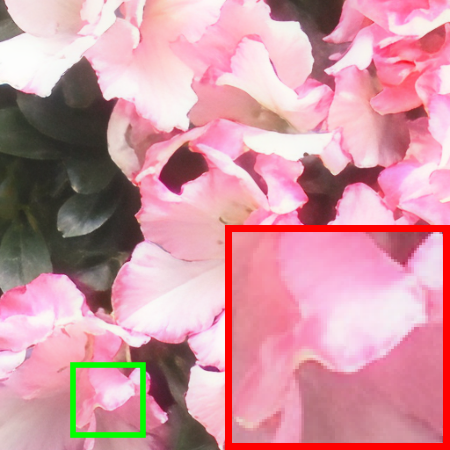}
& \includegraphics[width=0.14\textwidth]{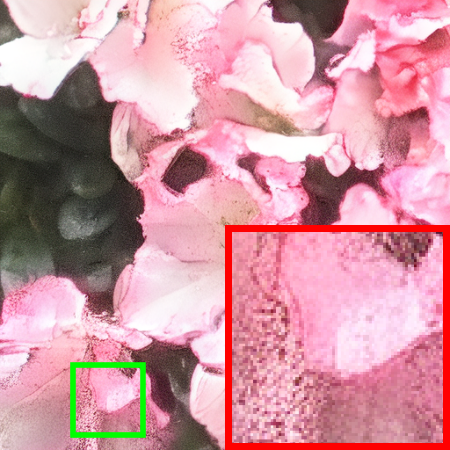}
& \includegraphics[width=0.14\textwidth]{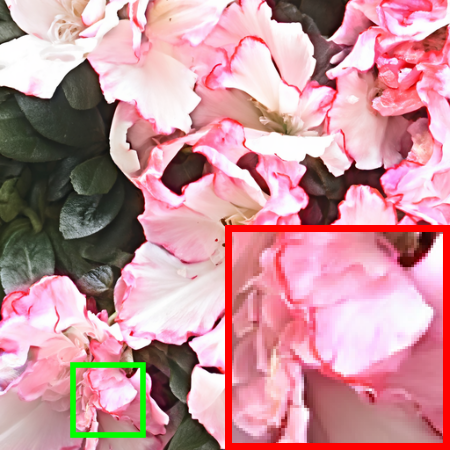} \\
\end{tabular}
\end{minipage}}
\vspace{-10pt}
\caption{Visual comparison on the image named ``DSC\_1425\_x1'' from the DRealSR dataset.}
\label{fig:Q5_DRealSR_1425}
\end{figure*}

\begin{figure*}[!t]
\setlength{\tabcolsep}{0.5pt}
\centering
\footnotesize
\resizebox{\linewidth}{!}{%
\begin{minipage}{\linewidth}
\centering
\begin{tabular}{@{}ccccccc@{}}
Input
& Real-ESRGAN~\cite{Real-esrgan}
& FeMASR~\cite{chen2022real}
& SwinIR~\cite{liang2021swinir}
& StableSR~\cite{wang2024exploiting}
& DiffBIR~\cite{lin2024diffbir}
& SeeSR~\cite{wu2024seesr} \\
\includegraphics[width=0.14\textwidth]{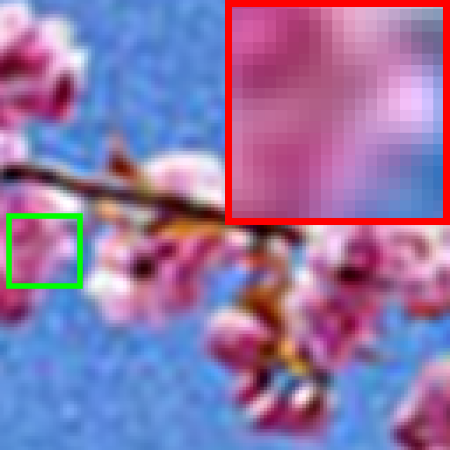}
& \includegraphics[width=0.14\textwidth]{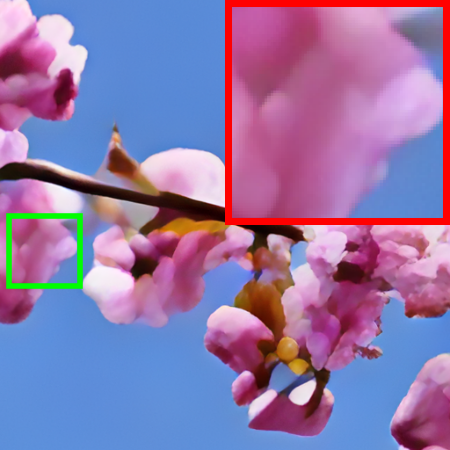}
& \includegraphics[width=0.14\textwidth]{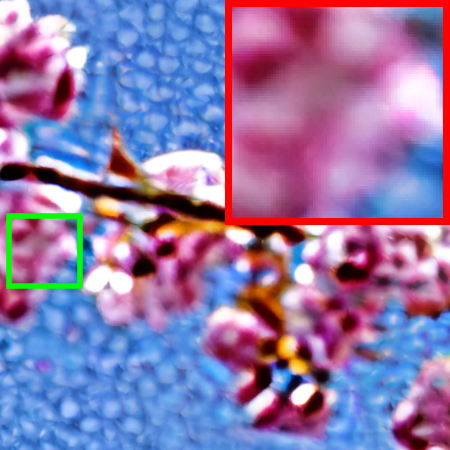}
& \includegraphics[width=0.14\textwidth]{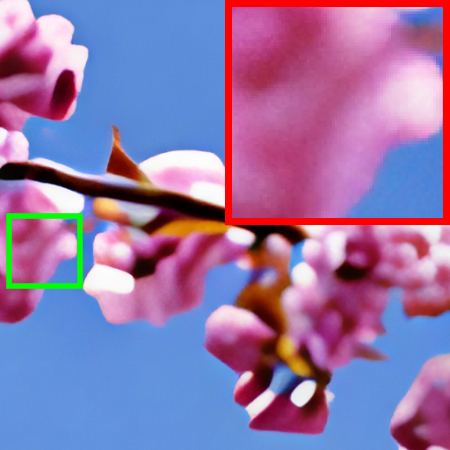}
& \includegraphics[width=0.14\textwidth]{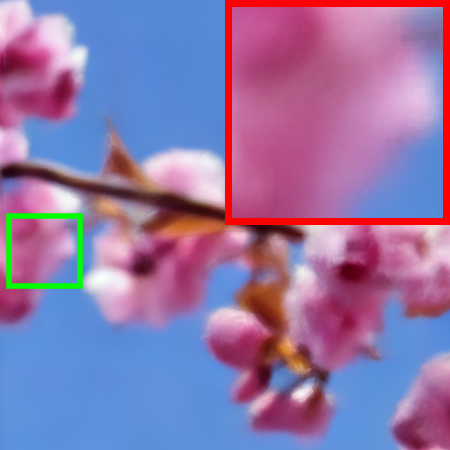}
& \includegraphics[width=0.14\textwidth]{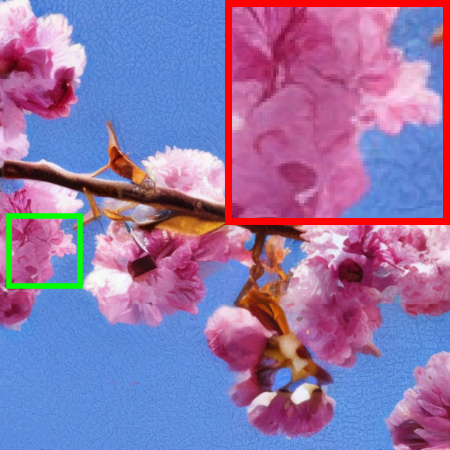}
& \includegraphics[width=0.14\textwidth]{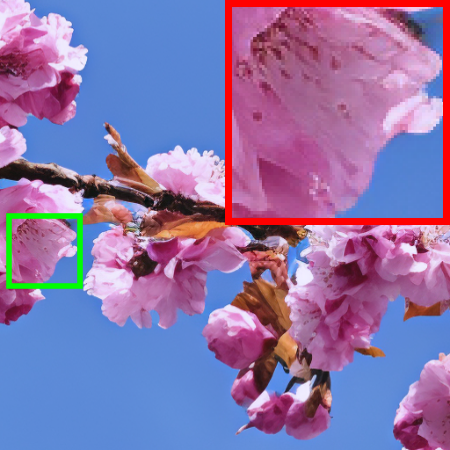} \\[1pt]

PASD~\cite{yang2024pixel}
& ResShift~\cite{yue2024resshift}
& SinSR~\cite{wang2024sinsr}
& OSEDiff~\cite{wu2024one}
& S3Diff~\cite{zhang2024degradation}
& PURE~\cite{wei2025perceive}
& \textbf{UARE (Ours)} \\
\includegraphics[width=0.14\textwidth]{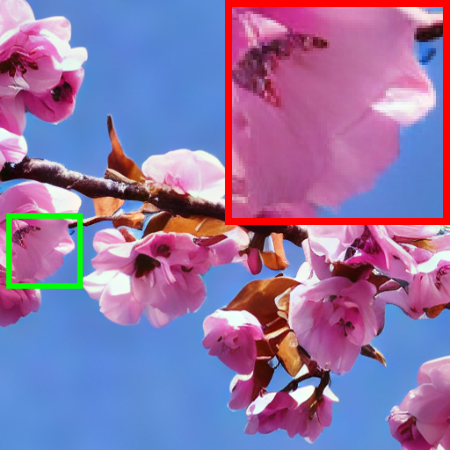}
& \includegraphics[width=0.14\textwidth]{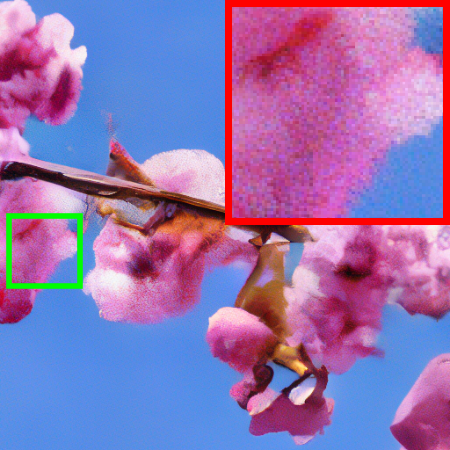}
& \includegraphics[width=0.14\textwidth]{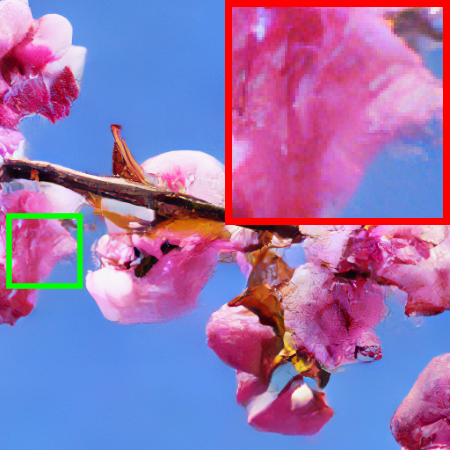}
& \includegraphics[width=0.14\textwidth]{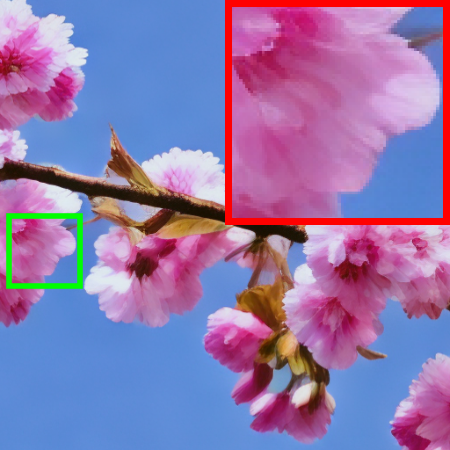}
& \includegraphics[width=0.14\textwidth]{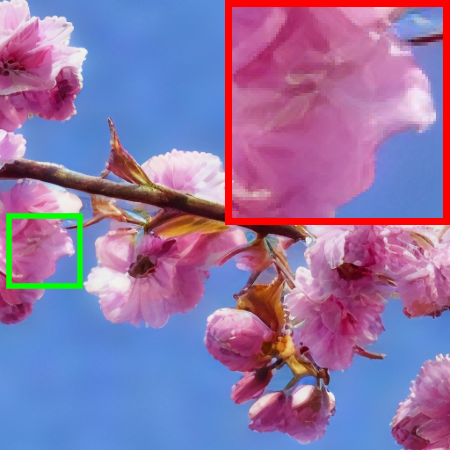}
& \includegraphics[width=0.14\textwidth]{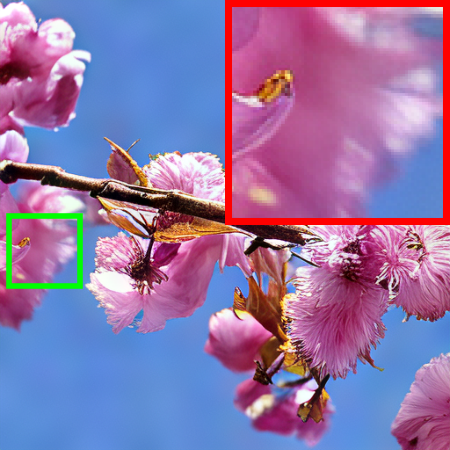}
& \includegraphics[width=0.14\textwidth]{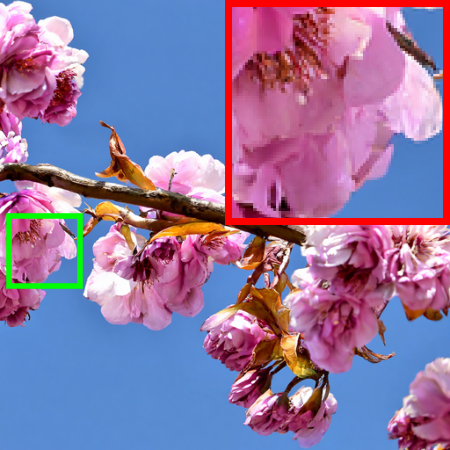} \\
\end{tabular}
\end{minipage}}
\vspace{-10pt}
\caption{Visual comparison on the image named ``0000098'' from the DIV2K dataset.}
\label{fig:Q9_DIV2K512_0000098}
\end{figure*}

\begin{table}[!t]
\centering
\caption{User study results of different SR methods.}
\vspace{-6pt}
\setlength{\tabcolsep}{5.2pt}
\resizebox{\linewidth}{!}{
\renewcommand{\arraystretch}{1.15}
\begin{tabular}{l | c c c c}
\shline
Method &  OSEDiff & S3Diff & PURE & \textbf{UARE (Ours)} \\
\hline \hline
Total Votes & 25 & \secondbest{48} & \third{31} & \best{186} \\
Voting Rate (\%) & 8.62 & \secondbest{16.55} & \third{10.69} & \best{64.14} \\
\shline
\end{tabular}}
% \vspace{-10pt}
\label{tab:user_study_sr}
\end{table}
\begin{table}[!t]
\centering
\caption{User study results of different restoration methods.}
\vspace{-6pt}
\setlength{\tabcolsep}{5.2pt}
\resizebox{\linewidth}{!}{
\renewcommand{\arraystretch}{1.15}
\begin{tabular}{l | c c c c}
\shline
Method & DiffIR & DiffUIR & FoundIR & \textbf{UARE (Ours)} \\
\hline \hline
Total Votes & 5 & \third{15} & \secondbest{20} & \best{250}\\
Voting Rate (\%)  & 1.72  & \third{5.17} & \secondbest{6.90} & \best{86.21} \\
\shline
\end{tabular}}
\label{tab:user_study_ir}
\end{table}

\subsection{User Study}
\label{sec:B.3}
To further evaluate the effectiveness of our UARE, we conduct a user study comparing four SR and restoration methods, respectively. We employ ten LR images from the RealSR, DRealSR and DIV2K datasets, and ten LQ images from the FoundIR test set. Compared SR methods include OSEDiff~\cite{wu2024one}, S3Diff~\cite{zhang2024degradation} and PURE~\cite{wei2025perceive}, while restoration methods include DiffIR~\cite{DiffIR}, DiffUIR~\cite{DiffUIR} and FoundIR~\cite{li2024foundir}. Twenty-nine expert researchers are invited to choose the best super-resolution/restored image for each test sample based on two equally weighted criteria: (1) perceptual quality, focusing on clarity, detail, and realism, and (2) content consistency with the LR/LQ input, including alignment in image structure and texture.

As reported in Tab.~\ref{tab:user_study_sr}, UARE achieves a high voting rate of 64.14\% in comparison with SR methods, which is significant better preference than other methods. Besides, as shown in Tab.~\ref{tab:user_study_ir}, UARE achieves a voting rate of 86.21\% in comparison of different all-in-one restoration methods. These results show that users overwhelmingly prefer UARE over both SR and restoration baselines. In particular, UARE receives nearly four times as many votes as the best competing SR method and more than an order of magnitude more votes than the strongest all-in-one restoration baseline. These consistent user preferences demonstrate that UARE achieves a better balance between perceptual quality and content fidelity, as well as our unified IQA-and-restoration scheme.

% \clearpage
\section{Discussion and Limitations}
\label{sec:C}
Due to the large number of parameters in the unified model, UARE has a relatively large model size and slow inference speed, which limits its deployment on resource-constrained devices. In addition, although we demonstrate that IQA can boost restoration and enhancement performance, how restoration and enhancement, in turn, can better improve IQA remains an open question and requires further study.

% \clearpage
{
    \small
    \bibliographystyle{ieeenat_fullname}
    \bibliography{main}
}

% \input{X_suppl}

\end{document}